\documentclass[10pt]{article} 
\usepackage[preprint]{tmlr}

\usepackage{amsmath,amsfonts,bm}

\def\eqref#1{equation~\ref{#1}}

\def\1{\bm{1}}

\DeclareMathAlphabet{\mathsfit}{\encodingdefault}{\sfdefault}{m}{sl}
\SetMathAlphabet{\mathsfit}{bold}{\encodingdefault}{\sfdefault}{bx}{n}

\usepackage{hyperref}
\usepackage{url}

\usepackage[utf8]{inputenc} 
\usepackage[T1]{fontenc}    
\usepackage{hyperref}       
\usepackage{url}            
\usepackage{booktabs}       
\usepackage{amsfonts}       
\usepackage{nicefrac}       
\usepackage{microtype}      
\usepackage{xcolor}         
\usepackage{graphicx}
\usepackage{svg}
\usepackage{array} 
\usepackage{tabularx}
\newcolumntype{C}{>{\centering\arraybackslash}X}
\usepackage{wrapfig}
\usepackage{pifont}
\newcommand{\cmark}{\ding{51}} 
\newcommand{\xmark}{\ding{55}} 
\usepackage{multirow}
\usepackage[most]{tcolorbox}
\usepackage{subcaption}
\usepackage{pifont}
\usepackage{enumitem}
\tcbuselibrary{breakable}
\newtcolorbox{promptbox}[1]{
  breakable,
  colback=gray!10,
  colframe=gray!50,
  boxrule=0.5pt,
  arc=4pt,
  left=6pt,
  right=6pt,
  top=6pt,
  bottom=6pt,
  parskip=6pt,
  fontupper=\ttfamily,
  title=#1,
  coltitle=black,
  fonttitle=\normalfont\bfseries\large,
  halign title=center
}

\title{Observe, Share, or Predict? Disentangling Partner\\ Information in Embodied Multi-Agent Collaboration}

\author{\name Adrián Szlatincsán \email aszlatincsan@fbk.eu \\
      \addr Fondazione Bruno Kessler \\
      University of Trento
      \AND
      \name Filippo Ziliotto \email fziliotto@fbk.eu\\
      \addr Fondazione Bruno Kessler
      \AND
      \name Bruno Lepri \email lepri@fbk.eu\\
      \addr Fondazione Bruno Kessler
      \AND
      \name Luciano Serafini \email serafini@fbk.eu\\
      \addr Fondazione Bruno Kessler
      \AND
      \name Massimiliano Luca \email mluca@fbk.eu\\
      \addr Fondazione Bruno Kessler
      \AND
      \name Tommaso Campari \email tcampari@fbk.eu\\
      \addr Fondazione Bruno Kessler}

\def\month{MM}  
\def\year{YYYY} 
\def\openreview{\url{https://openreview.net/forum?id=XXXX}} 

\begin{document}

\maketitle

\begin{abstract}
Large language models (LLMs) are increasingly used as high-level controllers for embodied agents, but it remains unclear whether they support collaborative reasoning, such as anticipating a collaborator's actions in real-time teamwork without dialogue. We ask how three ingredients of partner information, namely observing the partner's state, exchanging object knowledge, and explicitly predicting the partner's next action, separately shape task outcomes and coordination when agents can share percepts but never intentions, plans, or task-level messages, and whether the two agents occupy symmetric roles. Using the Habitat-based PARTNR benchmark, we introduce a Decoupled Partial Observability regime that enforces strictly individual perception, and we augment decentralized zero-shot ReAct planners with Other Agent's State (OAS) observation, object Knowledge Exchange (KE) between co-located agents, and Other Agent's Next Action (OANA) prediction, evaluated on the full $2\times2\times2$ grid with four planners on 1,000 validation tasks. OAS observation primarily improves task quality, reducing extraneous effort for every planner, whereas KE primarily improves execution efficiency, cutting simulation steps by up to 25\% with little change in success. The two are largely redundant for strong planners but complementary for the weakest, for which knowledge exchange without the partner state is the worst configuration. OANA prediction lowers the success rate in fifteen of sixteen model$\times$context pairs while raising explicit partner mentions to 55--100\%, and its accuracy gains lie mainly in the coarse action type, partly because agents recognize the action the partner is already performing. On average, the robot picks up objects earlier and usually finishes its share first, while the simulated human keeps working afterwards, and shared context helps mainly by shortening this period in which only the human is still working. Coordination gains thus come from reacting to shared context, chiefly the partner's observed state, rather than from explicit anticipation, and partner modeling in this setting is shallow and planner-dependent.
\end{abstract}

\section{Introduction}
\label{sec:introduction}


\begin{figure}[th] 
    \centering
    \includegraphics[width=0.98\linewidth]{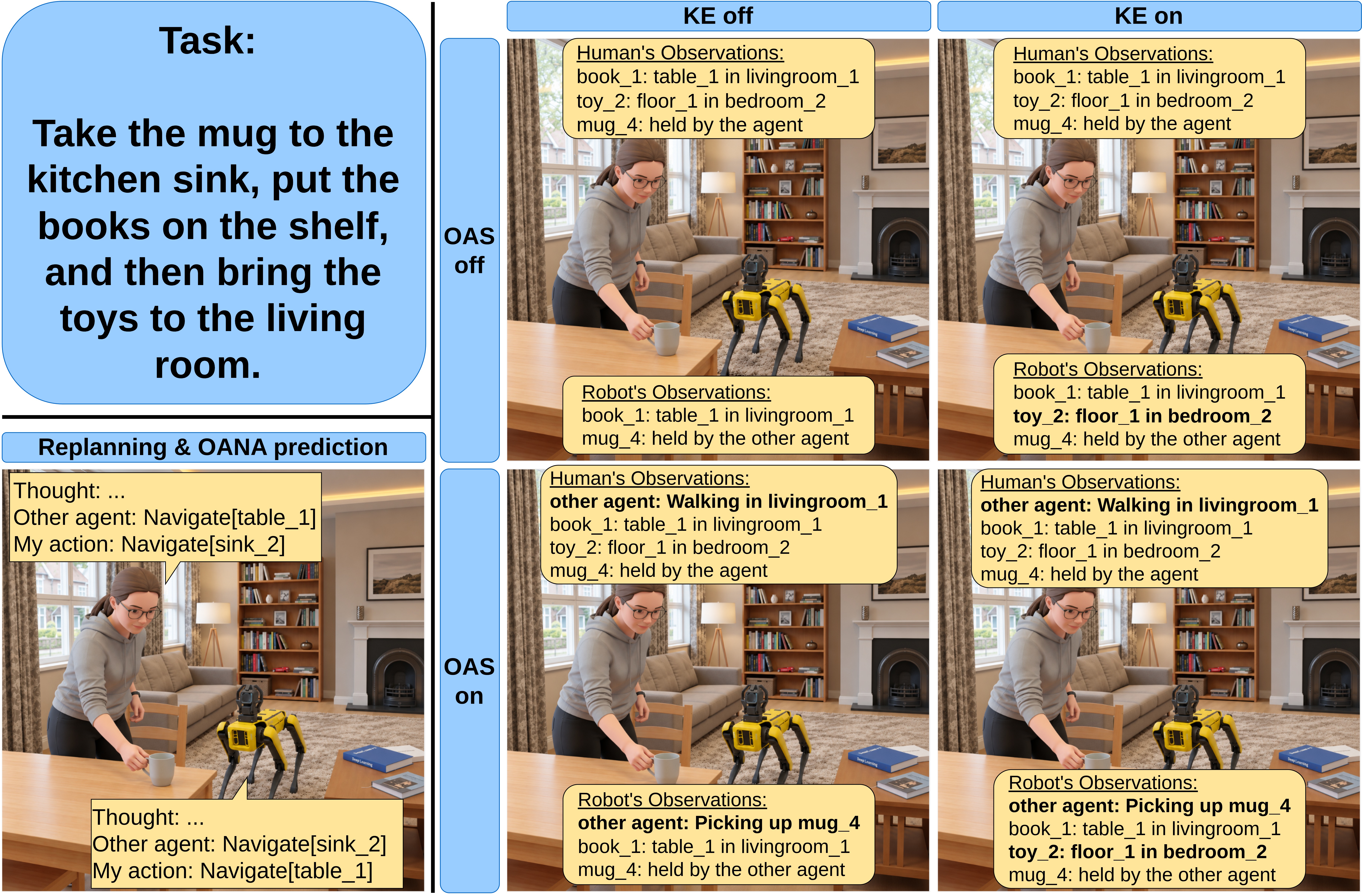}
    \caption{\textbf{Observation settings and OANA prediction.} In the object-only baseline (KE and OAS both off) of Decoupled Partial Observability, agents observe only objects and their positions (including whether an agent holds them). With object KE, co-located agents synchronize their accumulated object knowledge. With OAS observation, agents also observe the partner's state, but only when co-located. With OANA prediction, each replanning step predicts the partner's next action before the agent selects its own. Text in the figure is simplified relative to the prompts; the background image is AI-generated.}
    \label{fig:teaser}
    \label{fig:overview}
\end{figure}

Large language models (LLMs) are increasingly used as high-level controllers for embodied agents that collaborate with each other or with humans in complex, partially observable environments, such as household assistance and human--robot teamwork in realistic homes \citep{Glocker2025LLM-Empowered, Li2025Large, Chang2024PARTNR:, Chen2023Scalable, Zhang2023Building}. LLM-based multi-agent systems coordinate, deliberate, and negotiate through rich communication channels and plan flexibly on long-horizon embodied tasks \citep{Xi2023The, Glocker2025LLM-Empowered, Li2025Large, Chen2023Scalable, Zhang2023Building}, and a growing literature suggests that LLMs exhibit emergent Theory-of-Mind-like (ToM-like) abilities \citep{Kosinski2023Evaluating, Li2023Theory, Yang2025Large}, which would help in decentralized control under partial observability, a hard problem in multi-agent planning and learning \citep{Amato2019Modeling, Omidshafiei2017Deep, DeWeese2024Locally}, although whether such abilities are robust, and whether they transfer from text-based question answering into behavior, remains contested \citep{Shapira2023Clever, Gu2024SimpleToM:}.
 
However, in decentralized embodied settings, LLM agents still struggle with coordination, frequently exhibiting extraneous behaviors such as redundant task execution, repeated subtasks, and irrelevant actions \citep{Chang2024PARTNR:, Chen2023Scalable, Li2025Evaluation, Zhang2023Building}. PARTNR \citep{Chang2024PARTNR:} shows that access to the partner's intent, enabled via centralized control, can improve task efficiency, but such benefits are less understood under decentralized control \citep{Chang2024PARTNR:, Li2025Evaluation}. Current evaluations also often confound perception, planning, communication, and social or collaborative reasoning, and rarely isolate how a collaborator's observable state and explicit partner prediction affect real-time coordination \citep{Chang2024PARTNR:, Chen2023Scalable, Zhang2023Building}. Three conflations are particularly consequential: (i) what an agent knows about its partner and what it knows about the world its partner acts in are often supplied together as one notion of shared context; (ii) coordination metrics are often aggregated over the team, presupposing that agents running identical planners occupy symmetric roles; and (iii) reacting to a partner's observable state is rarely separated from anticipating its behavior, so improvements in coordination are readily read as evidence of partner modeling. It therefore remains unclear what partner modeling LLM agents actually perform, and when it improves planning, division of labor, temporal alignment, and task success, particularly when agents can exchange percepts but no intentions or plans \citep{Chang2024PARTNR:, Li2023Theory, Zhang2023Building}.
 
To address these gaps, we propose a controlled framework
\footnote{Code and evaluation scripts are available at \url{https://github.com/szlAdrian/observe-share-or-predict}.} that disentangles observability, partner-state access, object-knowledge sharing, and explicit partner-action prediction. Building on the PARTNR benchmark \citep{Chang2024PARTNR:}, which runs in Habitat 3.0 \citep{puig2023habitat}, we introduce (i) a \textbf{Decoupled Partial Observability} regime that, unlike PARTNR's partial observability, enforces strictly individual perception for both agents; (ii) a structured \textbf{Other Agent's State (OAS)} observation and (iii) an object \textbf{Knowledge Exchange (KE)} mechanism for co-located agents that transmits object positions and states without sender attribution, which we manipulate independently of each other; and finally (iv) explicit \textbf{Other Agent's Next Action (OANA)} prediction within decentralized zero-shot ReAct \citep{yao2022react} planners, scored both leniently and strictly so that recognizing a partner's ongoing action is not credited as anticipating it. We evaluate the full $2\times2\times2$ grid of these factors with four planners (Llama-3.1-8B, 4-bit Llama-3.1-70B, Gemini 3.1 Flash Lite, and DeepSeek V3.2) on 1,000 validation tasks, add a full-observability control, and report coordination metrics per agent as well as for the team. Figure~\ref{fig:teaser} illustrates the observation settings and OANA prediction. Our main findings are:
\begin{itemize}[leftmargin=*,nosep]
\item \textbf{The two context channels act on different axes} (Section~\ref{sec:task}). OAS observation improves task quality, reducing extraneous effort for every planner, while KE improves execution efficiency, cutting simulation steps by up to 25\% with little change in success. They are largely redundant for strong planners but complementary for the weakest, for which object knowledge without partner state is the worst configuration we measure.
\item \textbf{Explicit prediction lowers success} (Sections~\ref{sec:task} and~\ref{sec:robust}). OANA prediction lowers the success rate in fifteen of sixteen model$\times$context pairs, and in nineteen of twenty once full observability is included, while driving partner mentions to 55--100\%.
\item \textbf{Partner modeling is shallow} (Sections~\ref{sec:prediction} and~\ref{sec:robust}). Shared context improves predictions of the coarse action type much more than of its arguments. Part of the measured accuracy reflects recognizing the partner's ongoing action rather than anticipating its next one, and even with complete world knowledge, the Llama planners name their partner's next action type wrongly more than half the time.
\item \textbf{The two agents take different roles} (Section~\ref{sec:roles}). The robot picks up its first object earlier and usually completes its share of the work first, whereas the simulated human keeps working after the robot has finished and therefore determines when the episode ends. Shared context helps mainly by shortening this final period in which only the human is still working, while the division of labor between the two agents stays nearly the same.
\end{itemize}
The coordination gains we observe therefore come from agents reacting to shared context, chiefly structured partner state; explicit anticipation of the partner does not reliably add to them. For the debate on Theory of Mind in LLMs, this is a cautionary result: in embodied collaboration, neither coordination gains nor partner-mention frequency nor prediction accuracy that credits a partner's ongoing action should necessarily be read as evidence of mental-state attribution.

\section{Related Work}
\label{sec:relworks}

\paragraph{Implicit Multi-Agent Coordination in Embodied LLMs.} Much of the recent research on LLM-based multi-agent systems focuses on explicit communication, such as dialogue \citep{Zhang2023Exploring,Estornell2024ACC-Collab:,Han2024LLM,Li2024A,Plaat2025Agentic}. While these works establish that LLMs can coordinate through deliberation, they do not fully capture the constraints of embodied settings where messaging and dialogue are limited or may be unavailable. 
Social conventions can emerge in LLM populations through repeated interaction alone, without explicit negotiation \citep{Ashery2024Emergent}, and LLM-generated narratives can steer cooperation in multi-agent systems \citep{Curto2025LLM-Driven}, although scaling LLM agents to populations raises validity concerns \citep{Taillandier2025Integrating}. Communication itself does not guarantee coordination: it can reinforce biases \citep{Buscemi2025Strategic}, and collaborative outcomes depend on whether agents reach common ground \citep{Nath2025Lets}.
In contrast to dialogue-based systems, we focus on embodied coordination in which agents can exchange percepts but never intentions, plans, or task-level messages, so that partner intent must be inferred from task dynamics and observation rather than declared.

\paragraph{Embodied Planning and Partial Observability.} In embodied AI, LLMs serve as high-level controllers for task decomposition and coordination across agents in shared environments \citep{Zhang2023Building,White2025Collaborating,Wang2025Multi-Agent,Seo2026From}. While LLMs improve flexibility in long-horizon tasks, decentralized collaboration frequently suffers from partial observability and redundant or extraneous effort \citep{White2025Collaborating, Chang2024PARTNR:}, and an agent's world model goes stale when its partner changes the world out of view. Modern frameworks mitigate these problems through modular agent architectures or collaborative belief models \citep{Zhang2023Building, Wang2025CoBel-World:}. However, most evaluations assess collaboration holistically, making it difficult to isolate whether gains stem from perception, planning, or social or collaborative reasoning. Our work differs by assuming oracle action execution and privileged perception to factor out low-level errors, allowing us to directly test whether access to partners' observable states, exchanging object knowledge, and explicit modeling improve decentralized performance.


\paragraph{Theory of Mind and Partner Modeling.} A central debate is whether LLM agents possess a functional Theory of Mind (ToM), the ability to represent others' latent mental states \citep{Li2023Theory,Gandhi2023Understanding,Street2024LLMs,Yang2025Large}. While LLMs perform well on text-only ToM benchmarks, whether these abilities are robust and transfer into behavior is contested \citep{Shapira2023Clever,Gu2024SimpleToM:}. We do not probe mental-state attribution directly: our agents are never queried about a partner's beliefs, and OAS observations deliberately exclude internal state. Instead, we study partner modeling at the level of behavior that ToM is ultimately invoked to explain, anticipating a collaborator's next action, and ask whether it improves joint behavior. Related work infers a partner's goals or intentions in order to coordinate \citep{Wang2021ToM2C:, Trivedi2025An} or reasons from a partner's visual perspective \citep{Patania2025PerspAct:}. 
Work that equips LLM agents with explicit partner models, such as Hypothetical Minds \citep{Cross2024Hypothetical}, which generates and refines natural-language hypotheses about other agents' strategies, evaluates the modeling component by ablation under a fixed observation of the other agents. We additionally manipulate that observation: observable partner-state access (OAS) and explicit action prediction (OANA) are varied independently, and predictions are scored both leniently and strictly so that recognizing what a partner is already doing is not credited as anticipating it.
This yields a behavioral test of whether explicit partner modeling translates into more efficient and successful collaboration, an embodied counterpart to the gap between explicit ToM inference and its application reported by \citet{Gu2024SimpleToM:}.
\section{Method}
\label{sec:method}

\begin{figure*}[t]
    \centering
    \includegraphics[width=\textwidth]{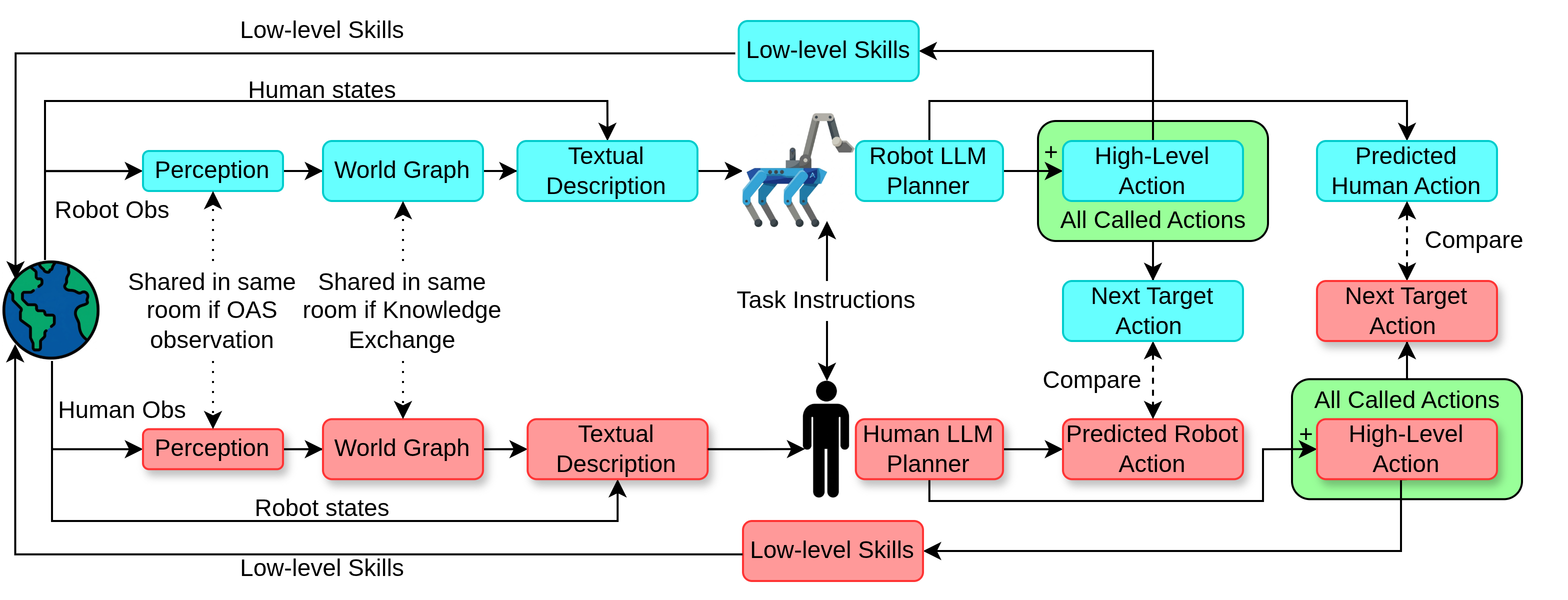}
    \caption{\textbf{Decoupled architecture with Other Agent's State (OAS) observation, object Knowledge Exchange (KE), and Other Agent's Next Action (OANA) prediction}. Both agents build a world graph using their own environmental object observations and their actions. The graph's nodes represent the house, rooms, furniture, receptacles, objects, and agents. The edges represent the type of relation between them. If allowed, the agents also observe the OAS and exchange their knowledge about the objects when they are in the same room. Their high-level LLM planners call a High-Level Action (at replanning) and the corresponding low-level skills (action execution in the simulator), and predict the OANA. The predicted OANA is then compared with the other agent's current and next actions (union target, Section \ref{sec:oana}), retrieved from its list of all called high-level actions.}
    \label{fig:decoupled_architecture}
\end{figure*}

Our goal is to characterize how two embodied LLM agents coordinate while collaborating on a shared task. We use \emph{collaboration} for the joint pursuit of that task and judge it by team outcomes (Success Rate, Percent Complete, Simulation Steps), and \emph{coordination} for how the agents manage the dependencies between their actions within it \citep{Malone1994Interdisciplinary}: avoiding duplicated work, aligning in time, and dividing the subtasks. \emph{Partner modeling} refers to what an agent represents about its partner, from reacting to its observed state to anticipating its next action. We ask whether observing the OAS, sharing object knowledge through KE, and predicting the OANA reduce coordination failures such as timing conflicts, race conditions, and extraneous actions; how each affects task outcomes and coordination; and whether the two agents take different roles. 
Section~\ref{sec:setting} describes the agents and how they plan, Section~\ref{sec:info} describes what each agent knows about the objects and its partner under each condition, and Section~\ref{sec:oana} describes how agents predict their partner's next action and how these predictions are scored. Figure~\ref{fig:decoupled_architecture} illustrates the architecture, adapted from PARTNR \citep{Chang2024PARTNR:}.

\subsection{Setting and Agents}
\label{sec:setting}
 
A simulated human ($h$) and a robot ($r$) work on a shared household task, such as moving plates from a dining table to the kitchen. The tasks come from the PARTNR benchmark \citep{Chang2024PARTNR:}, which runs in Habitat~3.0 \citep{puig2023habitat} in multi-room homes from the Habitat Synthetic Scenes Dataset \citep{khanna2024habitat}. We write $i \in \{h, r\}$ for an agent and $j$ for its partner. Control is decentralized: each agent runs its own LLM planner, and the agents send each other no messages; the only channels between them are those of Section~\ref{sec:info}. Both agents know the layout of the house (rooms and furniture) from the start, but have to find the objects.
 
\textbf{World graph.} Each agent stores what it knows in a world graph $g_{i,t}$, whose nodes are the house, rooms, furniture, receptacles, objects, and agents, and whose edges are the relations between them (e.g., \texttt{cup\_1 on table\_2}). The index $t$ counts low-level simulator steps, the unit of the Simulation Steps metric. How $g_{i,t}$ is updated is the subject of Section~\ref{sec:info}.
 
\textbf{Planning loop.} Each agent is a zero-shot ReAct planner \citep{yao2022react}. Let $\tau_i^{(k)}$ denote the simulator step at which agent $i$ replans for the $k$-th time; because execution is asynchronous, the replanning steps of the two agents rarely coincide. Agent $i$ keeps a prompt history $H_i^{(k)}$ that starts with the instructions and task, the descriptions of its action tools, and the layout. At its $k$-th replanning step, the agent receives a textual observation $o_i^{(k)}$, a summary of the objects in $g_{i,\tau_i^{(k)}}$, and its LLM outputs a reasoning trace $R_i^{(k)}$ and a high-level action $a_i^{(k)}$:
\begin{equation}
\label{eq:react}
\big(R_i^{(k)},\, a_i^{(k)}\big) \sim \mathrm{LLM}_i\big(\cdot \mid H_i^{(k-1)} \oplus o_i^{(k)}\big),
\qquad
H_i^{(k)} = H_i^{(k-1)} \oplus o_i^{(k)} \oplus R_i^{(k)} \oplus a_i^{(k)},
\end{equation}
where $\oplus$ appends text to the prompt. A high-level action has the form \texttt{<action\_type>[<parameters>]}, e.g., \texttt{Navigate[table\_0]} or \texttt{Explore[bedroom\_1]}. Oracle skills execute it as a sequence of low-level simulator steps, concurrently with the partner's ongoing action, so a single high-level action usually spans many steps~$t$.
 
\textbf{Action spaces.} We use the PARTNR action tools. The robot acts only through navigation and rearrangement primitives (e.g., \texttt{Navigate, Pick, Place, Open, Close}), whereas the human additionally has the object-state actions \texttt{Clean, Fill, Pour, PowerOn,} and \texttt{PowerOff,} so the human must contribute whenever these are needed. This gives the human agent 14 action types and the robot agent 9 (Appendix~\ref{app:skills}). An episode ends when both agents have called \texttt{Done[]}.

\subsection{Information Conditions}
\label{sec:info}
 
\textbf{Decoupled Partial Observability.} In PARTNR's partial observability setting, the robot also has access to the human's perception. To remove this asymmetric advantage and keep perception realistic, we introduce \emph{Decoupled Partial Observability}, in which each agent's world graph is updated only from its own field of view. Let $p_{i,t}$ be the objects in agent $i$'s field of view at step $t$, with their positions and states (including whether an agent holds them), and let $U(g, p)$ return $g$ with every object in $p$ added or moved to its perceived position and state. In the baseline, $g_{i,t} = U(g_{i,t-1}, p_{i,t})$. We read $p_{i,t}$ from the simulator (privileged perception), which removes the noise of learned object detectors and isolates the effects of the manipulations below. Two optional channels change this update, and only while the agents are co-located, that is, in the same room. With OAS observation, the update also uses the partner's current percepts $p_{j,t}$. With object KE, the agents then synchronize: $\mathrm{Sync}$ sets every object known to either agent, in both world graphs, to its true position and state in the simulator's global world graph $g^{\mathrm{global}}_t$. Writing $\tilde g_{i,t}$ for the graph after the perception update,
\begin{align}
\tilde g_{i,t} &=
\begin{cases}
U\big(g_{i,t-1},\, p_{i,t} \cup p_{j,t}\big) & \text{if OAS is enabled and the agents are co-located,}\\
U\big(g_{i,t-1},\, p_{i,t}\big) & \text{otherwise;}
\end{cases} \label{eq:oas_update}\\[4pt]
g_{i,t} &=
\begin{cases}
\mathrm{Sync}\big(\tilde g_{i,t},\, \tilde g_{j,t},\, g^{\mathrm{global}}_t\big) & \text{if KE is enabled and the agents are co-located,}\\
\tilde g_{i,t} & \text{otherwise.}
\end{cases} \label{eq:ke_update}
\end{align}
 
\textbf{Other Agent's State (OAS) observation.} To mimic real-world co-presence, OAS observation lets an agent see what its partner is doing. While the agents are co-located, the agent observes its partner's visible state $\mathrm{oas}_{j,t}$, such as \texttt{Walking in hallway\_1}, \texttt{Picking apple\_0}, or \texttt{Standing}; while they are in different rooms, it receives \texttt{Acting in another room}. These states contain no internal state of the partner, only what would be visible if the agent looked at it. They are read from simulator logs, so no action-recognition errors enter (a real system would need such a model, e.g., on RGB-D input), and the coordination capabilities of the planners can be studied in isolation. The observation $o_i^{(k)}$ in Eq.~(\ref{eq:react}) then consists of the object summary followed by the partner's states during $\big(\tau_i^{(k-1)}, \tau_i^{(k)}\big]$; if the partner's state did not change in this interval, the last observed state is repeated. In addition, on co-location, each agent adds its partner's current percepts to its own world graph, as in Eq.~(\ref{eq:oas_update}). This keeps its object knowledge consistent with the manipulations it watches the partner perform (e.g., an object the partner is carrying is no longer recorded at its old location). Unlike KE, it shares only what the partner perceives at that moment, not the object knowledge the partner has accumulated earlier in the episode.
 
\textbf{Object Knowledge Exchange (KE).}
\label{sec:kevsoas}
The synchronization in Eq.~(\ref{eq:ke_update}) resolves discrepancies that arise when the agents act independently in different rooms. Although Sync reads the simulator's global world graph $g^{\mathrm{global}}_t$, it adds no information that the two agents do not jointly hold: only the agents move objects or change their states, and each agent records its own manipulations, so for every object known to either agent, the more recent of the two records is up to date. Reading the value from $g^{\mathrm{global}}_t$ is therefore equivalent to keeping the more recent record and spares the agents from resolving conflicting records, which they could not do without dialogue. Object KE is an information channel between the agents, and we characterize it explicitly rather than describing our setting as communication-free. It carries only object positions and states, never intentions, plans, or task-level messages (\emph{content}); it merges exchanged knowledge into the receiver's world graph with no sender tag, so an agent cannot tell whether it perceived a change itself or acquired it from its partner (\emph{attribution}); it operates only while the agents share a room, as a consequence of physical co-presence (\emph{gating}); and it is read from the simulator rather than produced and parsed by the planners, so it is lossless and instantaneous, an upper bound on what an equivalent real protocol could deliver (\emph{fidelity}). Where KE fails to help, or harms a planner (Section~\ref{sec:task}), the cause therefore cannot be communication overhead, latency, or message error; it must lie in how the planner uses the information, most directly in whether it can infer agency from an unexplained change in the world.
 
Figure~\ref{fig:overview} shows the agents' observations under each condition, and Appendix~\ref{app:prompts} shows an example prompt.
 
\subsection{Other Agent's Next Action (OANA) Prediction}
\label{sec:oana}
 
When OANA prediction is enabled, the planner of agent $i$ first reasons about both agents' next actions, then predicts the partner's next high-level action $\hat a_j^{(k)}$, and then selects its own, so Eq.~(\ref{eq:react}) becomes
\begin{equation}
\label{eq:oana}
\big(R_i^{(k)},\, \hat a_j^{(k)},\, a_i^{(k)}\big) \sim \mathrm{LLM}_i\big(\cdot \mid H_i^{(k-1)} \oplus o_i^{(k)}\big),
\end{equation}
and all three outputs are appended to the history. With OANA enabled, the prompt also contains the descriptions of the partner's action tools, so part of any OANA effect could come from knowing what the partner can do. The human predicts over the robot's 9 action types and the robot over the human's 14, so the robot faces the larger prediction space. On average, an episode offers 11.7 ($\pm$3.8) rooms, 56.3 ($\pm$20.3) pieces of furniture, and 5.4 ($\pm$1.9) objects as possible action parameters.
 
\textbf{Targets.} Because the agents replan asynchronously, the partner is usually in the middle of an action when agent $i$ makes a prediction at $\tau = \tau_i^{(k)}$. Let $a_j^{\mathrm{cur}}(\tau)$ be the action the partner is executing at step $\tau$ and $a_j^{\mathrm{next}}(\tau)$ the action it selects at its first replanning step after $\tau$. We score each prediction against two target sets:
\begin{equation}
\label{eq:targets}
\mathcal{T}^{\mathrm{union}}_j(\tau) = \big\{a_j^{\mathrm{cur}}(\tau),\; a_j^{\mathrm{next}}(\tau)\big\},
\qquad
\mathcal{T}^{\mathrm{strict}}_j(\tau) = \big\{a_j^{\mathrm{next}}(\tau)\big\},
\end{equation}
and a prediction counts as correct if it matches any element of the target set. The union target allows for the asynchrony: an agent that observes \texttt{Walking in living\_room\_1} may name the ongoing \texttt{Navigate[table\_1]}, which the union target accepts, although the partner's next action is \texttt{Place[cup\_1, on, table\_1]}. Because this credits recognizing an ongoing action as if it were anticipation, we also report the strict target (Section~\ref{sec:robust}). Under both targets, once the partner has called \texttt{Done[]}, the only correct prediction is \texttt{Done[]}.
 
\textbf{Matching.} Predictions are compared at the class level: instance IDs are masked (e.g., \texttt{spoon\_3} $\rightarrow$ \texttt{spoon}, giving \texttt{Rearrange[spoon, on, table, next\_to, plate]}), so that an agent is not penalized for referring to an object it has not yet seen or to another instance of the same class. Matches are evaluated at several granularities, from the action type alone to all parameters (Section~\ref{sec:prediction}; protocol in Appendix~\ref{app:comparison}). For the Llama planners, the decoding grammar restricts predicted objects to the classes present in the scene, which are available from the start of the episode (Appendix~\ref{app:constraint}), an advantage the API-accessed planners lack. Both OAS observation and OANA prediction also lengthen the prompt. Appendix~\ref{app:impl} details the action space; further details are in \citet{Chang2024PARTNR:}.

\section{Experiments and Analysis}
\label{sec:experiments}

\subsection{Experimental Setup}
\label{sec:setup}

We conduct our experiments in the Habitat 3.0 \citep{puig2023habitat} simulator, extending the PARTNR \citep{Chang2024PARTNR:} framework with our proposed modifications. All experiments build on the PARTNR \textit{decentralized\_zero\_shot\_react\_summary} baseline configuration, which includes decentralized control, a zero-shot ReAct planner, privileged perception, oracle skills (i.e., simulator-based execution rather than learned action policies), and summarized object observations. We keep all these configuration settings unchanged, with a maximum timeout of 50 replanning steps and 20,000 simulator steps per agent, the LLM decoding temperature set to 0, and at most 250 generated tokens per replanning step. Evaluation is performed on the PARTNR validation set, comprising 13 house scenes, a diverse set of household objects, and 1,000 tasks. Due to limited computational resources, we report results from a single run per configuration. Prior work on the PARTNR baseline \citep{Chang2024PARTNR:} reports low variance across seeds. With 1,000 episodes, the binomial standard error of a single Success Rate is 1.2--1.6 points, so we interpret differences of a few points between individual runs with caution and base our conclusions mainly on effects that recur across configurations; for example, OANA prediction lowers Success Rate in 15 of 16 model$\times$context pairs, which a sign test treating the pairs as independent would assign a two-sided probability of $5 \times 10^{-4}$ under no effect. Multi-seed evaluation would nevertheless give more robust estimates, and we leave it for future work.


\textbf{Metrics.} We report task outcomes, which measure the collaboration as a whole, and coordination metrics, which measure how the two agents' activities fit together and are reported per agent where applicable. \textit{Simulation Steps} measures the number of simulator steps (simulator time) required to complete a task, reflecting execution efficiency. \textit{Success Rate} (SR) indicates whether all task constraints are satisfied (0 or 1 for an episode), while \textit{Percent Complete} (PC) captures partial progress by measuring the fraction of completed task propositions (equal to 1 for fully successful tasks). To assess planning efficiency, \textit{Planning Cycles} counts the number of high-level LLM calls made by each agent. We further analyze coordination through \textit{Other-Agent Mention Ratio}, defined as the fraction of replanning steps in which an agent explicitly mentions “other agent” or “agent b” in the reasoning part (the prompts refer to the partner by these terms; Appendix~\ref{app:prompts}). It therefore measures how often an agent considers its partner in its reasoning, not the quality of that reasoning. We also measure \textit{Exploration Efficiency} as the first low-level simulation step at which an agent successfully picks up an object. \textit{Task Offloading} quantifies the proportion of subtasks completed by the robot, while \textit{Extraneous Effort} measures the ratio of unnecessary rearrangements to total actions. Finally, \textit{Extra Planning Ratio} captures the proportion of replanning steps performed after the partner has finished acting, indicating inefficiencies in coordination, and \textit{Action Success Rate} reports the fraction of successfully executed actions. With OANA prediction enabled, the Extra Planning Ratio also gives the share of an agent's predictions whose target is Done[] (Section \ref{sec:oana}), since every prediction made after the partner has finished is scored against Done[].

\subsection{Task Outcomes and Coordination}
\label{sec:task}

\begin{table}[t]
\centering
\caption{\textbf{Task outcomes and redundant work under Decoupled Partial Observability} for every combination of Other Agent's State (OAS) observation, object Knowledge Exchange (KE) on co-location, and Other Agent's Next Action (OANA) prediction. SR: Success Rate (\%), PC: Percent Complete (\%), Steps: Simulation Steps, EE: Extraneous Effort (\%). The best value per model is bolded. All coordination metrics (Planning Cycles, Other-Agent Mention Ratio, Exploration Efficiency, Task Offloading, Extra Planning, Action Success), reported per agent, are in Table~\ref{tab:main_full} (Appendix~\ref{app:partial}).}
\label{tab:main}
\setlength{\tabcolsep}{2.6pt}
\footnotesize
\begin{tabular}{ccc|cccc|cccc|cccc|cccc}
\toprule
 & & & \multicolumn{4}{c|}{Llama-3.1-8B} & \multicolumn{4}{c|}{Llama-3.1-70B (4-bit)} & \multicolumn{4}{c|}{Gemini 3.1 Flash Lite} & \multicolumn{4}{c}{DeepSeek V3.2} \\
OAS & KE & OANA & SR$\uparrow$ & PC$\uparrow$ & Steps$\downarrow$ & EE$\downarrow$ & SR$\uparrow$ & PC$\uparrow$ & Steps$\downarrow$ & EE$\downarrow$ & SR$\uparrow$ & PC$\uparrow$ & Steps$\downarrow$ & EE$\downarrow$ & SR$\uparrow$ & PC$\uparrow$ & Steps$\downarrow$ & EE$\downarrow$ \\
\midrule
\xmark & \xmark & \xmark & 60.9 & \textbf{77.1} & 4214 & 62.3 & 76.0 & 86.9 & 3056 & 39.4 & 80.4 & 89.8 & 4118 & 44.3 & 82.3 & 91.0 & 3857 & 38.2 \\
\xmark & \xmark & \cmark & 55.0 & 73.0 & 4243 & 62.0 & 72.9 & 85.6 & 3648 & 43.5 & 76.0 & 88.6 & 4304 & 43.5 & 75.2 & 87.1 & 5486 & 35.0 \\
\xmark & \cmark & \xmark & 54.3 & 72.9 & 4914 & 69.6 & 76.9 & \textbf{87.6} & 3041 & 43.0 & 79.9 & 89.4 & 3081 & 38.2 & 82.2 & 90.6 & 2951 & 34.8 \\
\xmark & \cmark & \cmark & 56.2 & 72.9 & 4121 & 62.7 & 72.1 & 84.6 & 3125 & 42.1 & 75.8 & 88.3 & 3416 & 35.1 & 75.6 & 87.4 & 4249 & 31.9 \\
\midrule
\cmark & \xmark & \xmark & 58.8 & 74.7 & 4022 & \textbf{58.5} & 76.2 & 86.5 & 3425 & \textbf{36.5} & \textbf{81.2} & 89.8 & 3193 & 33.0 & \textbf{83.6} & \textbf{91.5} & 3112 & 28.1 \\
\cmark & \xmark & \cmark & 56.6 & 74.2 & 4017 & \textbf{58.5} & 72.5 & 84.3 & 3019 & 38.0 & 79.1 & 88.6 & 3453 & 32.2 & 78.4 & 88.8 & 3927 & \textbf{26.7} \\
\cmark & \cmark & \xmark & \textbf{62.3} & 76.8 & \textbf{3781} & \textbf{58.5} & \textbf{77.5} & \textbf{87.6} & 3095 & \textbf{36.5} & 81.1 & \textbf{90.3} & \textbf{2911} & 33.3 & 82.4 & 90.5 & \textbf{2794} & 27.5 \\
\cmark & \cmark & \cmark & 57.2 & 74.5 & 3890 & 59.1 & 72.8 & 84.6 & \textbf{2849} & 36.8 & 78.6 & 89.1 & 3107 & \textbf{32.1} & 80.0 & 89.4 & 3551 & 27.0 \\
\bottomrule
\end{tabular}
\end{table}
 
\begin{figure}[t]
\centering
\includegraphics[width=\textwidth]{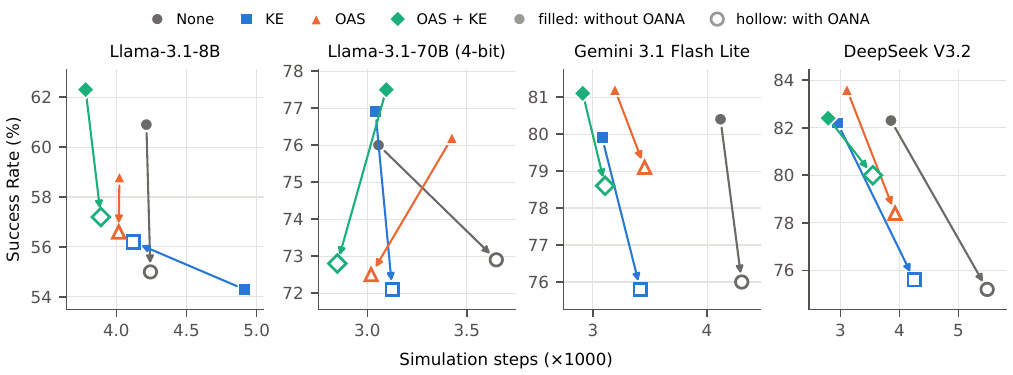}
\caption{\textbf{Success Rate against episode length (Simulation Steps) for every configuration in Table~\ref{tab:main}.} Marker shape and color give the shared context (none, KE, OAS, OAS + KE); filled markers are runs without OANA prediction, and each arrow points to the same context with prediction enabled. For Gemini and DeepSeek, KE and OAS both shorten episodes, and OAS also raises Success Rate; all arrows but one point down, since prediction lowers Success Rate in fifteen of sixteen model$\times$context pairs (the exception is the Llama 8B with KE only). The y-axis ranges differ per panel.}
\label{fig:sr_steps}
\end{figure}

Table~\ref{tab:main} reports task outcomes and redundant work for all 32 runs under Decoupled Partial Observability, and Figure~\ref{fig:sr_steps} plots each configuration's Success Rate against its episode length; the remaining coordination metrics are in Table~\ref{tab:main_full} (Appendix~\ref{app:partial}). The Llama planners run locally with PARTNR's grammar-constrained decoding \citep{Chang2024PARTNR:}, extended for OANA prediction (Appendix~\ref{sec:appendix_constraint_grammar}); Gemini and DeepSeek are accessed through an API, where constrained decoding is unavailable, but still achieve the highest task performance.
 
\textbf{OAS observation primarily improves task quality; KE primarily improves execution efficiency.} Adding OAS alone to the object-only baseline reduces Extraneous Effort for every planner (Llama 8B 62.3\%$\rightarrow$58.5\%, Llama 70B 39.4\%$\rightarrow$36.5\%, Gemini 44.3\%$\rightarrow$33.0\%, DeepSeek 38.2\%$\rightarrow$28.1\%), raises Action Success for the three stronger planners, and yields the best Success Rate of all configurations for Gemini (81.2\%) and DeepSeek (83.6\%). KE alone leaves Success Rate essentially unchanged for the same three planners (70B 76.0\%$\rightarrow$76.9\%, Gemini 80.4\%$\rightarrow$79.9\%, DeepSeek 82.3\%$\rightarrow$82.2\%) but shortens episodes by 25\% for Gemini (4118$\rightarrow$3081 steps) and 23\% for DeepSeek (3857$\rightarrow$2951), with fewer Planning Cycles and less replanning after the partner has finished (Table~\ref{tab:main_full}). This matches the semantics of the two channels: OAS tells an agent what its partner is currently doing, and therefore what not to duplicate, whereas KE removes redundant search by synchronizing object knowledge on co-location.
 
\textbf{The two channels are largely redundant for strong planners but complementary for the weakest one.} For Gemini and DeepSeek, OAS-only and KE-only each recover most of the attainable efficiency gain, and combining them adds little (Gemini 3193 and 3081$\rightarrow$2911 steps; DeepSeek 3112 and 2951$\rightarrow$2794). For the Llama 8B, KE without OAS is its single worst configuration (Success Rate 60.9\%$\rightarrow$54.3\%, 4914 steps, Extraneous Effort 69.6\%), OAS alone does not help either (58.8\%), and only the combination recovers its best Success Rate (62.3\%) and fewest steps (3781). We interpret this as an attribution failure: synchronized object knowledge without any signal about what the partner is doing with those objects (Section~\ref{sec:info}) can enlarge the world graph the 8B must reason over without giving it a way to assign newly appearing objects to the partner's activity, so it replans more often and duplicates work. The Llama 70B moves in the same direction on Extraneous Effort (39.4\%$\rightarrow$43.0\% with KE alone) without losing Success Rate (76.0\%$\rightarrow$76.9\%; Appendix~\ref{app:partial_task}). Shared context is therefore not uniformly beneficial; it must be interpretable by the planner that receives it.

\textbf{OAS + KE is the most efficient operating point, while peak Success Rate moves to OAS-only for the strongest planners.} Without prediction, OAS + KE gives the fewest Simulation Steps for the Llama 8B (3781), Gemini (2911), and DeepSeek (2794), comes within 2\% of the fastest setting for the Llama 70B (3095 vs.\ 3041), and gives the lowest or near-lowest (within 0.1 points) Extra Planning ratios for every model. For Gemini and DeepSeek, OAS + KE comes within 1.2 points of the OAS-only Success Rate (81.1\% vs.\ 81.2\%; 82.4\% vs.\ 83.6\%), within the uncertainty of a single run (Section~\ref{sec:setup}), so we read the two settings as equivalent in accuracy and separated by how efficiently that accuracy is obtained. When only one channel can be implemented, KE is the cheaper choice and OAS the more accurate one (Appendix~\ref{app:partial_task}).

\textbf{OANA prediction lowers Success Rate and Percent Complete in fifteen of sixteen model$\times$context pairs.} All arrows but one in Figure~\ref{fig:sr_steps} point down. Across these fifteen pairs, the Success Rate drop ranges from $-2.1$ points (Gemini, OAS only) to $-7.1$ points (DeepSeek, no shared context), while the Other-Agent Mention Ratio rises from 10--53\% to 55--100\%: the agents reason explicitly about the partner far more, but that reasoning does not convert into completed task propositions. The behavior is consistent with premature commitment: agents act on an anticipated partner action instead of the observed situation, and a wrong prediction can leave a subtask undone as ``already handled.'' Duplication is not the main cost, since Extraneous Effort falls under prediction in 11 of 16 pairs; prediction makes agents yield to each other more, not collide more. The failures it adds are also ordinary ones: the average completion of a failed episode, $(\mathrm{PC}-\mathrm{SR})/(100-\mathrm{SR})$, stays between 38\% and 53\% in all 32 runs (Appendix~\ref{app:partial_task}). The one exception is the Llama 8B with KE but no OAS, the configuration in which KE alone was most harmful: there, prediction raises Success Rate from 54.3\% to 56.2\% (less than the 2.2-point standard error of a difference between two single runs) and leaves Percent Complete unchanged at 72.9\%, consistent with prediction partly supplying the partner attribution that ungrounded KE lacks (Appendix~\ref{app:partial_task}).
 
\textbf{For the API-accessed planners, the cost of prediction shrinks as shared context grows; for the Llama 70B, only its time cost does.} DeepSeek's Success Rate penalty decreases monotonically with available context ($-7.1$, $-6.6$, $-5.2$, $-2.4$ points for none, KE, OAS, and OAS + KE), as does its step overhead (+1629, +1298, +815, +757); Gemini shows the same ordering more weakly ($-4.4$/$-4.1$ without OAS vs.\ $-2.1$/$-2.5$ with it). For the Llama 70B, prediction costs 592 and 84 additional steps without OAS but saves 406 and 246 once OAS is available, giving its fastest configuration overall (2849 steps). Anticipating a partner whose state is unobserved amounts to guessing from object positions alone; once the partner's state is visible, predictions can build on it, and part of their overhead is recovered. How prediction changes planning volume and the quality of individual actions also differs by planner family (Appendix~\ref{app:partial_task}).
 
\textbf{Partner-mention frequency tracks task success only when it is spontaneous.} Without prediction, the Other-Agent Mention Ratio rises with planner strength across our four planners (robot: 9.8\%, 13.3\%, 30.4\%, 39.1\% for the Llama 8B, Llama 70B, Gemini, and DeepSeek) and orders the models exactly as Success Rate does (60.9\%, 76.0\%, 80.4\%, 82.3\%). Once prediction is instructed, mentions reach 55--92\% for the Llama and Gemini planners and essentially 100\% for DeepSeek, yet Success Rate falls in all but one case, and the OAS + KE configuration without prediction, the most efficient operating point above, has comparatively low mention rates in every model (Llama 8B 21.0/30.9\%, Llama 70B 18.1/24.7\%, Gemini 27.5/36.5\%, DeepSeek 48.6/51.4\%). The metric is informative as an unprompted behavioral signal and uninformative once instructed, so on its own, it cannot separate endogenous partner modeling from prompt-compliant mentioning. Observation-grounded partner reasoning is less frequent but more useful than mandated prediction.

\subsection{Role Asymmetry Between the Agents}
\label{sec:roles}

\begin{figure}[t]
\centering
\includegraphics[width=\textwidth]{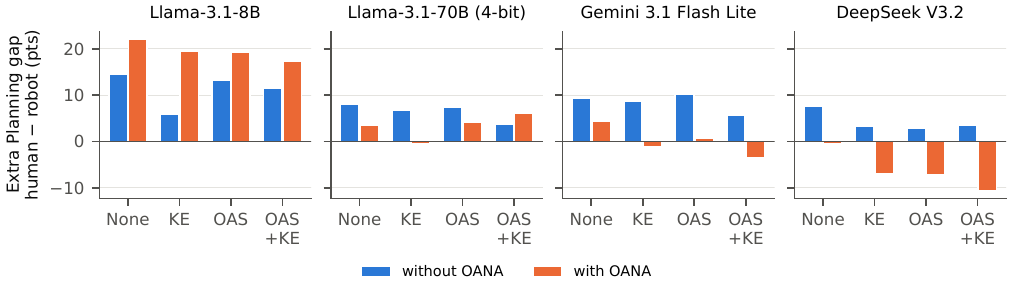}
\caption{\textbf{Completion lag between the agents:} human minus robot Extra Planning Ratio (percentage points) for each shared context, without (blue) and with (orange) OANA prediction. Positive values mean the human keeps replanning after the robot has finished. Without prediction, the human trails in every configuration, and the full shared context (OAS + KE) shrinks the lag in every model; prediction widens it for the Llama 8B and reverses it for DeepSeek, for Gemini in two contexts, and marginally for the Llama 70B with KE only ($-0.5$ points).}
\label{fig:roles}
\end{figure}

The two agents differ in their action spaces, since the human additionally holds the object-state manipulation actions \textit{Clean, Fill, Pour, PowerOn,} and \textit{PowerOff} (Appendix~\ref{sec:appendix_skill_api_library}), and in sensor height, and the prompts inherited from PARTNR make the human the task giver and the robot the task receiver (Appendix~\ref{sec:appendix_prompts}). Both run the same planner, prompt structure, and observability configuration, so systematic differences between them reflect the action spaces, the prompted roles, and the task structure these induce, and possibly sensor height, rather than model capability; our design does not separate these sources.
 
\textbf{The robot acquires objects first, and the human finishes last.} The robot reaches its first object interaction earlier than the human in 28 of 32 runs (mean gap 482, 205, 180, and 18 steps for the Llama 8B, Llama 70B, DeepSeek, and Gemini), since its repertoire consists of navigation and rearrangement primitives that converge on a pickup. Conversely, the human replans more after its partner has finished than the robot does in 25 of 32 runs (Figure~\ref{fig:roles}), although Task Offloading exceeds 50\% in all 32 runs: the robot completes more subtasks by count while going idle first. Together with the prompted roles, a likely explanation is that the human's exclusive state-modification actions sit at the ends of dependency chains, since an object usually must be transported before it is cleaned, filled, or switched on, so they are serialized behind the robot's rearrangements. The division of labor itself barely moves: Task Offloading stays within 52.4--60.7\% across all 32 runs while Success Rate varies from 54.3\% to 83.6\% (Appendix~\ref{app:partial_roles}).
 
\textbf{Shared context helps chiefly by shortening the human's trailing window.} Without prediction, moving from object-only observation to OAS + KE shrinks the completion lag in every model (Llama 8B +14.5$\rightarrow$+11.4, Llama 70B +8.1$\rightarrow$+3.8, Gemini +9.4$\rightarrow$+5.6, DeepSeek +7.7$\rightarrow$+3.5 points), and the human's own Extra Planning falls by more than the robot's in every case. Because episode length is set by the later-finishing agent, this is what turns into the Simulation Step reductions of Section~\ref{sec:task}, and it explains why OAS and KE improve efficiency more reliably than Success Rate.
 
\textbf{Prediction reverses the lag for strong planners and amplifies it for the weakest one.} Under OANA, DeepSeek's robot Extra Planning roughly doubles (20.3\%$\rightarrow$39.0\% at baseline), turning the lag negative in all four contexts; Gemini reverses in two of four, and the Llama 70B marginally in one (KE only, $-0.5$ points); for the Llama 8B, the lag widens instead (+14.5$\rightarrow$+22.1 points at baseline). Since the robot predicts over the human's 14 action types while the human predicts over the robot's 9, prediction imposes an asymmetric burden that strong planners absorb as robot-side deliberation. The human also executes less reliably (robot Action Success is at least as high in 28 of 32 runs), sees its Action Success change less favorably than the robot's when prediction is enabled in 13 of 16 pairs, and mentions the partner more often in all 16 no-prediction runs; which agent replans more overall depends on the planner (Appendix~\ref{app:partial_roles}). Under partial observability, the robot thus acts as a transport layer, and the human as a finisher that determines when the episode ends. For assistive settings in which the human side is a person, this pattern calls for caution: a robot that maximizes completed subtasks can still leave its partner working alone at the end of the task, and partner-state access plus knowledge exchange address that failure more directly than raising the robot's share of the work would. Much of the lag, however, is produced by partial observability itself (Section~\ref{sec:robust}).

\subsection{Partner Prediction}
\label{sec:prediction}

\begin{figure}[t]
\centering
\includegraphics[width=\textwidth]{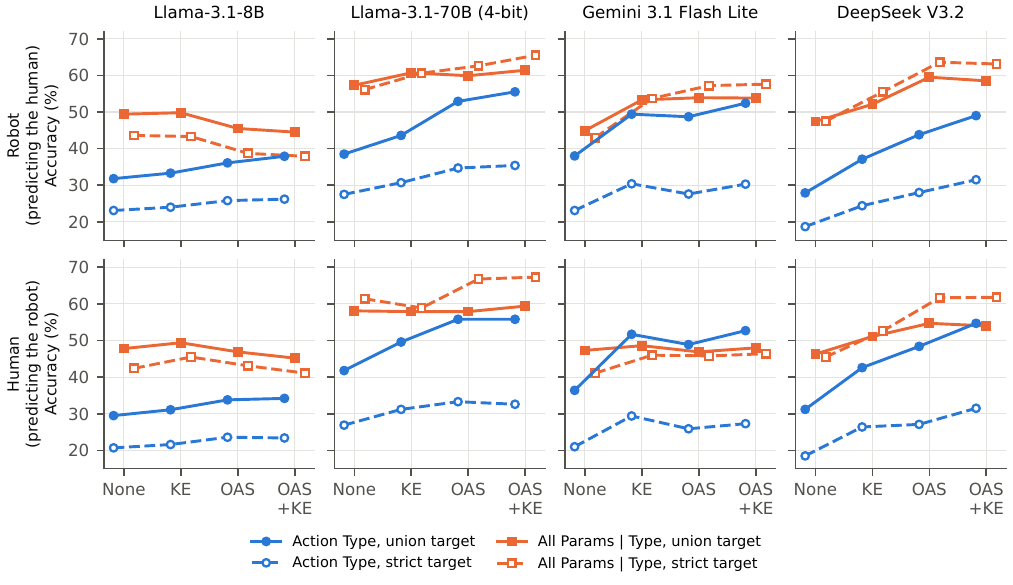}
\caption{\textbf{Partner-prediction accuracy by shared context, for the robot predicting the human (top) and the human predicting the robot (bottom):} Action Type accuracy (blue) and all-parameter accuracy given a correct action type (orange), under the union target (solid, filled markers; Table~\ref{tab:oana_prediction_accuracy}) and the strict next-action target (dashed, hollow markers; Table~\ref{tab:oana_prediction_accuracy_next}). Each point is a table entry; markers are offset horizontally so that close values remain visible. From None to OAS\,+\,KE, Action Type accuracy rises more than the conditional term for every planner and agent (+14.7/+14.6 against +4.8/+1.8 points for robot/human, averaged over planners, union target). The strict target lowers Action Type accuracy by 36.5\% on average (27--48\% across cells), while the conditional term is unchanged on average (52.2\% vs.\ 52.4\%) but planner-dependent: lower in every cell for the Llama 8B and higher in most cells for the Llama 70B and DeepSeek.}
\label{fig:prediction}
\end{figure}

Table~\ref{tab:oana_prediction_accuracy} reports how accurately agents predict each other's next action in the OANA-enabled runs of Table~\ref{tab:main}, scored against the union target (Section~\ref{sec:oana}) at several levels of abstraction: the action type alone (e.g., Pick, Navigate), the principal parameter (e.g., cup in Place[cup, on, table]), all parameters, their combinations, and conditional accuracies scored only over predictions whose action type already matches the target. By construction, $\mathrm{Acc}(\text{type} \wedge \text{params}) = \mathrm{Acc}(\text{type}) \times \mathrm{Acc}(\text{params} \mid \text{type})$, an identity we verified to within rounding in all 64 model$\times$context$\times$agent$\times$granularity cells of Table~\ref{tab:oana_prediction_accuracy}. We use it to assign every change to the coarse term (what the partner will do) or the conditional term (which objects, furniture, or room it will act on). Figure~\ref{fig:prediction} plots both terms for each planner, context, and agent, under both targets of Section~\ref{sec:oana}; Section~\ref{sec:robust} discusses the strict one.
 
\textbf{Shared context tells an agent what its partner will do far more than which object it will act on.} OAS + KE gives the top Action Type accuracy for every model and agent (strictly in seven of eight cases), with gains over the object-only baseline of +17.0/+14.0 points for the Llama 70B, +14.4/+16.3 for Gemini, and +21.1/+23.5 for DeepSeek, although neither channel transmits intentions or plans, only observed states and object locations. The two channels overlap: the sum of the single-channel gains exceeds the combined gain in seven of eight model$\times$agent cases (Gemini realizes about 60\% of the sum, DeepSeek 83\%). Averaged over the four planners, the conditional principal-parameter accuracy improves by only +4.2/+1.0 points, against +14.7/+14.6 for Action Type, and even the highest conditional accuracy in the table (Llama 70B with KE only, 67.6/64.5\%) leaves about a third of correctly typed predictions with the wrong principal parameter. Predicting the arguments remains the main difficulty. For the Llama 8B, richer context even lowers the conditional term (all-parameter accuracy given the type, 49.4\%$\rightarrow$45.5\% with OAS, robot) while raising the type term, so its joint accuracy barely moves (15.7\%$\rightarrow$16.4\%).

\textbf{Whether object knowledge alone suffices depends on the planner.} For Gemini, KE alone matches or exceeds the Action Type gain of OAS alone (+11.4 vs.\ +10.7 for the robot, +15.3 vs.\ +12.5 for the human); for DeepSeek, it recovers about 60\% of it, for the Llama 70B about a third on the robot side and about half on the human side (+5.1 vs.\ +14.4; +7.8 vs.\ +14.0), and for the Llama 8B about a third. A plausible account is self-projection: once both agents hold the same object beliefs, an agent can predict its partner by asking what it would itself do next, a route that the API-accessed planners, Gemini in particular, appear to exploit more than the Llama planners, which rely mainly on reading off the partner's state. Constrained decoding cannot explain this difference in action type, since with KE only the unconstrained Gemini still predicts the partner's action type better than the constrained Llama 70B (49.4/51.7\% vs.\ 43.6/49.6\%), although it may contribute to the Llama 70B's higher conditional parameter accuracy (Appendix~\ref{app:partial_pred}).
 
\textbf{Task-solving strength does not predict partner-modeling strength.} DeepSeek achieves the best Success Rate of all four planners, yet at the object-only baseline its robot is the weakest partner predictor in the table (27.9\% Action Type, below the Llama 8B's 31.8\%), whereas the third-best task solver, the 4-bit Llama 70B, is the best predictor under full shared context (55.5/55.8\%). Strong task performance thus comes mainly from individual planning, and partner-mention frequency is again no guide, since DeepSeek mentions its partner in essentially every replanning step under OANA. Within the API-accessed planners, however, accuracy does track the cost of prediction: as DeepSeek's robot Action Type accuracy rises through 27.9\%, 37.1\%, 43.8\%, and 49.0\% for none, KE, OAS, and OAS + KE, its Success Rate penalty falls through $-7.1$, $-6.6$, $-5.2$, and $-2.4$ points, consistent with agents acting on their predictions. Gemini follows this only loosely and the Llama planners not at all, so we present it as an account of the API-accessed models rather than a general law.
 
\textbf{The two agents predict different parts of the action.} The human predicts over the robot's 9 action types, and the robot over the human's 14. Under the union target, the human predicts the action type more accurately for the Llama 70B and DeepSeek in all four contexts and for Gemini in three, while the robot predicts the parameters more accurately once the type is right, a lead that appears only with shared context ($-3.1$ to $-4.2$ points H$-$R for the Llama 70B, $-5.7$ to $-7.1$ for Gemini, $-2.0$ to $-4.9$ for DeepSeek). The latter plausibly reflects arity: the robot's Place and Rearrange take five argument slots, whereas the five human-only actions take one. The two asymmetries cancel in the joint metric, whose H$-$R gap stays within $\pm$3.3 points in all 16 cells, so joint accuracies alone would hide both. The Llama 8B inverts the type asymmetry, and neither asymmetry is robust to the strict target (Section~\ref{sec:robust}; details in Appendices~\ref{app:partial_pred} and~\ref{subsubsec:appendix_action_pred_eval_next}).

\begin{table*}[t]
\centering
\setlength{\tabcolsep}{1.6pt}
\renewcommand{\arraystretch}{1.6}
\scriptsize
\caption{\textbf{Other Agent's Next Action (OANA) prediction accuracy (\%)} across action abstraction levels for robot (R) and human (H) agents. The target action is the \textbf{union of the current and next actions} of the other agent. Results highlight that Other Agent's State (OAS) observation and object Knowledge Exchange (KE) in the same room both improve prediction, that their combination gives the best action-type accuracy for every model and agent, and that the gains are concentrated in the action type rather than in its parameters. Per-model bests are underlined; overall bests are bolded.}
\label{tab:oana_prediction_accuracy}
\begin{tabularx}{\textwidth}{
>{\centering\arraybackslash}p{0.9cm} 
>{\centering\arraybackslash}p{0.6cm} 
>{\centering\arraybackslash}p{0.6cm} 
>{\centering\arraybackslash}p{1.6cm} 
>{\centering\arraybackslash}p{1.9cm} 
>{\centering\arraybackslash}p{2.0cm} 
>{\centering\arraybackslash}p{1.9cm} 
>{\centering\arraybackslash}p{1.9cm} 
>{\centering\arraybackslash}p{1.9cm} 
>{\centering\arraybackslash}p{2.0cm} 
}
\textbf{Model} & \textbf{OAS} & \textbf{KE} 
& \textbf{Action Type (R/H) $\uparrow$} 
& \textbf{Principal Par. (R/H) $\uparrow$} 
& \textbf{Action Type+Princ. Par. (R/H) $\uparrow$} 
& \textbf{All Params (R/H) $\uparrow$} 
& \textbf{Action Type +All Params (R/H) $\uparrow$} 
& \textbf{Principal Par. if Type (R/H) $\uparrow$} 
& \textbf{All Params if Type (R/H) $\uparrow$} \\
\hline
\multirow{4}{0.9cm}{\centering Llama-3.1-8B-Instruct}
& \xmark & \xmark 
& 31.8 / 29.5 & 24.6 / 21.7 & 17.2 / 15.8 & 17.8 / 15.4 & 15.7 / 14.1 & 54.2 / 53.5 & 49.4 / 47.8 \\
& \xmark & \cmark 
& 33.3 / 31.1 & \underline{26.1} / \underline{24.2} & 18.4 / 17.8 & \underline{18.8} / \underline{17.2} & 16.6 / 15.4 & \underline{55.3} / \underline{57.2} & \underline{49.8} / \underline{49.4} \\
& \cmark & \xmark 
& 36.1 / 33.8 & 22.5 / 21.5 & 17.8 / \underline{17.9} & 18.1 / 17.0 & 16.4 / \underline{15.8} & 49.2 / 53.1 & 45.5 / 46.9 \\
& \cmark & \cmark 
& \underline{37.9} / \underline{34.2} & 23.2 / 21.6 & \underline{18.9} / 17.2 & 18.5 / 16.8 & \underline{16.9} / 15.5 & 49.9 / 50.4 & 44.5 / 45.2 \\
\hline
\multirow{4}{0.9cm}{\centering Llama-3.1-70B-Instruct 4-bit}
& \xmark & \xmark 
& 38.5 / 41.8 & 34.2 / 37.7 & 24.1 / 26.5 & 26.7 / 30.3 & 22.1 / 24.3 & 62.5 / 63.3 & 57.3 / 58.1 \\
& \xmark & \cmark 
& 43.6 / 49.6 & 39.1 / 41.2 & 29.5 / 32.0 & 30.8 / 34.0 & 26.5 / 28.7 & \textbf{\underline{67.6}} / \textbf{\underline{64.5}} & 60.7 / 57.9 \\
& \cmark & \xmark 
& 52.9 / \textbf{\underline{55.8}} & 42.4 / 41.8 & 35.2 / 34.7 & 36.1 / 36.8 & 31.7 / 32.3 & 66.4 / 62.2 & 59.9 / 57.9 \\
& \cmark & \cmark 
& \textbf{\underline{55.5}} / \textbf{\underline{55.8}} & \textbf{\underline{45.2}} / \textbf{\underline{43.6}} & \textbf{\underline{37.3}} / \textbf{\underline{35.8}} & \textbf{\underline{38.5}} / \textbf{\underline{38.2}} & \textbf{\underline{34.1}} / \textbf{\underline{33.2}} & 67.2 / 64.1 & \textbf{\underline{61.4}} / \textbf{\underline{59.4}} \\
\hline
\multirow{4}{0.9cm}{\centering Gemini 3.1 Flash Lite}
& \xmark & \xmark 
& 38.0 / 36.4 & 28.0 / 26.4 & 19.7 / 19.1 & 21.9 / 21.0 & 17.1 / 17.2 & 51.7 / \underline{52.6} & 44.9 / 47.3 \\
& \xmark & \cmark 
& 49.4 / 51.7 & 38.5 / 36.2 & 28.8 / \underline{27.2} & 32.7 / 30.9 & 26.3 / 25.1 & \underline{58.3} / \underline{52.6} & 53.3 / \underline{48.6} \\
& \cmark & \xmark 
& 48.7 / 48.9 & 39.0 / 36.4 & 28.1 / 24.7 & 32.6 / 29.0 & 26.3 / 23.0 & 57.7 / 50.6 & \underline{53.9} / 46.9 \\
& \cmark & \cmark 
& \underline{52.4} / \underline{52.7} & \underline{42.7} / \underline{39.6} & \underline{30.4} / 27.1 & \underline{35.4} / \underline{31.8} & \underline{28.1} / \underline{25.3} & 58.1 / 51.5 & 53.8 / 48.0 \\
\hline
\multirow{4}{0.9cm}{\centering DeepSeek V3.2}
& \xmark & \xmark 
& 27.9 / 31.2 & 21.3 / 23.9 & 14.4 / 15.6 & 16.8 / 18.6 & 13.2 / 14.4 & 51.7 / 50.2 & 47.4 / 46.3 \\
& \xmark & \cmark 
& 37.1 / 42.6 & 28.6 / 33.8 & 21.0 / 23.2 & 23.2 / 27.0 & 19.3 / 21.8 & 56.6 / 54.6 & 52.1 / 51.1 \\
& \cmark & \xmark 
& 43.8 / 48.4 & 34.2 / 37.2 & 27.2 / 27.7 & 29.7 / 31.6 & 26.0 / 26.5 & \underline{62.2} / 57.3 & \underline{59.5} / \underline{54.7} \\
& \cmark & \cmark 
& \underline{49.0} / \underline{54.7} & \underline{38.4} / \underline{40.6} & \underline{30.2} / \underline{31.4} & \underline{32.9} / \underline{34.1} & \underline{28.7} / \underline{29.6} & 61.6 / \underline{57.4} & 58.5 / 54.0 \\
\hline
\end{tabularx}
\end{table*}

\subsection{Robustness: Strict Target and Full Observability}
\label{sec:robust}
 
We test the main findings under two stricter conditions (details in Appendices~\ref{sec:appendix_task_solving_analysis} and~\ref{sec:appendix_action_pred_eval}). The \emph{strict target} (Section~\ref{sec:oana}) re-scores the same runs against the partner's next action only, so naming an ongoing action no longer counts (Table~\ref{tab:oana_prediction_accuracy_next}). \emph{Full observability} gives both agents the complete ground-truth world graph at all times, which removes perceptual uncertainty, makes KE inapplicable, and makes OAS available continuously rather than only on co-location; we run it with the two Llama planners (Tables~\ref{tab:task_solving_full_obs}, \ref{tab:oana_prediction_accuracy_union} and \ref{tab:oana_prediction_accuracy_full}).
 
\textbf{Part of what the union target credits is recognition rather than anticipation.} The strict target lowers Action Type accuracy by 36.5\% on average while leaving the all-parameter conditional term essentially unchanged on average (Figure~\ref{fig:prediction}), and the gain from enabling OAS roughly halves (Llama 70B: +14.4/+14.0$\rightarrow$+7.2/+6.4 points), because an observation such as \emph{Walking in hallway\_1} lets an agent name the ongoing Navigate action without anticipating anything. Under full observability, enabling OAS raises the Llama 70B's union-target Action Type accuracy by +8.6/+6.6 points but leaves its strict accuracy unchanged or lower ($-0.3$/$-2.8$, Tables \ref{tab:oana_prediction_accuracy_union} and \ref{tab:oana_prediction_accuracy_full}). The agent-level asymmetries of Section~\ref{sec:prediction} also depend on the scoring: under the strict target, the robot matches or leads the human on action type in 14 of 16 cells, and its conditional parameter lead survives clearly only for Gemini.
 
\textbf{Anticipation stays weak even with complete world knowledge, and prediction still costs success.} The best strict-target Action Type accuracy is 35.4/33.3\% under partial observability (Llama 70B) and 43.8/45.9\% under full observability without OAS, so even an agent that knows the entire world names its partner's next action type wrongly more than half the time. OANA prediction still lowers Success Rate and Percent Complete in all four full-observability configurations (nineteen of twenty pairs across both regimes, Table \ref{tab:task_solving_full_obs}), so prediction does not hurt merely because agents anticipate partners they cannot see. Its cost depends on capacity: under full observability, the Llama 70B's penalty shrinks to $-1.6$ and $-0.6$ points, whereas the Llama 8B's penalties ($-6.2$, $-2.3$) match its partial-observability penalties ($-5.9$, $-2.2$).
 
\textbf{Full observability mainly shortens episodes and removes much of the role asymmetry.} Complete object knowledge shortens episodes by 43\% at the object-only baseline for both planners (Table~\ref{tab:compare_dp_full_obs_without_oana}) while leaving the Llama 70B's Success Rate essentially unchanged (76.9\% against 77.5\% at its best partial-observability setting), and about 94\% (Llama 8B) and 86\% (Llama 70B) of Extraneous Effort remains: most redundant work is a coordination failure that object knowledge does not fix. OAS keeps its coordination benefits, lowering Extraneous Effort and Extra Planning, but loses its task-quality benefit, which under partial observability partly compensated for missing world knowledge. The human's completion lag halves for the Llama 70B (+8.1$\rightarrow$+3.9 points) and reverses for the Llama 8B (+14.5$\rightarrow$$-1.8$), so much of it was an exploration artifact of partial observability, whereas the robot's execution advantage persists in all eight full-observability runs.

\section{Limitations}

Our experiments run in the Habitat 3.0 simulator on the PARTNR validation set with privileged perception and oracle skills, which may limit generalization to real deployments with perception error, actuation noise, and imperfect state estimation. We report a single run per configuration (Section~\ref{sec:setup}), so small differences, such as the Success Rate gap between OAS-only and OAS + KE for the strongest planners, should not be over-interpreted. The Llama 70B planner is 4-bit quantized, so its results may not transfer to the full-precision model. Exact reproducibility of the API-accessed planners is outside our control: the providers determine which model version is served under a given identifier, how requests are routed and executed, and how long a model remains available, so repeated queries with identical prompts and settings may not return identical outputs over time (Appendix \ref{sec:appendix_implementation_details}). The API-accessed planners cannot use PARTNR's grammar-constrained decoding (Appendix~\ref{app:constraint}), so prediction accuracy is not perfectly comparable across model families, although the context-dependent pattern of the differences argues against decoding alone explaining them. The study covers four planners and two-agent teams, and the full-observability control covers only the two Llama planners. The simulated human is itself an LLM agent, so the role asymmetries we report describe an LLM--LLM pair rather than an LLM--person pair. Because the PARTNR role prompts make the human the task giver and the robot the task receiver, these asymmetries are also not attributable to the action spaces alone. Several interpretations, notably the attribution-failure reading of ungrounded KE and the arity-based account of the parameter asymmetry, are consistent with the data but are not tested directly by our design. Next-action prediction is a behavioral proxy for partner modeling: we do not probe beliefs, false beliefs, or other mental-state attributions, so our results bear on the behavioral consequences usually attributed to ToM rather than on ToM itself. Finally, several design choices bound the interpretation: (i) OAS observation also merges the partner's current percepts on co-location (Section~\ref{sec:info}), so part of its effect may come from shared perception rather than partner state; (ii) enabling OANA adds the partner's action descriptions to the prompt as well as the prediction instruction; (iii) predictions are scored at the class level after masking instance IDs (Section~\ref{sec:oana}); and (iv) both OAS observation and OANA prediction lengthen the prompt, so their effects are confounded with added context length and may scale differently on longer-horizon or more constrained tasks.
\section{Conclusions}

We presented a controlled study of coordination between two embodied LLM agents in PARTNR, introducing Decoupled Partial Observability and manipulating Other Agent's State (OAS) observation, object Knowledge Exchange (KE), and Other Agent's Next Action (OANA) prediction independently across a full factorial grid and four planners, with coordination metrics reported per agent as well as for the team. The two context channels act on different axes: OAS mainly improves task quality and KE mainly improves efficiency, and they are largely redundant for strong planners but complementary for the weakest, for which object knowledge without partner state is the worst configuration we measure. Shared context thus helps only a planner that can make use of it, and, if our attribution reading is right, tagging shared observations with their source is a cheap and concrete design choice for real systems, since KE carries no sender identity. OANA prediction lowers the Success Rate in nineteen of the twenty model$\times$context pairs we ran; its accuracy gains concentrate in the coarse action type, and part of that accuracy reflects recognizing the partner's ongoing action. Even with complete world knowledge, the Llama planners name their partner's next action type wrongly more than half the time. The two agents occupy distinguishable roles under Decoupled Partial Observability: the robot usually finishes its share first, while the simulated human keeps working afterwards and determines when the episode ends, and shared context improves throughput chiefly by shortening this final, human-only period. Embodied LLM agents therefore coordinate better when partner state is supplied in a structured way, but our evidence indicates that they do so mainly by reacting to that state rather than by anticipating the partner, and the partner modeling that does occur is shallow and planner-dependent, with agent-level asymmetries that appear only under lenient scoring. For claims about Theory of Mind in LLMs, this argues for behavioral tests that separate recognition from anticipation and anticipation from its use. Three methodological cautions follow: (i) in our data, partner-mention frequency tracks task success only when it is spontaneous; (ii) prediction accuracy that credits a partner's ongoing action conflates recognition with anticipation; and (iii) team-averaged metrics conceal role asymmetries that per-agent reporting makes visible.
\subsubsection*{Broader Impact Statement}

This work studies coordination between two LLM agents in simulation; no human participants or personal data are involved, and the simulated ``human'' is itself an LLM agent. Its most direct application is assistive household robotics, where our results suggest concrete design choices: giving a robot its partner's observable state and shared object knowledge improves coordination, whereas instructing it to predict its partner's next action lowered task success in nearly all settings. Two risks follow. First, a robot acting on an incorrect prediction of what a person will do may leave tasks undone or get in the person's way; deployed systems should therefore treat partner predictions as hypotheses to be confirmed by observation before acting, and should be evaluated with per-agent metrics, since our role analysis shows that a robot that completes many subtasks can still leave the person the last and hardest part of the work. Second, partner-state observation and object knowledge exchange involve monitoring people and their belongings at home. This can be limited by restricting observation to coarse, task-relevant states such as those we use (e.g., \emph{Walking in hallway\_1}), processing them on the device without storing them, gating them by co-location as in our design, and making them visible to and switchable by the user.

\bibliography{main}
\bibliographystyle{tmlr}

\appendix
\section{Appendix}
\label{sec:appendix}



This appendix supplements the main paper. Appendix~\ref{sec:appendix_implementation_details} details the agents' action space, its generation, and the implementation; Appendix~\ref{sec:appendix_action_comparison} explains the action comparison protocol; Appendix~\ref{app:partial} gives the complete metrics and extended analyses behind Sections~\ref{sec:task}--\ref{sec:prediction} under Decoupled Partial Observability; Appendix~\ref{sec:appendix_task_solving_analysis} analyzes task performance and coordination under full observability; Appendix~\ref{sec:appendix_action_pred_eval} evaluates partner prediction under the strict next-action target and under full observability; and Appendix~\ref{sec:appendix_prompts} shows the prompts used in the experiments.

\subsection{Implementation details for the ReAct agents}\label{app:impl}
\label{sec:appendix_implementation_details}

Most of our experiments with the Llama models were conducted locally on four NVIDIA A100 GPUs, each with 64 GB of VRAM, while a smaller portion used three NVIDIA A40 GPUs with 48 GB of VRAM each. For the 4-bit quantized Llama-3.1-70B-Instruct planner, we were able to run 2 and 1 parallel processes on these respective setups, whereas for the Llama-3.1-8B-Instruct planner, we used 5 and 3 parallel processes to complete the 1,000 validation episodes via Hugging Face Transformers inference. We used Gemini 3.1 Flash Lite and DeepSeek V3.2 through the OpenRouter API (model identifiers: google/gemini-3.1-flash-lite, deepseek/deepseek-v3.2). Some later runs used vLLM instead of Hugging Face Transformers for inference, with the same model weights and decoding settings.

Due to limited hardware and CPU hour availability, some experiments had to be split across these two types of resources, which operate at different speeds, so the runtimes below are not directly comparable across configurations. 
According to PARTNR’s \textit{runtime} metric, the fastest configuration averaged 299 seconds per episode (DeepSeek V3.2 baseline experiment without OANA prediction), while the slowest required 1008 seconds per episode on average (Llama-3.1-70B-Instruct, Decoupled Partial Observability without OAS and with OANA prediction). We also observed that with OANA prediction, episodes generally required longer \textit{runtime} (9\% to 60\% more), partly due to additional logging and metric computation and partly because more tokens are generated. In the case of OANA prediction, more tokens are generated at the replannings (with a 250 limit) because of the reasoning about both the other agent’s next action and the agent’s own action, as well as predicting OANA itself. Furthermore, the instruction prompt becomes longer and more complex, and incorporating OAS observations adds even more input tokens.

Despite this increase in runtime, Table \ref{tab:main_full} shows that OANA prediction often reduces the number of planning cycles and, in some cases, also decreases the number of simulation steps.

\subsubsection{Skill API Library}\label{app:skills}
\label{sec:appendix_skill_api_library}

Below is the set of API libraries of action skill tools shared across all experiments that we use from PARTNR \citep{Chang2024PARTNR:}, which enable the agents to take actions in the simulator. The robot agent has no access to the state-changing tools (Clean, Fill, Pour, PowerOff, PowerOn).
\begin{itemize}
\item \textbf{Clean}: Used for cleaning an object.
\item \textbf{Close}: Used for closing an articulated entity.
\item \textbf{Explore}: Search a specific room by visiting various receptacles or furniture in that room.
\item \textbf{Fill}: Used for filling an object.
\item \textbf{Navigate}: Used for navigating to an entity.
\item \textbf{Open}: Used for opening an articulated entity.
\item \textbf{Pick}: Used for picking up an object; the agent cannot hold more than one object at a time.
\item \textbf{Place}: Used for placing an object on a target location; requires object name, furniture name, and spatial relation (“on” or “within”). The object must already be held. Optionally supports placement \texttt{next\_to} a reference object (which must already be on the target furniture). API: \texttt{Place[<object>, <relation>, <furniture>, <constraint>, <reference>]}, with \texttt{None} when unused.
\item \textbf{Pour}: Used for pouring from one container to another; pours from the container currently held by the agent.
\item \textbf{PowerOff}: Used for turning off a powered object.
\item \textbf{PowerOn}: Used for turning on a powered object.
\item \textbf{Rearrange}: Used for moving an object to a target location; requires object name, furniture name, and spatial relation (“on” or “within”), and automatically performs pick and place. Optionally supports \texttt{next\_to} with a reference object (must already be on target furniture). API: \texttt{Rearrange[<object>, <relation>, <furniture>, <constraint>, <reference>]}, with \texttt{None} when unused.
\item \textbf{Wait}: Used to make the agent stay idle for some time.
\item \textbf{Done}: Used to indicate that the agent has finished the task.
\end{itemize}

\subsubsection{Constraint Generation}\label{app:constraint}
\label{sec:appendix_constraint_grammar}

We adopt PARTNR’s constrained generation approach, following \cite{geng-etal-2023-grammar}, which restricts token sampling to sequences consistent with the specified grammar. For each Llama LLM call, we construct a grammar that permits only valid tool calls over observed entities. The base grammar used in all experiments is shown below, while the rules for objects, furniture, and rooms are dynamically updated based on the agent’s current world graph. This ensures that skills are invoked only for entities known to the agent. In the construction, the human agent's grammar includes all the actions, but the robot agent's grammar does not contain the object-state modification actions.

\begin{promptbox}{Tool Call Grammar}
root ::= Navigate | Pick | Place | Open | Close | Rearrange | PowerOn | PowerOff | Clean | Fill | Pour | Explore | Wait | Done

Navigate ::= "Navigate[" nav\_target "]" \\
Pick ::= "Pick[" object "]" \\
Place ::= "Place[" object ", " WS spatial\_relation ", " WS furniture ", " WS ((spatial\_constraint ", " WS obj\_or\_furniture) | (("none" | "None") WS ", " WS ("none" | "None"))) "]" \\
Open ::= "Open[" furniture "]" \\
Close ::= "Close[" furniture "]" \\
Rearrange ::= "Rearrange[" object ", " WS spatial\_relation ", " WS furniture ", " WS ((spatial\_constraint ", " WS obj\_or\_furniture) | (("none" | "None") WS ", " WS ("none" | "None"))) "]" \\
PowerOn ::= "PowerOn[" obj\_or\_furniture "]" \\
PowerOff ::= "PowerOff[" obj\_or\_furniture "]" \\
Clean ::= "Clean[" obj\_or\_furniture "]" \\
Fill ::= "Fill[" object "]" \\
Pour ::= "Pour[" object "]" \\
Explore ::= "Explore[" room "]" \\
Wait ::= "Wait[]" \\
Done ::= "Done[]" \\
nav\_target ::= (furniture | room | object) \\
object ::= "object\_1" | "object\_2" | ... \\
obj\_or\_furniture ::= (furniture | object) \\
furniture ::= "furniture\_1" | "furniture\_2" | ... \\
room ::= "room\_1" | "room\_2" | ... \\
spatial\_constraint ::= "next\_to" \\
spatial\_relation ::= "on" | "within" \\
free\_text ::= ['.:,!a-zA-Z\_0-9 ]* \\
WS ::= [ ]*
\end{promptbox}

\textbf{For OANA prediction,} the same rules apply, except that agents use only object classes rather than object classes with object instance IDs (see Section \ref{sec:oana}). Accordingly, the \textit{object} rule changed to \textit{object\_class ::= "object\_class\_1" | "object\_class\_2" | ...}, and all occurrences of \textit{object} in other rules are replaced with \textit{object\_class} as well. The set of object classes is available from the start of each episode based on the simulation information; thus, predictions can be made at any replanning containing the object classes, whereas specific object instances (with IDs) are only known once they are explored and observed.

Note: This type of Constraint Generation could only be applied to the locally run Llama models, where the output tensors were available, whereas for the models accessed via API, this type of Constraint Generation is not applicable. Consequently, these models (Gemini 3.1 Flash Lite and DeepSeek V3.2) can hallucinate entities as action parameters that are not in the world graph of the agents.

\subsection{Action comparison}\label{app:comparison}
\label{sec:appendix_action_comparison}

As stated in Section \ref{sec:oana} and shown in Table~\ref {tab:oana_prediction_accuracy}, we evaluated the OANA prediction across multiple levels of abstraction. The three examples below illustrate the comparison levels. Note that for the comparison, we ignore the instance IDs of the entities.

\textbf{Example 1: }

OANA prediction: Place[cup,on,table,None,None]

Target action: Place[cup,on,table,next\_to,plate]

\textbf{Example 2:}

OANA prediction: Navigate[book]

Target action: Pick[book]

\textbf{Example 3:}

OANA prediction: Explore[bedroom]

Target action: Explore[livingroom]

\textbf{Action comparison levels:}

\begin{itemize}
\item Action Type \\
Example 1 - match: Place $\leftrightarrow$ Place \\
Example 2 - no match: Navigate $\leftrightarrow$ Pick \\ 
Example 3 - match: Explore $\leftrightarrow$ Explore \\ 
\item Principal Parameter \\
Example 1 - match: cup $\leftrightarrow$ cup \\
Example 2 - match: book $\leftrightarrow$ book \\ 
Example 3 - no match: bedroom $\leftrightarrow$ livingroom \\ 
\item All Parameters \\
Example 1 - no match: (cup,on,table,None,None) $\leftrightarrow$ (cup,on,table,next\_to,plate) \\
Example 2 - match: (book) $\leftrightarrow$ (book) \\
Example 3 - no match: (bedroom) $\leftrightarrow$ (livingroom) \\ 
\item Action Type + Principal Parameter \\
Example 1 - match: (Place,cup) $\leftrightarrow$ (Place,cup) \\
Example 2 - no match: (Navigate,book) $\leftrightarrow$ (Pick,book) \\
Example 3 - no match: (Explore,bedroom) $\leftrightarrow$ (Explore,livingroom) \\ 
\item Action Type + All Parameters \\
Example 1 - no match: \\ 
(Place,cup,on,table,None,None) $\leftrightarrow$ (Place,cup,on,table,next\_to,plate) \\
Example 2 - no match: (Navigate,book) $\leftrightarrow$ (Pick,book) \\
Example 3 - no match: (Explore,bedroom) $\leftrightarrow$ (Explore,livingroom) \\ 
\item Principal Parameter if Action Type is matched \\
Example 1 - match: (cup) $\leftrightarrow$ (cup) \\
Example 2 - no count: no action type match\\ 
Example 3 - no match: (bedroom) $\leftrightarrow$ (livingroom) \\ 
\item All Parameters if Action Type is matched \\
Example 1 - no match: (cup,on,table,None,None) $\leftrightarrow$ (cup,on,table,next\_to,plate) \\
Example 2 - no count: no action type match\\
Example 3 - no match: (bedroom) $\leftrightarrow$ (livingroom) \\ 
\end{itemize}
\subsection{Additional Results under Decoupled Partial Observability}
\label{app:partial}
 
This section gives the complete metrics behind Sections~\ref{sec:task}--\ref{sec:prediction} and the analyses that the main text summarizes.
 
\begin{table*}[t]
\centering
\setlength{\tabcolsep}{1.6pt}
\renewcommand{\arraystretch}{1.6}
\scriptsize
\caption{\textbf{Task performance and coordination metrics under Decoupled Partial Observability (full version of Table~\ref{tab:main})} with different Other Agent's State (OAS) observation, Knowledge Exchange (KE) in the same room, and Other Agent's Next Action (OANA) prediction settings, using four different LLM planners for both agents. Results show that OAS observation mainly improves task quality while KE mainly improves execution efficiency, that the best coordination is reached with both enabled and without OANA prediction, and that OANA prediction raises partner mentioning but lowers Success Rate in all but one configuration. SR: Success Rate, PC: Percent Complete, OA Mention: Other-Agent Mention Ratio, R/H: Robot/Human agent. Subgroup bests are underlined; overall bests are bolded for each model.}
\label{tab:main_full}

\begin{subtable}{\textwidth}
\centering
\caption{Llama-3.1-8B-Instruct}
\label{tab:main_8B}
\begin{tabularx}{\textwidth}{
>{\centering\arraybackslash}p{0.6cm} 
>{\centering\arraybackslash}p{0.5cm} 
>{\centering\arraybackslash}p{0.8cm} 
>{\centering\arraybackslash}p{1.0cm} 
>{\centering\arraybackslash}p{0.55cm} 
>{\centering\arraybackslash}p{0.55cm} 
>{\centering\arraybackslash}p{1.8cm} 
>{\centering\arraybackslash}p{1.7cm} 
>{\centering\arraybackslash}p{1.8cm} 
>{\centering\arraybackslash}p{0.8cm} 
>{\centering\arraybackslash}p{1.3cm} 
>{\centering\arraybackslash}p{1.6cm} 
>{\centering\arraybackslash}p{2.1cm} 
}
\textbf{OAS} & \textbf{KE} & \textbf{OANA} & \textbf{Sim Steps $\downarrow$} & \textbf{SR \%$\uparrow$} & \textbf{PC \%$\uparrow$} & \textbf{Plan. Cycles (R / H) $\downarrow$} & \textbf{OA Mention (R / H) \%$\uparrow$} & \textbf{Exploration Eff. (R/H) $\downarrow$} & \textbf{Offload \%$\uparrow$} & \textbf{Extra Effo. \%$\downarrow$} & \textbf{Extra Plan. (R / H) \%$\downarrow$} & \textbf{Action Success (R / H) \%$\uparrow$} \\
\hline
\xmark & \xmark & \xmark & \underline{4214} & \underline{60.9} & \textbf{\underline{77.1}} & 24.2 / 23.2 & 9.8 / 14.6 & \underline{1348} / \underline{1795} & 57.0 & 62.3 & 23.2 / \underline{37.7} & 70.2 / \underline{69.4} \\
\xmark & \xmark & \cmark & 4243 & 55.0 & 73.0 & \underline{21.2} / \underline{20.9} & \underline{85.5} / \underline{88.7} & 1590 / 2244 & \textbf{\underline{60.7}} & \underline{62.0} & \underline{17.7} / 39.8 & \underline{70.9} / 66.0 \\
\hline
\xmark & \cmark & \xmark & 4914 & 54.3 & \underline{72.9} & 27.7 / 25.1 & 9.5 / 17.3 & \textbf{\underline{1300}} / \textbf{\underline{1741}} & 56.9 & 69.6 & 26.5 / \underline{32.3} & 64.5 / 62.4 \\
\xmark & \cmark & \cmark & \underline{4121} & \underline{56.2} & \underline{72.9} & \underline{20.7} / \underline{20.4} & \textbf{\underline{87.7}} / \textbf{\underline{92.1}} & 1487 / 2047 & \underline{58.9} & \underline{62.7} & \textbf{\underline{16.1}} / 35.5 & \underline{71.6} / \underline{66.0} \\
\hline
\cmark & \xmark & \xmark & 4022 & \underline{58.8} & \underline{74.7} & 22.1 / 21.1 & 20.5 / 32.7 & 1793 / \underline{2166} & 55.1 & \textbf{\underline{58.5}} & 20.5 / \underline{33.8} & 72.0 / 66.4 \\
\cmark & \xmark & \cmark & \underline{4017} & 56.6 & 74.2 & \underline{19.8} / \underline{19.4} & \underline{74.7} / \underline{81.2} & \underline{1670} / 2200 & \underline{56.9} & \textbf{\underline{58.5}} & \underline{20.0} / 39.3 & \textbf{\underline{73.6}} / \underline{69.6} \\
\hline
\cmark & \cmark & \xmark & \textbf{\underline{3781}} & \textbf{\underline{62.3}} & \underline{76.8} & 21.5 / 19.8 & 21.0 / 30.9 & 1779 / 2148 & 55.5 & \textbf{\underline{58.5}} & 20.6 / \textbf{\underline{32.0}} & 72.0 / 68.3 \\
\cmark & \cmark & \cmark & 3890 & 57.2 & 74.5 & \textbf{\underline{19.2}} / \textbf{\underline{18.7}} & \underline{75.6} / \underline{80.8} & \underline{1653} / \underline{2138} & \underline{56.4} & 59.1 & \underline{18.3} / 35.6 & \underline{73.4} / \textbf{\underline{70.1}} \\
\hline
\end{tabularx}
\end{subtable}

\vspace{0.6em}

\begin{subtable}{\textwidth}
\centering
\caption{4-bit quantized Llama-3.1-70B-Instruct}
\label{tab:main_70B}
\begin{tabularx}{\textwidth}{
>{\centering\arraybackslash}p{0.6cm} 
>{\centering\arraybackslash}p{0.5cm} 
>{\centering\arraybackslash}p{0.8cm} 
>{\centering\arraybackslash}p{1.0cm} 
>{\centering\arraybackslash}p{0.55cm} 
>{\centering\arraybackslash}p{0.55cm} 
>{\centering\arraybackslash}p{1.8cm} 
>{\centering\arraybackslash}p{1.7cm} 
>{\centering\arraybackslash}p{1.8cm} 
>{\centering\arraybackslash}p{0.8cm} 
>{\centering\arraybackslash}p{1.3cm} 
>{\centering\arraybackslash}p{1.6cm} 
>{\centering\arraybackslash}p{2.1cm} 
}
\textbf{OAS} & \textbf{KE} & \textbf{OANA} & \textbf{Sim Steps $\downarrow$} & \textbf{SR \%$\uparrow$} & \textbf{PC \%$\uparrow$} & \textbf{Plan. Cycles (R / H) $\downarrow$} & \textbf{OA Mention (R / H) \%$\uparrow$} & \textbf{Exploration Eff. (R/H) $\downarrow$} & \textbf{Offload \%$\uparrow$} & \textbf{Extra Effo. \%$\downarrow$} & \textbf{Extra Plan. (R / H) \%$\downarrow$} & \textbf{Action Success (R / H) \%$\uparrow$} \\
\hline
\xmark & \xmark & \xmark & \underline{3056} & \underline{76.0} & \underline{86.9} & \underline{15.7} / \underline{15.1} & 13.3 / 17.2 & \textbf{\underline{909}} / \textbf{\underline{1100}} & \underline{52.4} & \underline{39.4} & \underline{17.3} / \underline{25.4} & \underline{79.9} / \underline{80.0} \\
\xmark & \xmark & \cmark & 3648 & 72.9 & 85.6 & 17.0 / 15.8 & \underline{71.3} / \underline{61.1} & 1312 / 1455 & \underline{52.4} & 43.5 & 23.0 / 26.4 & 77.1 / 75.8 \\
\hline
\xmark & \cmark & \xmark & \underline{3041} & \underline{76.9} & \textbf{\underline{87.6}} & \underline{15.6} / 14.7 & 13.1 / 16.8 & 1239 / \underline{1365} & \underline{53.0} & 43.0 & \underline{15.5} / \underline{22.2} & 78.4 / \underline{79.3} \\
\xmark & \cmark & \cmark & 3125 & 72.1 & 84.6 & 16.0 / \underline{14.0} & \underline{56.2} / \underline{54.7} & \underline{1215} / 1416 & 52.9 & \underline{42.1} & 18.2 / \textbf{\underline{17.7}} & \underline{78.9} / 78.1 \\
\hline
\cmark & \xmark & \xmark & 3425 & \underline{76.2} & \underline{86.5} & 15.6 / 13.8 & 16.9 / 26.6 & \underline{1220} / 1550 & \underline{55.3} & \textbf{\underline{36.5}} & \underline{13.9} / 21.2 & \underline{81.7} / \textbf{\underline{81.7}} \\
\cmark & \xmark & \cmark & \underline{3019} & 72.5 & 84.3 & \underline{14.4} / \underline{13.5} & \underline{74.9} / \underline{64.3} & 1239 / \underline{1419} & 53.3 & 38.0 & 15.7 / \underline{19.8} & 80.5 / 79.9 \\
\hline
\cmark & \cmark & \xmark & 3095 & \textbf{\underline{77.5}} & \textbf{\underline{87.6}} & 15.4 / 13.1 & 18.1 / 24.7 & 1180 / 1456 & \textbf{\underline{55.8}} & \textbf{\underline{36.5}} & 13.9 / \textbf{\underline{17.7}} & 81.5 / \underline{81.4} \\
\cmark & \cmark & \cmark & \textbf{\underline{2849}} & 72.8 & 84.6 & \textbf{\underline{13.9}} / \textbf{\underline{12.8}} & \textbf{\underline{75.3}} / \textbf{\underline{65.4}} & \underline{1178} / \underline{1371} & 52.9 & 36.8 & \textbf{\underline{12.9}} / 18.9 & \textbf{\underline{82.1}} / 80.7 \\
\hline
\end{tabularx}
\end{subtable}
\end{table*}

\begin{table*}[t]
\ContinuedFloat
\centering
\setlength{\tabcolsep}{1.6pt}
\renewcommand{\arraystretch}{1.6}
\scriptsize
\caption{(continued)}

\begin{subtable}{\textwidth}
\centering
\caption{Gemini 3.1 Flash Lite}
\label{tab:main_gemini}
\begin{tabularx}{\textwidth}{
>{\centering\arraybackslash}p{0.6cm} 
>{\centering\arraybackslash}p{0.5cm} 
>{\centering\arraybackslash}p{0.8cm} 
>{\centering\arraybackslash}p{1.0cm} 
>{\centering\arraybackslash}p{0.55cm} 
>{\centering\arraybackslash}p{0.55cm} 
>{\centering\arraybackslash}p{1.8cm} 
>{\centering\arraybackslash}p{1.7cm} 
>{\centering\arraybackslash}p{1.8cm} 
>{\centering\arraybackslash}p{0.8cm} 
>{\centering\arraybackslash}p{1.3cm} 
>{\centering\arraybackslash}p{1.6cm} 
>{\centering\arraybackslash}p{2.1cm} 
}
\textbf{OAS} & \textbf{KE} & \textbf{OANA} & \textbf{Sim Steps $\downarrow$} & \textbf{SR \%$\uparrow$} & \textbf{PC \%$\uparrow$} & \textbf{Plan. Cycles (R / H) $\downarrow$} & \textbf{OA Mention (R / H) \%$\uparrow$} & \textbf{Exploration Eff. (R/H) $\downarrow$} & \textbf{Offload \%$\uparrow$} & \textbf{Extra Effo. \%$\downarrow$} & \textbf{Extra Plan. (R / H) \%$\downarrow$} & \textbf{Action Success (R / H) \%$\uparrow$} \\
\hline
\xmark & \xmark & \xmark & \underline{4118} & \underline{80.4} & \underline{89.8} & \underline{10.2} / \underline{11.1} & 30.4 / 37.8 & 2177 / 2262 & 54.7 & 44.3 & \underline{26.9} / \underline{36.3} & \underline{81.1} / \underline{80.1} \\
\xmark & \xmark & \cmark & 4304 & 76.0 & 88.6 & 10.8 / 12.4 & \underline{91.7} / \textbf{\underline{91.1}} & \underline{1735} / \underline{1653} & \underline{55.1} & \underline{43.5} & 33.6 / 38.0 & 80.5 / 75.7 \\
\hline
\xmark & \cmark & \xmark & \underline{3081} & \underline{79.9} & \underline{89.4} & \underline{8.2} / \underline{8.7} & 24.3 / 31.6 & 1744 / 1798 & 54.6 & 38.2 & \underline{16.4} / 25.1 & 85.1 / \underline{84.6} \\
\xmark & \cmark & \cmark & 3416 & 75.8 & 88.3 & 8.7 / 8.9 & \textbf{\underline{92.3}} / \underline{90.2} & \underline{1649} / \underline{1634} & \underline{55.3} & \underline{35.1} & 23.6 / \underline{22.5} & \underline{85.2} / 83.5 \\
\hline
\cmark & \xmark & \xmark & \underline{3193} & \textbf{\underline{81.2}} & \underline{89.8} & \underline{8.3} / \underline{8.9} & 27.7 / 38.2 & 1666 / 1772 & \underline{55.7} & 33.0 & \underline{16.6} / 26.7 & \textbf{\underline{85.6}} / \textbf{\underline{85.5}} \\
\cmark & \xmark & \cmark & 3453 & 79.1 & 88.6 & 8.8 / 9.4 & \underline{89.8} / \underline{88.3} & \underline{1438} / \underline{1424} & 55.3 & \underline{32.2} & 21.7 / \underline{22.5} & 85.0 / 83.4 \\
\hline
\cmark & \cmark & \xmark & \textbf{\underline{2911}} & \underline{81.1} & \textbf{\underline{90.3}} & \textbf{\underline{8.0}} / \textbf{\underline{8.5}} & 27.5 / 36.5 & 1617 / 1638 & \textbf{\underline{56.4}} & 33.3 & \textbf{\underline{14.9}} / 20.5 & \textbf{\underline{85.6}} / \underline{84.2} \\
\cmark & \cmark & \cmark & 3107 & 78.6 & 89.1 & 8.5 / 8.7 & \underline{90.6} / \underline{88.5} & \textbf{\underline{1368}} / \textbf{\underline{1359}} & 55.8 & \textbf{\underline{32.1}} & 20.0 / \textbf{\underline{16.5}} & 84.2 / 83.7 \\
\hline
\end{tabularx}
\end{subtable}

\vspace{0.6em}

\begin{subtable}{\textwidth}
\centering
\caption{DeepSeek V3.2}
\label{tab:main_deepseek}
\begin{tabularx}{\textwidth}{
>{\centering\arraybackslash}p{0.6cm} 
>{\centering\arraybackslash}p{0.5cm} 
>{\centering\arraybackslash}p{0.8cm} 
>{\centering\arraybackslash}p{1.0cm} 
>{\centering\arraybackslash}p{0.55cm} 
>{\centering\arraybackslash}p{0.55cm} 
>{\centering\arraybackslash}p{1.8cm} 
>{\centering\arraybackslash}p{1.7cm} 
>{\centering\arraybackslash}p{1.8cm} 
>{\centering\arraybackslash}p{0.8cm} 
>{\centering\arraybackslash}p{1.3cm} 
>{\centering\arraybackslash}p{1.6cm} 
>{\centering\arraybackslash}p{2.1cm} 
}
\textbf{OAS} & \textbf{KE} & \textbf{OANA} & \textbf{Sim Steps $\downarrow$} & \textbf{SR \%$\uparrow$} & \textbf{PC \%$\uparrow$} & \textbf{Plan. Cycles (R / H) $\downarrow$} & \textbf{OA Mention (R / H) \%$\uparrow$} & \textbf{Exploration Eff. (R/H) $\downarrow$} & \textbf{Offload \%$\uparrow$} & \textbf{Extra Effo. \%$\downarrow$} & \textbf{Extra Plan. (R / H) \%$\downarrow$} & \textbf{Action Success (R / H) \%$\uparrow$} \\
\hline
\xmark & \xmark & \xmark & \underline{3857} & \underline{82.3} & \underline{91.0} & \underline{13.3} / \underline{13.3} & 39.1 / 43.6 & \underline{1458} / \underline{1584} & 54.8 & 38.2 & \underline{20.3} / \underline{28.0} & 89.5 / 88.4 \\
\xmark & \xmark & \cmark & 5486 & 75.2 & 87.1 & 18.9 / 16.6 & \underline{99.9} / \textbf{\underline{100}} & 1641 / 1987 & \underline{55.8} & \underline{35.0} & 39.0 / 38.4 & \underline{92.1} / \underline{89.7} \\
\hline
\xmark & \cmark & \xmark & \underline{2951} & \underline{82.2} & \underline{90.6} & \textbf{\underline{12.2}} / \underline{11.4} & 35.9 / 40.9 & \underline{1380} / \underline{1471} & 53.8 & 34.8 & \underline{16.0} / \underline{19.3} & 91.9 / \underline{92.3} \\
\xmark & \cmark & \cmark & 4249 & 75.6 & 87.4 & 16.4 / 13.4 & \textbf{\underline{100}} / \textbf{\underline{100}} & 1521 / 1709 & \textbf{\underline{57.9}} & \underline{31.9} & 36.4 / 29.4 & \underline{93.7} / 91.5 \\
\hline
\cmark & \xmark & \xmark & \underline{3112} & \textbf{\underline{83.6}} & \textbf{\underline{91.5}} & \underline{12.6} / \underline{11.5} & 49.5 / 52.6 & \underline{1237} / \underline{1396} & 54.6 & 28.1 & \underline{15.1} / \underline{17.9} & 92.7 / \textbf{\underline{92.8}} \\
\cmark & \xmark & \cmark & 3927 & 78.4 & 88.8 & 14.8 / 12.5 & \textbf{\underline{100}} / \underline{99.9} & 1341 / 1571 & \underline{55.5} & \textbf{\underline{26.7}} & 30.9 / 23.6 & \underline{93.2} / 90.3 \\
\hline
\cmark & \cmark & \xmark & \textbf{\underline{2794}} & \underline{82.4} & \underline{90.5} & \textbf{\underline{12.2}} / \textbf{\underline{11.0}} & 48.6 / 51.4 & \textbf{\underline{1214}} / \textbf{\underline{1381}} & 52.5 & 27.5 & \textbf{\underline{13.6}} / \textbf{\underline{17.1}} & 92.8 / \underline{92.6} \\
\cmark & \cmark & \cmark & 3551 & 80.0 & 89.4 & 14.0 / 11.4 & \textbf{\underline{100}} / \textbf{\underline{100}} & 1328 / 1462 & \underline{53.6} & \underline{27.0} & 30.5 / 19.8 & \textbf{\underline{93.9}} / 91.3 \\
\hline
\end{tabularx}
\end{subtable}
\end{table*}
 
\subsubsection{Task Outcomes and Coordination}
\label{app:partial_task}

\textbf{What the OANA prediction buys is planner-family dependent.} For the Llama models, OANA reduces Planning Cycles for both agents in six of eight settings (8B in all four, e.g.\ 24.2/23.2$\rightarrow$21.2/20.9), yet Simulation Steps do not fall correspondingly for the 8B (4214$\rightarrow$4243) and Percent Complete declines in seven of the eight settings (it is unchanged for the 8B with KE only), indicating fewer but longer plans and earlier termination rather than more efficient planning. For Gemini the opposite holds: Planning Cycles rise slightly in all four settings ($+0.5$ on average) while Exploration Efficiency improves consistently (2177/2262$\rightarrow$1735/1653, 1744/1798$\rightarrow$1649/1634, 1666/1772$\rightarrow$1438/1424, 1617/1638$\rightarrow$1368/1359), so prediction makes this planner commit to objects sooner at a small planning cost. DeepSeek pays the highest overhead (Planning Cycles rise by up to 5.6, and Extra Planning nearly doubles, from 20.3\% to 39.0\% for the robot without shared context), but its robot Action Success improves under prediction in every setting (as does the Llama 8B's), peaking at 93.9\%; its human agent declines in three of the four. Prediction thus improves the quality of individual actions for some planners while degrading the joint outcome.

\textbf{Model capacity sets the baseline quality of coordination and determines how safely additional context can be added.} Extraneous Effort at each model's best setting decreases monotonically with planner strength (Llama 8B 58.5\%, Llama 70B 36.5\%, Gemini 32.1\%, DeepSeek 26.7\%), while Task Offloading stays within a narrow 52.4--60.7\% band across all 32 runs: the performance differences come from better temporal alignment and less duplicated work, not from shifting the workload between the agents. Exploration Efficiency separates the two families most clearly. For the Llama planners, the settings without OAS give the fastest first object pickup (70B 909/1100 at the baseline; 8B 1300/1741 with KE only, its worst Success Rate configuration), and richer context delays the first interaction; for Gemini and DeepSeek, the richest settings give the fastest pickup (1368/1359 and 1214/1381). Weaker planners, therefore, behave greedily and early, and additional interaction signals can hurt them, whereas stronger planners convert the same signals into faster search and better task outcomes alike.

\textbf{OANA turns successes into ordinary failures, not near-misses.} The Percent Complete--Success Rate gap widens under prediction in 15 of 16 model$\times$context pairs, but this follows arithmetically from the lower Success Rate: more episodes fail, and each failed episode still contributes partial progress. The informative quantity is the average completion of a failed episode, $(\mathrm{PC}-\mathrm{SR})/(100-\mathrm{SR})$, which lies between 38\% and 53\% in all 32 runs and rises under prediction in only 8 of the 16 pairs. Prediction, therefore, converts some would-be successes into failures that, like failures without prediction, complete about half of the task's propositions. Across the no-prediction settings, the same quantity is 38--41\% for the Llama 8B and 43--49\% for the three stronger planners, so weaker planners also leave more of a task undone when they fail, not merely fail more often.

\textbf{Object knowledge without partner state is a weak lever on redundant work and a counterproductive one for the Llama planners.} Adding KE to the object-only baseline raises Extraneous Effort for both Llama planners (8B 62.3\%$\rightarrow$69.6\%, 70B 39.4\%$\rightarrow$43.0\%), and although it lowers Extraneous Effort for Gemini (44.3\%$\rightarrow$38.2\%) and DeepSeek (38.2\%$\rightarrow$34.8\%), OAS alone lowers it substantially further in both cases (33.0\% and 28.1\%). OAS is thus the dominant lever on redundant rearrangement for every model, and KE without OAS is at best a weak substitute and at worst counterproductive. This generalizes the 8B failure discussed in Section~\ref{sec:task}: synchronizing object knowledge tells an agent that an object exists and where it is, but not that the partner is already acting on it, so the newly visible objects become candidate targets for both agents simultaneously. The 70B exhibits the same direction as the 8B in Extraneous Effort (though not in Extra Planning) without a Success Rate cost (76.0\%$\rightarrow$76.9\%), which suggests the mechanism is general and its severity is governed by planner capacity.

\textbf{Prediction partially repairs the ungrounded-KE failure mode, which supports the attribution account.} Adding OANA to the Llama 8B's pathological KE-only configuration reduces Simulation Steps from 4914 to 4121, Planning Cycles from 27.7/25.1 to 20.7/20.4, and Extraneous Effort from 69.6\% to 62.7\%, while raising Action Success from 64.5/62.4\% to 71.6/66.0\%. It is also the only setting in which prediction raises Success Rate (54.3\%$\rightarrow$56.2\%, with Percent Complete unchanged at 72.9\%), although by less than the standard error of a difference between two single runs (about 2.2 points). Under the attribution account, this follows naturally: KE supplies object knowledge that the 8B cannot assign to an owner, and OANA forces the missing assignment step by requiring an explicit hypothesis about what the partner will do with those objects. Prediction here substitutes, imperfectly, for the partner-state signal that OAS would otherwise provide.

\textbf{Spontaneous partner reasoning tracks task success across models; mandated partner reasoning does not.} In the Decoupled baseline, the Other-Agent Mention Ratio rises with planner strength (robot: 9.8\%, 13.3\%, 30.4\%, 39.1\% for Llama 8B, 70B, Gemini, DeepSeek; human: 14.6\%, 17.2\%, 37.8\%, 43.6\%) and orders the models identically to Success Rate (60.9\%, 76.0\%, 80.4\%, 82.3\%). Stronger planners attend to the partner more often without being asked to. OANA raises the same metric to 55--92\% for the Llama and Gemini planners and saturates it at essentially 100\% for DeepSeek, yet Success Rate falls in fifteen of the sixteen cases (the exception is the 8B with KE only, discussed above). The metric is therefore informative as an unprompted behavioral signal and uninformative once it is instructed, which we take as evidence that endogenous partner modeling and prompt-compliant partner mentioning are distinct phenomena that this metric cannot separate on its own.

\textbf{Planning volume is not a proxy for planning competence.} Gemini attains 81.1\% Success Rate using 8.0/8.5 planning cycles per episode, while Llama 8B attains 62.3\% using 21.5/19.8, and DeepSeek attains 82.4\% using an intermediate 12.2/11.0. Expressed as simulator steps per high-level action at each model's OAS + KE setting, Gemini issues roughly 176 steps per action (both agents' planning cycles pooled) against 92 for the 8B, 109 for the 70B, and 120 for DeepSeek. Gemini therefore commits to substantially longer-horizon actions and revises them rarely, whereas the 8B replans frequently over short horizons and converts that additional deliberation into neither fewer simulator steps nor higher success. Frequent replanning in these settings is a symptom of unstable commitment or a poor grasp of the environment and its actions, not careful planning.

\textbf{KE is the cheaper intervention and OAS is the more accurate one.} For all three stronger planners, KE alone yields fewer Simulation Steps than OAS alone (Llama 70B 3041 vs.\ 3425, Gemini 3081 vs.\ 3193, DeepSeek 2951 vs.\ 3112), while OAS alone yields the higher Success Rate in three of four models (Gemini 81.2\% vs.\ 79.9\%, DeepSeek 83.6\% vs.\ 82.2\%, Llama 8B 58.8\% vs.\ 54.3\%; the Llama 70B reverses this at 76.2\% vs.\ 76.9\%). Since object knowledge synchronization requires only shared world-graph state while partner-state observation requires continuous monitoring of the collaborator, this matters practically: deployments constrained to one channel should choose based on whether latency or task correctness is the binding constraint, and the two channels are close enough to be redundant for strong planners that implementing both is justified mainly by the modest additional efficiency it buys.

\subsubsection{Role Asymmetry Between the Agents}
\label{app:partial_roles}
 

\textbf{Division of labor is nearly invariant to every intervention; what changes is timing.} Task Offloading stays within 52.4--60.7\% across all 32 runs, a range of eight points, while Success Rate varies from 54.3\% to 83.6\% and Simulation Steps from 2794 to 5486 over the same runs. No setting causes agents to renegotiate who does what to any meaningful degree. 
The small movements that do occur are mostly toward the robot: OANA raises its share in 10 of 16 pairs, most for the 8B baseline (57.0\%$\rightarrow$60.7\%) and DeepSeek with KE only (53.8\%$\rightarrow$57.9\%), leaves it unchanged for the 70B baseline, and lowers it slightly, by at most 2.9 points, for the 70B with any shared context and for Gemini with OAS enabled.
The performance differences reported in Section \ref{sec:task} are consequently produced by better temporal alignment, fewer duplicated actions, and earlier object acquisition, not by a different allocation of work.

\textbf{The robot reaches its first object interaction earlier than the human in 28 of 32 runs.} The mean Exploration Efficiency gap (human minus robot) is 482 simulator steps for the Llama 8B, 
205 for the Llama 70B, 180 for DeepSeek, and 18 for Gemini. The four exceptions are exactly Gemini's four OANA configurations, where the human is marginally faster (1368/1359 at OAS + KE). The robot's action repertoire consists entirely of primitives that begin with locating and picking an object, so every robot plan converges on a pickup, whereas a substantial share of the human's available actions operate on objects already in place and do not require one. The gap is largest for the weakest planner and nearly vanishes for Gemini, indicating that it reflects how strongly a planner is steered by the shape of its action list rather than by a hard environmental constraint.

\textbf{The human is the trailing agent: it performs a higher rate of replanning than the robot after the other has stopped taking action in 25 of 32 runs.} Averaged over all configurations, the Extra Planning gap (human minus robot) is $+15.4$ points for the Llama 8B, $+4.9$ for the 70B, $+4.3$ for Gemini, and $-1.0$ for DeepSeek (the average gap in extra planning cycles shows the same pattern: +3.0 for the 8B, $+0.5$ for the 70B, $+0.6$ for Gemini, and $-0.6$ for DeepSeek). Restricting to the no-prediction settings, the richest shared context compresses this gap in every model: from $+14.5$ to $+11.4$ points for the 8B, $+8.1$ to $+3.8$ for the 70B, $+9.4$ to $+5.6$ for Gemini, and $+7.7$ to $+3.5$ for DeepSeek as the configuration moves from object-only observation to OAS + KE. This asymmetry coexists with Task Offloading above 50\% in all 32 runs, meaning the robot completes more subtasks by count while nonetheless going idle first. Together with the prompted roles, a natural reading is that the human's exclusive state-modification actions sit at the ends of dependency chains: an object usually must be transported to the correct receptacle before it needs to be cleaned, filled or switched on, so the human's distinctive contributions are serialized behind the robot's rearrangements and tend to be the last work remaining. Subtask-count offloading consequently overstates how much the robot relieves its partner, because the robot's contribution comes early in the episode and the human completes the remaining work.

\textbf{Shared context helps the team primarily by shortening the trailing window.} Extra Planning falls for the human from 37.7\% to 32.0\% (8B), 25.4\% to 17.7\% (70B), 36.3\% to 20.5\% (Gemini) and 28.0\% to 17.1\% (DeepSeek) between the object-only baseline and OAS + KE, in every case a larger absolute reduction than for the robot over the same comparison. Because episode length is determined by the later-finishing agent, improvements concentrated on the trailing agent translate directly into the Simulation Step reductions reported in Section~\ref{sec:task}. This also explains why OAS and KE improve efficiency more reliably than they improve Success Rate: they compress idle and redundant time at the end of the episode without changing which agent is capable of finishing the task.

\textbf{Prediction reverses the direction of the lag for strong planners and amplifies it for the weakest one.} Under OANA, DeepSeek's robot Extra Planning roughly doubles (20.3\%$\rightarrow$39.0\% at baseline, 13.6\%$\rightarrow$30.5\% at OAS + KE), turning the gap negative (human minus robot) in all four contexts ($-0.6$, $-7.0$, $-7.3$, $-10.7$ points); Gemini shows the same reversal in two of four contexts, and the Llama 70B a marginal one with KE only ($-0.5$ points). For the 8B, the gap instead widens from $+14.5$ to $+22.1$ points at baseline and from $+11.4$ to $+17.3$ at OAS + KE. Since the robot must predict from the human's 14 action types while the human predicts from the robot's 9, prediction imposes an asymmetric burden that strong planners absorb as additional robot-side deliberation after the partner has finished, while the weakest planner instead leaves its human agent with more remaining work.

\textbf{Without prediction, the human reasons about the partner more often; with prediction, the robot does so with the stronger planners.} The Other-Agent Mention Ratio is higher for the human in all 16 no-prediction runs, across every model and context (for example 21.0/30.9\% for the Llama 8B and 27.5/36.5\% for Gemini at OAS + KE). Enabling OANA flips the ordering for Gemini in all four contexts (90.6/88.5\% at OAS + KE) and for the Llama 70B in all four (75.3/65.4\% at OAS + KE), saturates it for DeepSeek near 100\% on both sides, and leaves it unchanged for the 8B. The spontaneous pattern is consistent with dependence driving attention: the agent whose work is blocked by the partner has more reason to model the partner. The induced pattern is consistent with prediction difficulty driving reasoning length: the robot faces the larger prediction space and spends more of its reasoning trace on the partner once prediction is mandatory. That these two orderings point in opposite directions is a further reason to treat mention frequency as a behavioral signal rather than a quality measure.

\textbf{The robot executes actions more reliably than the human, and prediction widens that gap.} Robot Action Success is at least as high as human Action Success in 28 of 32 runs, with the largest margins for the 8B (up to 5.6 points, e.g., 72.0\% vs.\ 66.4\% with OAS only) and near parity for the 70B and DeepSeek. Under OANA, the change in Action Success is less favorable for the human than for the robot in 13 of 16 model$\times$context pairs, for instance, $-0.6$ vs.\ $-4.4$ points for Gemini at baseline and $+0.5$ vs.\ $-2.5$ for DeepSeek with OAS only. The human's larger action space contains the actions with the most restrictive preconditions, since state modification requires the object to be present, accessible, and in the appropriate configuration, so mispredicting the partner has more ways to invalidate a human action than a robot action. Anticipatory execution is thus riskier for whichever agent holds the precondition-heavy repertoire.

\textbf{Planning-cycle asymmetry is planner-specific rather than environment-imposed.} The robot issues more planning cycles than the human in all eight 8B runs, all eight 70B runs, and seven of eight DeepSeek runs, but fewer in all eight Gemini runs (8.0/8.5 at OAS + KE). Gemini is also the planner with the smallest exploration gap between the agents. The same asymmetric action spaces therefore produce opposite planning-cycle orderings depending on the planner, which indicates that role specialization here is a joint property of the task structure and the model, not a mechanical consequence of the environment. Reporting per-agent rather than averaged metrics is necessary to observe this; agent-averaged planning cycles are nearly identical for Gemini and would conceal the reversal entirely.
 
\subsubsection{Partner Prediction}
\label{app:partial_pred}

\textbf{Combining OAS observation with object KE gives the best Action Type prediction accuracy for every model and both agents.} The OAS + KE configuration gives the top Action Type accuracy in all eight model×agent combinations, strictly in seven and tied in the eighth (Llama 8B 37.9/34.2\%, Llama 70B 55.5/55.8\%, Gemini 52.4/52.7\%, DeepSeek 49.0/54.7\%, R/H), and the top accuracy at every other unconditional abstraction level for the Llama 70B and DeepSeek planners. Relative to the object-only baseline, the gains are substantial for the three stronger planners: $+17.0/{+}14.0$ points for the Llama 70B, $+14.4/{+}16.3$ for Gemini, and $+21.1/{+}23.5$ for DeepSeek on Action Type. Access to the partner's state and to a synchronized object world graph therefore both supply information that predicts the partner's next action, which is not obvious a priori: neither channel transmits intentions, plans, or language, only observed states and object locations.

\textbf{OAS observation is the stronger single channel for three of the four planners, but the two channels are substantially redundant rather than additive.} Taking Action Type as the reference metric, the sum of the two single-channel gains exceeds the gain from enabling both in seven of the eight model$\times$agent combinations, and often by a wide margin: for Gemini, the separate gains total +22.1/+27.8 points while the combination delivers only +14.4/+16.3, a realization of roughly 60\%. DeepSeek realizes about 83\% (+25.1/+28.6 separately vs.\ +21.1/+23.5 together), the Llama 70B 87\% and 64\% (+19.5/+21.8 vs.\ +17.0/+14.0), and only the 8B robot is marginally super-additive (+5.8 vs.\ +6.1). Partner state and shared object knowledge thus overlap in what they reveal about the partner's next action, which is consistent with their mechanisms: observing that the partner is \textit{Picking apple\_0} and learning that \textit{apple\_0} has appeared in the shared world graph or its position has changed, despite the agent not interacting with the object, are two routes to the same inference. The practical consequence mirrors Section~\ref{sec:task}: implementing the second channel buys considerably less than implementing the first.

\textbf{Shared context tells an agent what its partner will do far more than which object it will act on.} Averaged over the four planners, moving from the object-only baseline to OAS + KE improves Action Type accuracy by +14.7 points for the robot and +14.6 for the human, but improves the conditional principal-parameter accuracy by only +4.2 and +1.0 points, respectively. The asymmetry is visible in the extreme cases: Gemini's human agent improves its Action Type accuracy from 36.4\% to 52.7\% while its conditional principal-parameter accuracy moves from 52.6\% to 51.5\%, so essentially the entire joint improvement for that agent comes from the type factor. Even at the best-performing cell in the whole table (70B with OAS + KE), roughly a third of correctly typed predictions still carry the wrong principal parameter (67.2/64.1\% conditional accuracy) and about 40\% carry a wrong full parameter set (61.4/59.4\%). The residual difficulty in partner modeling, therefore, lies in the arguments rather than the verbs, and it is the component least responsive to the interventions we study.

\textbf{For the smallest planner, richer context improves the type term and degrades the parameter term, leaving joint accuracy nearly unchanged.} The Llama 8B is the only model whose conditional accuracies fall clearly below baseline for both agents (elsewhere the largest drop is 2.0 points, Gemini's human with OAS): enabling OAS drops conditional principal-parameter accuracy from 54.2\% to 49.2\% (robot) and conditional all-parameter accuracy from 49.4\% to 45.5\%, with OAS + KE no better. Because its Action Type accuracy simultaneously rises from 31.8\% to 36.1\% (37.9\% with OAS+KE), the two factors nearly cancel, and the joint action-type-plus-all-parameters accuracy moves only from 15.7\% to 16.4\% (16.9\%). The decomposition reveals what the joint metric hides: the Llama 8B does extract partner information from OAS and convert it into better coarse predictions, but the additional prompt content displaces the object-level reasoning needed to fill in arguments, so the newly correctly typed predictions are disproportionately ones it cannot complete. This is the prediction-side counterpart of the capacity sensitivity documented in Section~\ref{sec:task}, and it recurs with the opposite polarity: in task solving, KE without OAS was the 8B's pathological configuration, whereas in prediction, KE is the only channel that helps it ($+1.1/{+}3.7$ points of conditional principal-parameter accuracy) and OAS is the one that hurts.

\textbf{Whether object knowledge alone suffices depends on the planner, not on the task.} For Gemini, KE alone recovers or exceeds the entire benefit of OAS alone on Action Type ($+11.4$ vs.\ $+10.7$ for the robot, $+15.3$ vs.\ $+12.5$ for the human, i.e.\ ratios above one), and for DeepSeek, it recovers roughly 60\% of it. For the Llama planners, the same channel recovers less: about a third on the robot side and about half on the human side for the 70B (+5.1 vs.\ +14.4 and +7.8 vs.\ +14.0, ratios of 0.35 and 0.56), and about a third for the 8B (at most 0.37). A plausible account is that a synchronized object graph supports prediction by self-projection: once both agents hold the same beliefs about where objects are, an agent can predict the partner by asking what it would itself do next, and this route is used more by planners that re-derive the partner's situation than by those that read off the partner's state. The Llama planners appear to rely mainly on the direct partner-state signal, whereas the API-accessed planners, Gemini in particular, exploit common ground. This distinction is invisible in the joint metrics and only emerges once the KE-only condition is run.

\textbf{Parameters are the binding constraint on partner prediction, and the errors are asymmetric.} Among predictions that name every parameter correctly, the action type is also correct 84.8\% of the time on average across the 32 model$\times$context$\times$agent cells (range 77.4--92.9\%), whereas among predictions with the correct action type, only 52.2\% carry a fully correct parameter set (range 44.5--61.4\%). Getting the arguments right, therefore, almost implies getting the verb right, while the converse fails roughly half the time. The residual error mode is nonetheless informative: about one in seven parameter-correct predictions still attaches the wrong verb, so agents that have correctly identified the objects their partner is engaged with can still misjudge the manipulation being performed on them. We note that the direction of this asymmetry follows arithmetically from the action type being the easier marginal (Bayes relates the two conditionals by the ratio of the marginals), so it quantifies the size of the parameter bottleneck rather than establishing it independently.

\textbf{Task-solving strength does not predict partner-modeling strength, and for the strongest planner, the two rankings are close to inverted.} 
DeepSeek V3.2 achieves the best Success Rate, Percent Complete, and Action Success of all four planners in Table~\ref{tab:main_full}, yet at the object-only baseline, its robot is the weakest partner predictor in the table, below even the 8B's (27.9\% vs. 31.8\% Action Type). Conversely, the 4-bit 70B is the third-best task solver but the best predictor under full shared context (55.5/55.8\%). Prediction accuracy and task competence are thus dissociable: a planner can be highly effective at deciding its own next action while modeling its partner poorly, as expected if strong task performance comes mainly from individual planning rather than from coordination. This also supplies independent confirmation of the argument in Section~\ref{sec:task} that partner-mention frequency is not a quality measure. DeepSeek mentions the other agent in essentially 100\% of replanning steps under OANA in every context, the highest rate of any planner, while producing some of the least accurate partner predictions.

\textbf{Within the API-accessed planners, prediction accuracy tracks the size of the success-rate penalty that OANA imposes.} For DeepSeek, the correspondence is monotone across all four contexts: as baseline Action Type accuracy rises through 27.9\%, 37.1\%, 43.8\% and 49.0\% for no context, KE, OAS and OAS + KE, the Success Rate cost of enabling prediction falls through $-7.1$, $-6.6$, $-5.2$ and $-2.4$ points. Gemini follows the ordering only loosely: its penalty tracks whether OAS is available ($-4.4$ and $-4.1$ points without OAS, $-2.1$ and $-2.5$ with it) more closely than accuracy, since KE-only reaches higher Action Type accuracy than OAS-only (49.4/51.7\% vs. 48.7/48.9\%) yet costs about twice as much. This quantifies the mechanism proposed in Section~\ref{sec:task}: agents act on their predictions, so inaccurate anticipation translates into premature commitment and lost task propositions, and improving the prediction reduces the damage. We note that the relationship does not extend to the Llama planners, whose penalties do not shrink as accuracy rises, so we present it as an account of the API-accessed models rather than a general law; disentangling it properly would require intervening on prediction quality directly, which our design does not do.

\textbf{Constrained decoding cannot explain the cross-family differences in action type, but may contribute to those in parameters.} The locally run Llama planners use PARTNR's grammar-constrained decoding (Appendix~\ref{app:constraint}), which restricts every predicted call to a valid action and, for OANA predictions, restricts objects to the classes present in the scene from the start of the episode; the API-accessed planners have neither restriction. On action type the grammar confers little, and the data agree: with KE only, the unconstrained Gemini still predicts the partner's action type better than the constrained Llama 70B (49.4/51.7\% against 43.6/49.6\%). On parameters, the picture differs: the Llama 70B's conditional principal-parameter accuracy (62.2--67.6\%) exceeds Gemini's (50.6--58.3\%) and DeepSeek's (50.2--62.2\%) in every context and for both agents, the direction expected if knowing the scene's object classes helps fill in arguments, although the equally constrained Llama 8B (49.2--57.2\%) shows that grammar alone is not sufficient. We therefore treat action-type comparisons across model families as reliable and parameter-level comparisons as favoring the Llama planners.

The two agents face structurally different prediction problems. The human predicts over the robot's 9 action types, while the robot predicts over the human's 14, which include the object state manipulation actions (\textit{Clean, Fill, Pour, PowerOn,} and \textit{PowerOff}) absent from the robot's repertoire. 
Both run the same planner, so systematic differences between the R and H columns of Table \ref{tab:oana_prediction_accuracy} reflect this asymmetry and the prompted roles (Section \ref{sec:roles}) rather than model capability.
Under the union target, we find a double dissociation; Appendix~\ref{sec:appendix_action_pred_eval} tests which parts of it survive the strict target and full observability.

\textbf{Under the union target, the human predicts the partner's action type more accurately.} For the Llama 70B and DeepSeek, the human leads in all four contexts (+0.3 to +6.0 and +3.3 to +5.7 points, respectively), and Gemini leads on the human side in three of four. The effect is in the direction implied by the action-space sizes. 
For the Llama 70B, it also shrinks once OAS is available, from $+3.3$ at baseline ($+6.0$ with KE only) to $+2.9$ with OAS and $+0.3$ with OAS + KE, indicating that richer context partially compensates the robot for its larger hypothesis space rather than helping both agents equally. This lead is not stable across scoring and regime: under the strict next-action target, it disappears in the partial regime, where the robot matches or leads in 14 of 16 cells, and for the two Llama planners, it reappears under full observability (Appendix \ref{sec:appendix_action_pred_eval}). The pattern suggests that the smaller hypothesis space helps only once object uncertainty is removed, and that under partial observability, part of the union-target lead comes from recognizing the robot's ongoing navigation.


\textbf{The robot predicts parameters more accurately once the action type is right, and this advantage appears only with shared context.} At the object-only baseline, the conditional principal-parameter gaps are negligible in every model ($-0.7$, $+0.8$, $+0.9$ and $-1.5$ points H$-$R for 8B, 70B, Gemini and DeepSeek). Enabling either channel opens a robot-side lead that persists in all three stronger planners: $-3.1$ to $-4.2$ points for the 70B, $-5.7$ to $-7.1$ for Gemini, and $-2.0$ to $-4.9$ for DeepSeek, with the same pattern at the all-parameter level. The likely source is the arity of the two action repertoires rather than the size of their parameter vocabularies, which the action grammar (Appendix \ref{sec:appendix_constraint_grammar}) makes comparable: \textit{Navigate} ranges over furniture, rooms, and objects alike, and the human's \textit{Clean}, \textit{PowerOn} and \textit{PowerOff} range over furniture and objects alike. What differs is structure. The five action types available only to the human (\textit{PowerOn}, \textit{PowerOff}, \textit{Clean}, \textit{Fill}, \textit{Pour}) all take a single argument, whereas \textit{Place} and \textit{Rearrange}, both in the robot's nine-type repertoire, each take five slots comprising an object, a spatial relation, a furniture target, and an optional constraint with its reference entity. Conditional on a correct action type, the agent predicting the robot therefore draws from a pool in which the two compound actions carry proportionally more weight, and a single wrong slot is enough to fail the all-parameter criterion; the agent predicting the human additionally draws from five single-slot actions on which parameter prediction is a much simpler problem. This also offers a reading of why the gap is absent at baseline: without shared context neither agent can name the objects and receptacles involved, so both directions fail on the arguments regardless of arity, and the asymmetry only becomes expressible once OAS and KE make the single-argument cases nearly determined while leaving the compound cases underdetermined. 
Under the strict target, this robot-side lead persists clearly only for Gemini (Appendix \ref{subsubsec:appendix_action_pred_eval_next}), so we offer the arity account as a hypothesis specific to union scoring.

\textbf{The smallest planner inverts the type asymmetry, which suggests it is not exploiting the smaller action space.} The 8B robot outpredicts the 8B human on Action Type in all four contexts ($-2.2$ to $-3.7$ points H$-$R), reversing the pattern seen in every stronger planner. A planner that reasons over the partner's available actions should find the 9-action problem easier than the 14-action one. That the 8B does not suggest that its predictions are driven less by the structure of the partner's action space than by surface features of the prompt, which is consistent with its failure to convert correct type predictions into correct arguments discussed above.

\textbf{The two asymmetries cancel, so the aggregate metrics look symmetric while the underlying competences are not.} On the strictest joint metric, action type plus all parameters, the H$-$R gap stays within $\pm3.3$ points in all 16 cells and within $\pm2.5$ for three of the four planners, even though its two factors reach $+6.0$ points in one direction (Action Type, 70B with KE) and $-7.0$ in the other (All Params if Type, Gemini with OAS). Reporting only the joint accuracies would therefore suggest that the two agents model each other about equally well, when in fact, under union scoring, the human is better at anticipating what its partner will do and the robot at anticipating what it will do it to. The decomposition into type and conditional parameter accuracy makes the asymmetry visible, and it is a reason to report conditional accuracies rather than joint accuracies alone in settings with heterogeneous action spaces.

\subsection{Evaluation under Full Observability}
\label{sec:appendix_task_solving_analysis}

\begin{table*}[t]
\centering
\setlength{\tabcolsep}{1.6pt}
\renewcommand{\arraystretch}{1.6}
\scriptsize

\caption{\textbf{Task solving performance and coordination metrics under full observability setting} with different Other Agent's State (OAS) observation and Other Agent's Next Action (OANA) prediction settings. The 4-bit quantized Llama-3.1-70B-Instruct and Llama-3.1-8B-Instruct serve as planners for both agents. MS: model size, SR: Success Rate, PC: Percent Complete, OA Mention: Other-Agent Mention Ratio, R/H: Robot/Human agent. Subgroup bests are underlined; overall bests are bolded.}
\label{tab:task_solving_full_obs}

\begin{tabularx}{\textwidth}{
>{\centering\arraybackslash}p{0.4cm} 
>{\centering\arraybackslash}p{0.6cm} 
>{\centering\arraybackslash}p{0.8cm} 
>{\centering\arraybackslash}p{1.0cm} 
>{\centering\arraybackslash}p{0.55cm} 
>{\centering\arraybackslash}p{0.55cm} 
>{\centering\arraybackslash}p{1.8cm} 
>{\centering\arraybackslash}p{1.7cm} 
>{\centering\arraybackslash}p{1.8cm} 
>{\centering\arraybackslash}p{0.8cm} 
>{\centering\arraybackslash}p{1.3cm} 
>{\centering\arraybackslash}p{1.6cm} 
>{\centering\arraybackslash}p{2.1cm} 
}
\textbf{MS} & \textbf{OAS} & \textbf{OANA} & \textbf{Sim Steps $\downarrow$} & \textbf{SR \%$\uparrow$} & \textbf{ PC \%$\uparrow$} & \textbf{Plan. Cycles (R / H) $\downarrow$} & \textbf{OA Mention (R / H) \%$\uparrow$} & \textbf{Exploration Eff. (R/H) $\downarrow$} & \textbf{Offload \%$\uparrow$} & \textbf{Extra Effo. \%$\downarrow$} & \textbf{Extra Plan. (R / H) \%$\downarrow$} & \textbf{Action Success (R / H)  \%$\uparrow$} \\
\hline
\multirow{4}{*}{8B} 
& \xmark & \xmark 
& 2383 & \underline{64.7} & \underline{79.5} 
& 20.6 / 19.0 
& 11.6 / 15.4 
& \underline{421} / 455
& 51.1 
& 60.0 
& 22.7 / 20.9
& \underline{73.0} / \underline{68.9} \\

& \xmark & \cmark 
& 3113 & 58.5 & 72.9 
& 17.2 / 17.6 
& 67.5 / 74.9
& 1054 / 1306 
& 54.6 
& 61.0 
& 19.1 / 28.9 
& 71.0 / 65.4 \\

& \cmark & \xmark 
& \underline{2001} & 62.6 & 78.5 
& 18.1 / 18.3 
& 23.5 / 32.4 
& 498 / \underline{452} 
& 51.5 
& \underline{54.8} 
& 18.9 / \underline{16.8}
& 70.9 / 62.4 \\

& \cmark & \cmark 
& 2890 & 60.3 & 75.3 
& \underline{17.1} / \underline{15.7} 
& \underline{72.1} / \textbf{\underline{77.9}} 
& 1053 / 1296 
& \textbf{\underline{56.7}} 
& 57.0 
& \underline{13.8} / 24.3 
& 72.4 / 68.6 \\

\hline
\multirow{4}{*}{70B} 
& \xmark & \xmark 
& 1742 & \textbf{\underline{76.9}} & \textbf{\underline{87.7}} 
& 12.8 / 12.5 
& 14.3 / 15.5 
& 387 / 428 
& 49.1 
& 35.6 
& 11.5 / 15.4 
& 84.0 / \textbf{\underline{83.3}} \\

& \xmark & \cmark 
& 1825 & 75.3 & 86.6 
& 13.4 / 12.0 
& 50.8 / 52.9 
& 456 / 539 
& \underline{51.8}
& 37.7 
& 14.5 / 13.4 
& 81.6 / 79.7 \\

& \cmark & \xmark 
& \textbf{\underline{1611}} & \textbf{\underline{76.9}} & 87.3 
& \textbf{\underline{12.1}} / \textbf{\underline{11.6}} 
& 19.8 / 21.8 
& \textbf{\underline{385}} / \textbf{\underline{411}} 
& 49.9 
& \textbf{\underline{31.5}} 
& 8.2 / \textbf{\underline{11.1}} 
& \textbf{\underline{84.2}} / 82.9 \\

& \cmark & \cmark 
& 1681 & 76.3 & 87.0 
& \textbf{\underline{12.1}} / \textbf{\underline{11.6}} 
& \textbf{\underline{73.3}} / \underline{65.4}
& 426 / 515 
& 51.4 
& 35.5 
& \textbf{\underline{7.7}} / 11.4 
& 81.2 / 79.5 \\

\hline
\end{tabularx}
\end{table*}

We additionally evaluate the effects of OAS observation and OANA prediction under PARTNR's \textit{Full} observability setting, where both agents use the complete global ground-truth world graph at all times. In this setting, at each replanning, beyond room and furniture information, agents observe all objects and their locations (textual description in the prompt), independent of their own positions. Consequently, explicit exploration is no longer required, although it remains available.

Since object information is globally accessible, the object KE is not applicable anymore, and the true OAS, if enabled, is also accessible continuously, irrespective of the agents’ relative locations (agents in different rooms no longer receive the ``\textit{Acting in another room}'' observation). The OANA prediction follows the same protocol as in the partial observability setting (Section~\ref{sec:oana}): at each replanning step, agents first reason about the partner’s and their own next actions and then predict the partner's next action before selecting their own action.



\subsubsection{Task Performance and Coordination under Full Observability}
\label{sec:appendix_task_analysis_full}

\begin{table*}[t]
\centering
\setlength{\tabcolsep}{1.4pt}
\renewcommand{\arraystretch}{1.8}
\scriptsize

\caption{\textbf{Comparison of task performance and coordination with enabled OANA prediction} across different observability scenarios. OAS: Other Agent's State, KE: Knowledge Exchange, SR: Success Rate, PC: Percent Complete, Plan. Cycles: Planning Cycles per agent (R: Robot / H: Human), OA Mention: Other-Agent Mention Ratio (R / H \%). Subgroup bests are underlined, overall bests are bolded. Larger models and fuller observability generally lead to better task success and more efficient planning. Note: For full observability, object KE is not applicable.}
\label{tab:compare_dp_full_obs_with_oana}

\begin{tabularx}{\textwidth}{
>{\centering\arraybackslash}p{1.5cm} 
>{\centering\arraybackslash}p{2.5cm} 
>{\centering\arraybackslash}p{0.7cm} 
>{\centering\arraybackslash}p{0.7cm} 
>{\centering\arraybackslash}p{1.0cm} 
>{\centering\arraybackslash}p{1.8cm} 
>{\centering\arraybackslash}p{1.1cm} 
>{\centering\arraybackslash}p{1.1cm} 
>{\centering\arraybackslash}X 
>{\centering\arraybackslash}X 
>{\centering\arraybackslash}X 
>{\centering\arraybackslash}X 
}
\hline
\multirow{2}{*}{\textbf{Model Size}} & \multirow{2}{*}{\textbf{Observability}} & \multirow{2}{*}{\textbf{OAS}} & \multirow{2}{*}{\textbf{KE}} & \multirow{2}{*}{\textbf{OANA}} & \multirow{2}{*}{\textbf{Sim Steps $\downarrow$}} & \multirow{2}{*}{\textbf{SR \%$\uparrow$}} & \multirow{2}{*}{\textbf{PC \%$\uparrow$}} & \multicolumn{2}{c}{\textbf{Plan. Cycles $\downarrow$}} & \multicolumn{2}{c}{\textbf{OA Mention \%$\uparrow$}} \\
\cline{9-12}
& & & & & & & & \textbf{R} & \textbf{H} & \textbf{R} & \textbf{H} \\
\hline
\multirow{5}{*}{8B} 
& Decoupled Partial & \xmark & \xmark & \cmark & 4243 & 55.0 & 73.0 & 21.2 & 20.9 & 85.5 & 88.7 \\
& Decoupled Partial & \xmark & \cmark & \cmark & 4121 & 56.2 & 72.9 & 20.7 & 20.4 & \textbf{\underline{87.7}} & \textbf{\underline{92.1}}\\
& Decoupled Partial & \cmark & \xmark & \cmark & 4017 & 56.6 & 74.2 & 19.8 & 19.4 & 74.7 & 81.2 \\
& Decoupled Partial & \cmark & \cmark & \cmark & 3890 & 57.2 & 74.5 & 19.2 & 18.7 & 75.6 & 80.8 \\
& Full & \xmark & - & \cmark & 3113 & 58.5 & 72.9 & 17.2 & 17.6 & 67.5 & 74.9 \\
& Full & \cmark & - & \cmark & \underline{2890} & \underline{60.3} & \underline{75.3} & \underline{17.1} & \underline{15.7} & 72.1 & 77.9 \\
\hline
\multirow{5}{*}{70B} 
& Decoupled Partial & \xmark & \xmark & \cmark & 3648 & 72.9 & 85.6 & 17.0 & 15.8 & 71.3 & 61.1 \\
& Decoupled Partial & \xmark & \cmark & \cmark & 3125 & 72.1 & 84.6 & 16.0 & 14.0 & 56.2 & 54.7 \\
& Decoupled Partial & \cmark & \xmark & \cmark & 3019 & 72.5 & 84.3 & 14.4 & 13.5 & 74.9 & 64.3 \\
& Decoupled Partial & \cmark & \cmark & \cmark & 2849 & 72.8 & 84.6 & 13.9 & 12.8 & \underline{75.3} & \underline{65.4} \\
& Full & \xmark & - & \cmark & 1825 & 75.3 & 86.6 & 13.4 & 12.0 & 50.8 & 52.9 \\
& Full & \cmark & - & \cmark & \textbf{\underline{1681}} & \textbf{\underline{76.3}} & \textbf{\underline{87.0}} & \textbf{\underline{12.1}} & \textbf{\underline{11.6}} & 73.3 & \underline{65.4} \\
\hline
\end{tabularx}
\end{table*}

\begin{table*}[t]
\centering
\setlength{\tabcolsep}{1.4pt}
\renewcommand{\arraystretch}{1.8}
\scriptsize

\caption{\textbf{Comparison of task performance and coordination with disabled OANA prediction} across different observability scenarios. OAS: Other Agent's State, KE: Knowledge Exchange, SR: Success Rate, PC: Percent Complete, Plan. Cycles: Planning Cycles per agent (R: Robot / H: Human), OA Mention: Other-Agent Mention Ratio (R / H \%). Subgroup bests are underlined, overall bests are bolded. Larger models and fuller observability generally lead to better task success and more efficient planning. Note: For full observability, object KE is not applicable.}
\label{tab:compare_dp_full_obs_without_oana}

\begin{tabularx}{\textwidth}{
>{\centering\arraybackslash}p{1.5cm} 
>{\centering\arraybackslash}p{2.5cm} 
>{\centering\arraybackslash}p{0.7cm} 
>{\centering\arraybackslash}p{0.7cm} 
>{\centering\arraybackslash}p{1.0cm} 
>{\centering\arraybackslash}p{1.8cm} 
>{\centering\arraybackslash}p{1.1cm} 
>{\centering\arraybackslash}p{1.1cm} 
>{\centering\arraybackslash}X 
>{\centering\arraybackslash}X 
>{\centering\arraybackslash}X 
>{\centering\arraybackslash}X 
}
\hline
\multirow{2}{*}{\textbf{Model Size}} & \multirow{2}{*}{\textbf{Observability}} & \multirow{2}{*}{\textbf{OAS}} & \multirow{2}{*}{\textbf{KE}} & \multirow{2}{*}{\textbf{OANA}} & \multirow{2}{*}{\textbf{Sim Steps $\downarrow$}} & \multirow{2}{*}{\textbf{SR \%$\uparrow$}} & \multirow{2}{*}{\textbf{PC \%$\uparrow$}} & \multicolumn{2}{c}{\textbf{Plan. Cycles $\downarrow$}} & \multicolumn{2}{c}{\textbf{OA Mention \%$\uparrow$}} \\
\cline{9-12}
& & & & & & & & \textbf{R} & \textbf{H} & \textbf{R} & \textbf{H} \\
\hline
\multirow{5}{*}{8B} 
& Decoupled Partial & \xmark & \xmark & \xmark & 4214 & 60.9 & 77.1 & 24.2 & 23.2 & 9.8 & 14.6 \\
& Decoupled Partial & \xmark & \cmark & \xmark & 4914 & 54.3 & 72.9 & 27.7 & 25.1 & 9.5 & 17.3 \\
& Decoupled Partial & \cmark & \xmark & \xmark & 4022 & 58.8 & 74.7 & 22.1 & 21.1 & 20.5 & \textbf{\underline{32.7}} \\
& Decoupled Partial & \cmark & \cmark & \xmark & 3781 & 62.3 & 76.8 & 21.5 & 19.8 & 21.0 & 30.9 \\
& Full & \xmark & - & \xmark & 2383 & \underline{64.7} & \underline{79.5} & 20.6 & 19.0 & 11.6 & 15.4 \\
& Full & \cmark & - & \xmark & \underline{2001} & 62.6 & 78.5 & \underline{18.1} & \underline{18.3} & \textbf{\underline{23.5}} & 32.4 \\
\hline
\multirow{5}{*}{70B} 
& Decoupled Partial & \xmark & \xmark & \xmark & 3056 & 76.0 & 86.9 & 15.7 & 15.1 & 13.3 & 17.2 \\
& Decoupled Partial & \xmark & \cmark & \xmark & 3041 & 76.9 & 87.6 & 15.6 & 14.7 & 13.1 & 16.8 \\
& Decoupled Partial & \cmark & \xmark & \xmark & 3425 & 76.2 & 86.5 & 15.6 & 13.8 & 16.9 & \underline{26.6} \\
& Decoupled Partial & \cmark & \cmark & \xmark & 3095 & \textbf{\underline{77.5}} & 87.6 & 15.4 & 13.1 & 18.1 & 24.7 \\
& Full & \xmark & - & \xmark & 1742 & 76.9 & \textbf{\underline{87.7}} & 12.8 & 12.5 & 14.3 & 15.5 \\
& Full & \cmark & - & \xmark & \textbf{\underline{1611}} & 76.9 & 87.3 & \textbf{\underline{12.1}} & \textbf{\underline{11.6}} & \underline{19.8} & 21.8 \\
\hline
\end{tabularx}
\end{table*}

Table~\ref{tab:task_solving_full_obs} reports the same metrics under PARTNR's full observability setting, in which both agents hold complete and continuously updated object knowledge from the start of the episode. Object KE is not applicable here: full observability already gives every agent all object positions at all times, so it subsumes and strictly exceeds the co-location-gated synchronization that KE provides. The design therefore reduces to a $2\times2$ grid of OAS observation and OANA prediction, run with the Llama-3.1-8B and 4-bit Llama-3.1-70B planners. Tables~\ref{tab:compare_dp_full_obs_with_oana} and~\ref{tab:compare_dp_full_obs_without_oana} place these runs alongside the corresponding Decoupled Partial Observability configurations with and without OANA prediction, respectively. Because the comparison spans only two planners from one family, we treat it as a controlled probe of what perceptual uncertainty contributes to the effects reported in Sections~\ref{sec:task}--\ref{sec:roles}, not as an independent evaluation.

\textbf{Full observability roughly halves execution time while leaving task success essentially unchanged for the stronger planner.} At matched settings, Simulation Steps fall by 43\% for both planners at the object-only baseline (Llama 8B 4214$\rightarrow$2383, Llama 70B 3056$\rightarrow$1742) and by 50\% and 53\% respectively with OAS enabled (4022$\rightarrow$2001 and 3425$\rightarrow$1611). Success Rate moves far less. The Llama 70B reaches 76.9\% under full observability, which does not exceed the 77.5\% it attains under Decoupled Partial Observability with OAS + KE, and its best Percent Complete is 87.7\% against 87.6\% in the partial regime, a difference well within the uncertainty of a single run. The Llama 8B does gain, from 62.3\% to 64.7\%, but 2.4 points of Success Rate is a modest return on the complete elimination of perceptual uncertainty. Task accuracy for these planners is therefore mostly not limited by perception: a perfect world model makes them much faster at the tasks they already solve but does not let them solve appreciably more.

\textbf{Continuous object knowledge improves speed but not accuracy.} The clearest comparison is the Llama 70B without OAS or prediction: KE-only under Decoupled Partial Observability yields 76.9\% Success Rate, 87.6\% Percent Complete and 3041 Simulation Steps, while full observability yields 76.9\%, 87.7\% and 1742 steps. Success Rate is identical, and Percent Complete differs by 0.1 points, while the episode is 43\% shorter. Upgrading from intermittent, proximity-triggered object synchronization to permanent, complete knowledge thus converts entirely into execution efficiency. This supports the reading of KE in Section \ref{sec:task}: KE mainly improves efficiency, and even complete object knowledge does not raise the accuracy ceiling.

\textbf{Exploration Efficiency collapses, which confirms what the metric measures in the partial regime.} First successful pickup arrives 57--79\% earlier under full observability at every matched setting without prediction (Llama 70B baseline 909/1100$\rightarrow$387/428, Llama 70B with OAS 1220/1550$\rightarrow$385/411, Llama 8B baseline 1348/1795$\rightarrow$421/455, Llama 8B with OAS 1793/2166$\rightarrow$498/452). The magnitude is expected, since with all object positions known in advance, there is nothing to search for, and the residual 385--500 steps are navigation time rather than exploration. This confirms that the Exploration Efficiency differences reported in Section~\ref{sec:roles} and Appendix \ref{app:partial_task} tracked search behavior and not, for example, differences in how quickly a planner commits to a first action.

\textbf{Redundant work is a coordination failure, not a perception failure: most extraneous effort survives perfect object knowledge.} Extraneous Effort at each planner's best full-observability setting is 54.8\% for the Llama 8B and 31.5\% for the Llama 70B, against 58.5\% and 36.5\% at their best partial-observability settings. Removing all perceptual uncertainty therefore eliminates only about 4 and 5 points of redundant rearrangement, leaving roughly 94\% and 86\% of it in place. Agents with identical, complete, and correct world models still perform a third to three-fifths of their rearrangements unnecessarily. Since neither agent can act on stale or wrong object information in this regime, the remaining redundancy must come from both agents independently selecting the same task-relevant object, a failure that partner-state access, not world knowledge, has to address.

\textbf{OAS observation keeps its coordination benefits under full observability but loses its task-quality benefits, and for the weakest planner it becomes net harmful.} With prediction disabled, enabling OAS still reduces Extraneous Effort for both planners (Llama 8B 60.0\%$\rightarrow$54.8\%, Llama 70B 35.6\%$\rightarrow$31.5\%) and still reduces Extra Planning for both agents (22.7/20.9\%$\rightarrow$18.9/16.8\% and 11.5/15.4\%$\rightarrow$8.2/11.1\%), with both reductions comparable to or larger than the corresponding ones in the partial regime. Success Rate, however, does not follow: it is unchanged for the Llama 70B (76.9\% in both cases, with Percent Complete slipping 87.7\%$\rightarrow$87.3\%) and falls for the Llama 8B (64.7\%$\rightarrow$62.6\%), whose human agent also drops to 62.4\% Action Success, the lowest value anywhere in Table~\ref{tab:task_solving_full_obs}. This happens even though OAS is more available here than in the partial regime: under full observability, the partner's state is observed continuously rather than only on co-location (Appendix \ref{sec:appendix_task_solving_analysis}), while the co-located percept sharing that OAS also provides (Section \ref{sec:info}) becomes redundant, since every object is already known. Read together with Section~\ref{sec:task}, this separates two things OAS was doing at once. Its coordination contribution, telling an agent what its partner is occupied with, and hence what not to duplicate, is intrinsic and survives. Its contribution to task outcomes was substantially a matter of compensating for incomplete world knowledge, and once that gap is closed, the added prompt content is a cost that the 8B, in particular, cannot absorb.

\textbf{OANA prediction lowers Success Rate and Percent Complete in all four full-observability configurations, bringing the count to nineteen of twenty across both regimes (the exception being the partial-observability Llama 8B with KE only).} The penalties are $-6.2$ and $-2.3$ points of Success Rate for the Llama 8B without and with OAS, and $-1.6$ and $-0.6$ points for the Llama 70B, with Percent Complete moving in the same direction in every case and the Other-Agent Mention Ratio rising from 12--32\% to 51--78\%. That the effect holds without exception under perfect perception rules out one otherwise plausible account of it, namely that prediction hurts because agents anticipate partners whose situation they cannot see. Agents that know everything about the world, and that are asked to predict a partner acting in that same fully known world, still complete fewer task propositions than agents that simply act.

\textbf{The size of that penalty, however, depends on planner capacity.} The Llama 70B's penalties under full observability ($-1.6$ and $-0.6$ points) are the smallest it incurs in any configuration in this paper, against $-3.1$ to $-4.8$ points across the four partial-observability settings, and its Simulation Step overhead falls to +83 and +70 steps (under 5\%), against +592 at the partial-observability baseline. The Llama 8B shows no such relief: its penalties under full observability ($-6.2$, $-2.3$) match those in the corresponding partial-observability settings ($-5.9$, $-2.2$) almost exactly, and its Simulation Step overhead grows to $+730$ and $+889$ steps, a 31\% and 44\% increase on episodes that are already the shortest it produces. Applying the grounding argument of Section \ref{sec:task} to this setting, complete world knowledge makes prediction nearly free for the capable planner and no cheaper at all for the weak one.

\textbf{Under full observability, the exploration cost of prediction is pure deliberation overhead, and it is large for the smaller planner.} Enabling OANA delays the first pickup from 421/455 to 1054/1306 steps for the Llama 8B and from 387/428 to 456/539 for the Llama 70B. In the partial regime, such a delay is ambiguous between slower search and slower decision-making; here, there is nothing to search for, so the entire 2.5--2.9$\times$ increase for the 8B is time spent deliberating before acting. Prediction consumes a substantial share of this planner's early decision-making without improving what it decides, which is consistent with its joint prediction accuracy barely moving with added context (Section~\ref{sec:prediction}).

\textbf{The human-trailing asymmetry is largely produced by partial observability rather than by the action spaces alone.} Defining the completion lag as human minus robot Extra Planning, as in Section~\ref{sec:roles}, the no-prediction lag shrinks from $+8.1$ to $+3.9$ points at baseline and from $+7.3$ to $+2.9$ with OAS for the Llama 70B, and reverses outright for the Llama 8B, from $+14.5$ to $-1.8$ at baseline and from $+13.3$ to $-2.1$ with OAS. Task Offloading moves in the same direction, falling from 55.1–60.7\% (8B) and 52.4–55.8\% (70B) in the partial regime to 51.1--56.7\% and 49.1--51.8\%, respectively; the 70B's share even falls below parity in both of its no-prediction settings (49.1\% and 49.9\%). The dependency-chain account of Section~\ref{sec:roles} predicts a residual lag, since state-modification actions still come after transport, and for the 70B, a residual of $+2.9$ to $+3.9$ points is what remains. But the bulk of the gap was an exploration artifact: under partial observability, the human's slower search pushed its work later into the episode and inflated the robot's share of completed subtasks. This qualifies the assistive-robotics caution in Section~\ref{sec:roles}: the finisher role is real but considerably milder once the partner is not also searching. Prediction restores the asymmetry in three of four cases, driving the 8B lag to $+9.8$ and $+10.5$ points, consistent with the robot's larger prediction space imposing an asymmetric deliberation burden.

\textbf{The robot's execution advantage, by contrast, is unaffected by observability, which supports attributing it to the action spaces.} Robot Action Success exceeds human Action Success in all eight full-observability runs, by $+3.8$ to $+8.5$ points for the Llama 8B and $+0.7$ to $+1.9$ for the Llama 70B, the same ordering and roughly the same magnitudes as the 28 of 32 runs in the partial regime. Full observability also lifts the 70B's best Action Success by about 2 points (best 84.2/83.3\% against 82.1/81.7\%). An asymmetry that persists unchanged when perceptual uncertainty is removed is most naturally explained by the precondition structure of the human's state-modification actions rather than by anything the agents can or cannot see.

Overall, full observability saves much time but adds little accuracy: it removes search almost entirely, leaves most redundant work untouched, and roughly halves the completion lag between the agents for the 70B while removing it for the 8B. What it does not do is substitute for knowing what one's partner is doing, which is why OAS retains its coordination effects here while losing its task-quality ones, and why prediction remains costly even when nothing is hidden.

\subsection{Further Action Prediction Evaluations}
\label{sec:appendix_action_pred_eval}

\subsubsection{Next Action Target for Decoupled Partial Observability}
\label{subsubsec:appendix_action_pred_eval_next}

\begin{table*}[t]
\centering
\setlength{\tabcolsep}{1.6pt}
\renewcommand{\arraystretch}{1.6}
\scriptsize
\caption{\textbf{Other Agent's Next Action (OANA) prediction accuracy (\%)} across action abstraction levels for robot (R) and human (H) agents under \textbf{Decoupled Partial} observability. The \textbf{target action is the next action} of the other agent. Results highlight that OAS observation, alone or combined with KE, improves prediction for the three stronger planners, while the Llama 8B gains almost nothing at the strictest level.}
\label{tab:oana_prediction_accuracy_next}
\begin{tabularx}{\textwidth}{
>{\centering\arraybackslash}p{0.9cm} 
>{\centering\arraybackslash}p{0.6cm} 
>{\centering\arraybackslash}p{0.6cm} 
>{\centering\arraybackslash}p{1.6cm} 
>{\centering\arraybackslash}p{1.9cm} 
>{\centering\arraybackslash}p{2.0cm} 
>{\centering\arraybackslash}p{1.9cm} 
>{\centering\arraybackslash}p{1.9cm} 
>{\centering\arraybackslash}p{1.9cm} 
>{\centering\arraybackslash}p{2.0cm} 
}
\textbf{Model} & \textbf{OAS} & \textbf{KE} 
& \textbf{Action Type (R/H) $\uparrow$} 
& \textbf{Principal Par. (R/H) $\uparrow$} 
& \textbf{A. Type+Princ. Par. (R/H) $\uparrow$} 
& \textbf{All Params (R/H) $\uparrow$} 
& \textbf{A. Type+All Params (R/H) $\uparrow$} 
& \textbf{Principal Par. if Type (R/H) $\uparrow$} 
& \textbf{All Params if Type (R/H) $\uparrow$} \\
\hline
\multirow{3}{0.9cm}{\centering Llama-3.1-8B-Instruct}
& \xmark & \xmark 
& 23.1 / 20.7 & 18.0 / 15.4 & 11.5 / 10.4 & 11.4 / 9.6 & 10.1 / 8.8 & 50.0 / 50.1 & \underline{43.6} / 42.4 \\
& \xmark & \cmark 
& 24.0 / 21.6 & \underline{19.2} / \underline{17.2} & \underline{12.2} / 12.2 & \underline{12.0} / \underline{10.8} & \underline{10.4} / 9.9 & \underline{50.8} / \underline{56.4} & 43.3 / \underline{45.5} \\
& \cmark & \xmark 
& 25.8 / \underline{23.6} & 15.2 / 15.0 & 11.3 / \underline{12.3} & 10.9 / \underline{10.8} & 10.0 / \underline{10.2} & 43.8 / 51.8 & 38.7 / 43.1 \\
& \cmark & \cmark 
& \underline{26.2} / 23.4 & 15.8 / 14.8 & 11.9 / 11.4 & 11.1 / 10.4 & 9.9 / 9.6 & 45.5 / 48.5 & 37.9 / 41.1 \\
\hline
\multirow{3}{0.9cm}{\centering Llama-3.1-70B-Instruct 4-bit}
& \xmark & \xmark 
& 27.5 / 26.9 & 26.2 / 26.2 & 17.3 / 18.5 & 19.0 / 19.9 & 15.4 / 16.5 & 63.0 / 68.7 & 56.1 / 61.4 \\
& \xmark & \cmark 
& 30.7 / 31.2 & 30.1 / 26.9 & 21.5 / 21.4 & 22.1 / 20.7 & 18.6 / 18.4 & 69.8 / 68.5 & 60.6 / 58.9 \\
& \cmark & \xmark 
& 34.7 / \textbf{\underline{33.3}} & 30.7 / 28.7 & 25.0 / \textbf{\underline{24.5}} & 24.8 / 24.2 & 21.7 / \textbf{\underline{22.2}} & 72.2 / 73.6 & 62.6 / 66.8 \\
& \cmark & \cmark 
& \textbf{\underline{35.4}} / 32.6 & \textbf{\underline{32.7}} / \textbf{\underline{28.8}} & \textbf{\underline{26.3}} / 24.4 & \textbf{\underline{26.2}} / \textbf{\underline{24.3}} & \textbf{\underline{23.2}} / 21.9 & \textbf{\underline{74.4}} / \textbf{\underline{75.0}} & \textbf{\underline{65.5}} / \textbf{\underline{67.3}} \\
\hline
\multirow{3}{0.9cm}{\centering Gemini 3.1 Flash Lite}
& \xmark & \xmark 
& 23.1 / 21.0 & 16.8 / 13.4 & 11.3 / 9.7 & 12.8 / 10.4 & 9.9 / 8.6 & 48.8 / 46.0 & 43.0 / 41.1 \\
& \xmark & \cmark 
& \underline{30.4} / \underline{29.4} & 23.5 / 18.8 & 17.6 / \underline{14.6} & 19.7 / \underline{15.9} & 16.3 / \underline{13.5} & 58.1 / 49.7 & 53.7 / 46.0 \\
& \cmark & \xmark 
& 27.6 / 25.9 & 24.6 / 18.8 & 16.7 / 12.9 & 20.0 / 14.3 & 15.8 / 11.9 & 60.6 / 49.8 & 57.2 / 45.8 \\
& \cmark & \cmark 
& 30.3 / 27.3 & \underline{27.3} / \underline{20.1} & \underline{18.7} / 13.7 & \underline{22.1} / 15.4 & \underline{17.4} / 12.7 & \underline{61.8} / \underline{50.3} & \underline{57.6} / \underline{46.4} \\
\hline
\multirow{3}{0.9cm}{\centering DeepSeek V3.2}
 & \xmark & \xmark 
& 18.7 / 18.5 & 15.3 / 15.0 & 10.0 / 9.6 & 11.4 / 10.9 & 8.9 / 8.4 & 53.6 / 51.8 & 47.5 / 45.5 \\
& \xmark & \cmark 
& 24.4 / 26.4 & 20.8 / 21.8 & 15.1 / 15.3 & 16.0 / 16.7 & 13.6 / 13.9 & 61.9 / 57.8 & 55.5 / 52.6 \\
& \cmark & \xmark 
& 28.0 / 27.1 & 24.5 / 23.5 & 19.0 / 18.0 & 20.7 / 19.4 & 17.8 / 16.7 & \underline{67.9} / 66.2 & \underline{63.6} / 61.7 \\
& \cmark & \cmark 
& \underline{31.5} / \underline{31.5} & \underline{27.7} / \underline{26.9} & \underline{21.3} / \underline{21.2} & \underline{22.9} / \underline{22.0} & \underline{19.8} / \underline{19.5} & 67.8 / \underline{67.2} & 63.1 / \underline{61.8} \\
\hline
\end{tabularx}
\end{table*}


Table~\ref{tab:oana_prediction_accuracy_next} re-scores the same OANA-enabled runs against a stricter target: only the partner's \emph{subsequent} high-level action counts as correct, whereas Table~\ref{tab:oana_prediction_accuracy} also credited a prediction that matched the action the partner was executing at the moment of prediction (Section~\ref{sec:oana}). The two tables therefore describe identical agent behavior scored under two readings of what it means to anticipate a partner. The union target measures whether an agent has a correct model of what its partner is doing \emph{or} about to do; the strict target isolates genuine anticipation. Comparing them separates the part of our prediction results that reflects reading the partner's present state from the part that reflects projecting its future.

\textbf{The strict target removes roughly a third of all prediction accuracy, and on average, the entire loss sits in the action-type term.} Averaged over the 32 model$\times$context$\times$agent cells, Action Type accuracy falls by 16.0 points, or 36.5\% in relative terms, with the drop ranging from $-8.7$ points (Llama 8B, baseline, robot) to $-25.4$ (Gemini, OAS + KE, human). 
The conditional accuracies do not fall on average (all-parameter 52.2\% $\rightarrow$ 52.4\%), but they split by planner. For the Llama 70B the conditional principal-parameter accuracy rises in all eight cells, by as much as $+11.4$ points (OAS only, human, 62.2\% → 73.6\%), and for DeepSeek in all eight, by up to $+9.8$; for the Llama 8B it falls in all eight, and for Gemini in six of eight. Since the joint metrics factorize as Acc(type) × Acc(params | type), an identity we again verified in all 64 cells of Table \ref{tab:oana_prediction_accuracy_next}, the stricter target makes the coarse problem markedly harder for every planner, while making the fine problem somewhat easier for the 70B and DeepSeek and somewhat harder for the 8B and Gemini.

\textbf{That divergence identifies what the union target was crediting: ongoing actions that are easy to type and hard to parameterize.} When an agent observes its partner \textit{Walking in hallway\_1}, answering \textit{Navigate}[$\cdot$] scores a correct action type under the union target without any anticipation, because the ongoing action is directly readable from the OAS observation, but the destination argument remains unconstrained. Such events inflate type accuracy and depress conditional parameter accuracy simultaneously. Excluding them, as the strict target does, therefore produces the pattern we observe for the Llama 70B and DeepSeek, the two planners whose union-target Action Type gains most from OAS (Section \ref{sec:prediction}); for the 8B and Gemini, the conditional term falls as well, so this account covers only part of the data. This also means the union results should be read as a mixture of two capabilities: state recognition, which is comparatively easy and which the strict target removes, and anticipation, which is what remains here. The gap between the two tables is a quantitative estimate of how much of apparent partner modeling is the former.

\textbf{Shared context still helps, and for the three stronger planners it helps substantially, but the smallest planner gains nothing once the ongoing action is excluded.} Moving from the object-only baseline to OAS + KE improves the strictest joint metric (action type plus all parameters) by $+7.8/{+}5.4$ points for Llama 70B, $+7.5/{+}4.1$ for Gemini and $+10.9/{+}11.1$ for DeepSeek (R/H). For the Llama 8B, the same move yields $-0.2/{+}0.8$ points: its robot side is marginally \emph{worse} with full shared context than without it (10.1\%$\rightarrow$9.9\%). The union-target analysis already found the 8B's gains nearly cancelling between the two factors (15.7\%$\rightarrow$16.9\%); under the strict target, the cancellation is complete. Whatever the 8B extracts from partner state and shared object knowledge supports recognizing what its partner is currently doing and does not support predicting what it will do next.

\textbf{Gemini's best strict-target prediction comes from object knowledge alone, not from partner state.} Under the union target, OAS + KE topped Action Type in all eight model×agent combinations (strictly in seven). Under the strict target, it tops only four. Gemini is best with KE only on both sides (30.4/29.4\% against 30.3/27.3\% with OAS + KE, and against 27.6/25.9\% with OAS alone), and both Llama planners peak on the human side with OAS only. The KE-versus-OAS contrast reported in Section~\ref{sec:prediction} consequently becomes stronger for Gemini, where KE alone now delivers a larger Action Type gain than OAS alone on both sides (+7.3/+8.4 against +4.5/+4.9), while for the Llama 70B it delivers about half of the OAS gain on the robot side and two-thirds on the human side (+3.2/+4.3 against +7.2/+6.4). The self-projection account we proposed, in which a synchronized object graph lets an agent predict its partner by asking what it would itself do next, is an account of \emph{anticipation} specifically, and it is under the anticipation-only target that Gemini's reliance on object knowledge is most visible.

\textbf{Under the strict target, the human's action-type advantage disappears in the partial regime.} Under the union target, the human agent led the robot on Action Type in 11 of 16 cells, and we attributed this to the human predicting over the robot's 9 action types while the robot predicts over the human's 14. Under the strict target, the ordering inverts: the robot matches or leads in 14 of 16 cells. The Llama 70B's gap moves from $+3.3$ to $-0.6$ points at baseline and from $+0.3$ to $-2.8$ with OAS + KE; DeepSeek's moves from $+3.3$ to $-0.2$ and from $+5.7$ to $0.0$. The relative accuracy loss is also consistently larger on the human side for the three stronger planners (for example, $-41.6\%$ against $-36.2\%$ for the Llama 70B with OAS + KE), and only slightly larger (0.7--2.6 points) for the 8B. A target-space-size effect should be indifferent to whether the target is the current or the next action; a state-recognition effect should not. This suggests that the human's apparent edge at coarse partner prediction came largely from the robot's ongoing actions being easy to read, which follows from the robot's role as the transport layer (Section~\ref{sec:roles}) and hence its disproportionate time spent navigating, rather than from the human facing a smaller hypothesis space.

\textbf{The robot's conditional parameter advantage is not robust either and survives clearly only for Gemini.} Under the union target, the robot led on conditional principal-parameter accuracy in the three stronger planners once either channel was enabled. Under the strict target, this holds only for Gemini, where it in fact widens considerably (from $-5.7$, $-7.1$, $-6.6$ to $-8.4$, $-10.8$, $-11.5$ points H$-$R for KE, OAS and OAS + KE). For the Llama 70B, the lead shrinks or reverses ($+5.7$, $-1.3$, $+1.4$, $+0.6$); for DeepSeek, it shrinks to at most $-4.1$; and for the 8B, the human leads in all four contexts. The double dissociation reported in Section~\ref{sec:prediction} is thus a property of the union scoring rather than a stable feature of the two prediction problems, and the arity-based account we offered for it applies to that scoring only.

\textbf{One consequence of this is that the joint metric is no longer close to symmetric for every planner.} The H$-$R gap on action type plus all parameters stayed within $\pm3.3$ points in all 16 union cells because the two factor-level asymmetries pointed in opposite directions and cancelled. Under the strict target, the cancellation survives for the Llama 8B ($\pm1.3$), the Llama 70B ($\pm1.3$) and DeepSeek ($\pm1.1$), but fails for Gemini, whose robot leads by 2.8, 3.9 and 4.7 points once any shared context is enabled, because its type asymmetry and its conditional asymmetry now both favor the robot. Symmetry in an aggregate metric is therefore coincidental rather than structural, which is a further argument for reporting the two factors separately.

\textbf{The parameter bottleneck is the same under both targets.} Among predictions naming every parameter correctly, the action type is also correct 85.7\% of the time on average (range 77.1--94.4\%), against 84.8\% under the union target; among correctly-typed predictions, 52.4\% carry a fully correct parameter set (range 37.9--67.3\%), against 52.2\%. The two scorings differ by 0.9 and 0.2 points on these conditionals, despite differing by 16 points on unconditional type accuracy. Getting the arguments right almost implies getting the verb right regardless of how the target is defined, and the reverse fails about half the time in both cases. The asymmetry between the coarse and fine sub-problems is thus a property of partner modeling in this environment and not an artifact of the temporal flexibility in our matching protocol.

\textbf{Absolute anticipation accuracy is low enough to reframe the trade-off documented in Section~\ref{sec:task}.} The best strict-target performance anywhere in Table \ref{tab:oana_prediction_accuracy_next} is 35.4/33.3\% on action type and 23.2/22.2\% on action type plus all parameters, from the Llama 70B with OAS + KE on the robot side and with OAS alone on the human side. The 8B never exceeds 26.2\% on action type or 10.4\% on the strictest metric. Agents required to predict their partner are therefore acting alongside an anticipation that is wrong roughly two-thirds of the time at the coarsest granularity and three-quarters of the time at the finest, even in the best-resourced configuration. That OANA prediction lowers Success Rate in fifteen of sixteen model$\times$context pairs is considerably less surprising in this light, and the monotone relationship we reported between prediction accuracy and the size of that penalty for DeepSeek becomes the more natural reading: prediction is not merely an overhead but an overhead built on a mostly incorrect signal.

In summary, the strict target leaves the mechanism-level conclusions of Section~\ref{sec:prediction} intact (the factorization, the parameter bottleneck, the planner dependence of how KE is used, and the 8B's inability to convert context into complete predictions), while substantially revising the agent-level ones. The asymmetries we attributed to the structure of the two action spaces turn out to depend on crediting ongoing actions, and next-action anticipation is both considerably weaker than the union numbers suggest and more evenly distributed between the two agents than those numbers imply.

\subsubsection{Action Prediction Evaluations for Full Observability}
\label{sec:prediction_full}

\begin{table*}[t]
\centering
\setlength{\tabcolsep}{1.6pt}
\renewcommand{\arraystretch}{1.6}
\scriptsize

\caption{\textbf{Full observability OANA prediction accuracy (\%)} across action abstraction levels for robot (R) and human (H) agents. The target action is defined as the \textbf{union of the current and next actions} of the other agent. Results show that allowing OAS observation mostly improves prediction accuracy in the case of the 8B. Using Llama-3.1-Instruct models as planners for the agents.}
\label{tab:oana_prediction_accuracy_union}

\begin{tabularx}{\textwidth}{
>{\centering\arraybackslash}p{1.0cm} 
>{\centering\arraybackslash}p{0.6cm} 
>{\centering\arraybackslash}p{1.6cm} 
>{\centering\arraybackslash}p{1.9cm} 
>{\centering\arraybackslash}p{2.0cm} 
>{\centering\arraybackslash}p{1.9cm} 
>{\centering\arraybackslash}p{2.3cm} 
>{\centering\arraybackslash}p{2.2cm} 
>{\centering\arraybackslash}p{2.0cm} 
}
\textbf{Model Size} & \textbf{OAS} 
& \textbf{Action Type (R/H) $\uparrow$} 
& \textbf{Principal Par. (R/H) $\uparrow$} 
& \textbf{A. Type+Princ. Par. (R/H) $\uparrow$} 
& \textbf{All Params (R/H) $\uparrow$} 
& \textbf{A. Type+All Params (R/H) $\uparrow$} 
& \textbf{Principal Par. if Type (R/H) $\uparrow$} 
& \textbf{All Params if Type (R/H) $\uparrow$} \\
\hline

\multirow{2}{*}{8B}
& \xmark 
& 29.3 / 33.2 
& 26.6 / \underline{28.6} 
& 17.0 / 20.9 
& 17.1 / 18.6 
& 14.1 / 16.4 
& 58.1 / \underline{62.7} 
& 48.1 / 49.3 \\

& \cmark 
& \underline{33.4} / \underline{35.5} 
& \underline{27.2} / 28.3 
& \underline{20.1} / \underline{22.1} 
& \underline{19.1} / \underline{20.0} 
& \underline{16.9} / \underline{18.4} 
& \underline{60.3} / 62.3 
& \underline{50.5} / \underline{51.6} \\

\hline

\multirow{2}{*}{70B}
& \xmark 
& 57.0 / 64.2 
& 49.7 / \textbf{\underline{53.3}} 
& 40.8 /  \textbf{\underline{44.6}} 
& 40.8 /  \textbf{\underline{44.4}} 
& 35.5 /  \textbf{\underline{39.3}} 
& \textbf{\underline{71.5}} / \textbf{\underline{69.5}} 
& \textbf{\underline{62.3}} / \textbf{\underline{61.3}} \\

& \cmark 
& \textbf{\underline{65.6}} / \textbf{\underline{70.8}} 
& \textbf{\underline{50.2}} / 48.2 
& \textbf{\underline{44.5}} / 43.2 
& \textbf{\underline{42.7}} / 41.7 
& \textbf{\underline{38.5}} / 38.5 
& 67.8 / 60.9 
& 58.7 / 54.3 \\

\hline
\label{tab:oana_pred_full}
\end{tabularx}
\end{table*}

\begin{table*}[t]
\centering
\setlength{\tabcolsep}{1.6pt}
\renewcommand{\arraystretch}{1.6}
\scriptsize

\caption{\textbf{Full observability OANA prediction accuracy (\%)} across action abstraction levels for robot (R) and human (H) agents. The target action is defined as the \textbf{next action} of the other agent. Results show that allowing Other Agent's State (OAS) observation mostly improves prediction accuracy in the case of the 8B model, except for the Principal Parameter, while for the 70B model, it hinders the performance. Using Llama-3.1-Instruct models as planners for the agents.}
\label{tab:oana_prediction_accuracy_full}

\begin{tabularx}{\textwidth}{
>{\centering\arraybackslash}p{1.0cm} 
>{\centering\arraybackslash}p{0.6cm} 
>{\centering\arraybackslash}p{1.6cm} 
>{\centering\arraybackslash}p{1.9cm} 
>{\centering\arraybackslash}p{2.0cm} 
>{\centering\arraybackslash}p{1.9cm} 
>{\centering\arraybackslash}p{2.3cm} 
>{\centering\arraybackslash}p{2.2cm} 
>{\centering\arraybackslash}p{2.0cm} 
}
\textbf{Size} & \textbf{OAS} 
& \textbf{Action Type (R/H) $\uparrow$} 
& \textbf{Principal Par. (R/H) $\uparrow$} 
& \textbf{A. Type+Princ. Par. (R/H) $\uparrow$} 
& \textbf{All Params (R/H) $\uparrow$} 
& \textbf{A. Type+All Params (R/H) $\uparrow$} 
& \textbf{Principal Par. if Type (R/H) $\uparrow$} 
& \textbf{All Params if Type (R/H) $\uparrow$} \\
\hline

\multirow{2}{*}{8B}
& \xmark 
& 23.3 / 26.3 
& \underline{22.9} / \underline{23.5}
& 14.6 / 17.6 
& 13.5 / 14.1 
& 11.7 / 13.3 
& 62.5 / 67.2 
& 50.1 / 50.7 \\

& \cmark 
& \underline{23.4} / \underline{26.7} 
& 21.7 / 23.0 
& \underline{15.5} / \underline{18.6} 
& \underline{13.8} / \underline{15.4} 
& \underline{12.2} / \underline{14.8} 
& \underline{66.1} / \underline{69.5} 
& \underline{52.4} / \underline{55.5} \\

\hline

\multirow{2}{*}{70B}
& \xmark 
& \textbf{\underline{43.8}} / \textbf{\underline{45.9}} 
& \textbf{\underline{41.8}} / \textbf{\underline{41.4}} 
& \textbf{\underline{35.3}} / \textbf{\underline{36.5}} 
& \textbf{\underline{33.2}} / \textbf{\underline{33.3}} 
& \textbf{\underline{30.3}} / \textbf{\underline{31.3}} 
& 80.6 /  \textbf{\underline{79.5}} 
& \textbf{\underline{69.2}} / \textbf{\underline{68.1}} \\

& \cmark 
& 43.5 / 43.1 
& 38.6 / 36.3 
& \textbf{\underline{35.3}} / 33.9 
& 31.4 / 30.5 
& 29.5 / 29.3 
& \textbf{\underline{81.2}} / 78.5 
& 67.8 / 67.9 \\

\hline
\label{tab:oana_pred_full_next}
\end{tabularx}
\end{table*}


Tables~\ref{tab:oana_pred_full} and~\ref{tab:oana_pred_full_next} report prediction accuracy under full observability for the union and the strict next-action targets, respectively, and Tables~\ref{tab:oana_prediction_current_next_action_dp_full} and~\ref{tab:oana_prediction_next_action_dp_full} place both alongside the corresponding Decoupled Partial Observability runs. Object KE is again inapplicable, since full observability already supplies every object position at all times, so the design is a $2\times2$ of OAS observation and target definition, run with the two Llama planners. Crossing the observability regime with target definition is what makes these runs informative: the regime controls how much the predictor knows about the \emph{world}, and the target controls whether credit requires recognizing the partner's \emph{present} action or anticipating its \emph{future} one. The joint metrics again factorize, to within rounding, as $\mathrm{Acc}(\text{type})\times\mathrm{Acc}(\text{params}\mid\text{type})$ in all 32 cells of the two tables, so we continue to attribute every movement to the type term or the conditional term.

\textbf{Full observability produces the strongest partner prediction anywhere in this work, but only for the larger planner.} The Llama 70B reaches 65.6/70.8\% Action Type accuracy under the union target with OAS, against a best of 55.5/55.8\% across all Decoupled Partial configurations, and 43.8/45.9\% under the strict target without OAS, against 35.4/33.3\% in the partial regime. The gains are $+10.1/{+}15.0$ and $+8.4/{+}12.6$ points, respectively, and they extend to the strictest joint metric ($+4.4/{+}6.1$ under union, $+7.1/{+}9.1$ under strict). Notably, full observability \emph{without} OAS already exceeds the best Decoupled Partial configuration with both OAS and KE on all eight R/H comparisons we can make, so complete object knowledge substitutes for, and then surpasses, the co-location-gated channels, at least for a planner able to use it.

\textbf{For the smaller planner, the same intervention improves parameters and not verbs, and on the robot side, it makes coarse prediction worse.} The Llama 8B's best Action Type accuracy falls from 37.9\% to 33.4\% (union) and from 26.2\% to 23.4\% (strict) on the robot side when moving from its best partial-observability configuration to full observability, while rising modestly on the human side ($+1.3$ and $+3.1$). Its conditional accuracies, by contrast, improve substantially: conditional principal-parameter accuracy rises by +5.0/+5.5 points under the union target and by +15.3/+13.1 under the strict target (50.8/56.4\%$\rightarrow$66.1/69.5\%). The decomposition localizes this cleanly. Perfect object knowledge resolves the ambiguity that the conditional term measures (which objects and receptacles are involved) and does nothing for the question of which verb the partner will choose, which for this planner is the binding constraint. The small robot-side regression on Action Type is consistent with the prompt-length effect noted in Appendix~\ref{app:partial_pred}: full observability enumerates every object in the scene, and the 8B pays for that context rather than exploiting it.

\textbf{Crossing the two targets isolates what OAS observation contributes, and under full observability, it contributes essentially nothing but present-state recognition.} For the Llama 70B, enabling OAS raises union-target Action Type by $+8.6/{+}6.6$ points but changes strict-target Action Type by $-0.3/{-}2.8$. The same intervention that appears to improve partner prediction by 6.6 to 8.6 points does not improve anticipation at all. This directly tests the mechanism proposed in Appendix \ref{subsubsec:appendix_action_pred_eval_next}: OAS observations describe what the partner is doing right now (\textit{Walking in hallway\_1}, \textit{Picking apple\_0}), so they let an agent name the partner's ongoing action, which the union target credits and the strict target does not. Under Decoupled Partial Observability, roughly half of the OAS benefit survived the strict target (70B: +14.4/+14.0 under union against +7.2/+6.4 under strict), because there, OAS also carried object information that was otherwise unavailable. Once full observability supplies that information independently, OAS only reports the partner's current state.

\begin{table*}[t]
\centering
\setlength{\tabcolsep}{1.4pt}
\renewcommand{\arraystretch}{1.8}
\scriptsize

\caption{\textbf{OANA prediction accuracy (\%)} comparison across Decoupled Partial and Full observability settings for robot (R) and human (H) agents. The target action is the \textbf{union of the current and next actions} of the other agent. Results show that full observability improves prediction for the Llama 70B and the 8B human agent, while the 8B robot's Action Type accuracy falls below its best partial-observability configuration. Subgroup bests (per model scale) are underlined, and overall bests are bolded.}
\label{tab:oana_prediction_current_next_action_dp_full}

\begin{tabularx}{\textwidth}{
>{\centering\arraybackslash}p{1.4cm} 
>{\centering\arraybackslash}p{2.2cm} 
>{\centering\arraybackslash}p{0.6cm} 
>{\centering\arraybackslash}p{0.5cm} 
>{\centering\arraybackslash}p{0.8cm} 
>{\centering\arraybackslash}X 
>{\centering\arraybackslash}X 
>{\centering\arraybackslash}X 
>{\centering\arraybackslash}X 
>{\centering\arraybackslash}X 
>{\centering\arraybackslash}X 
}
\hline
\multirow{2}{*}{\textbf{Model Size}} & \multirow{2}{*}{\textbf{Observability}} & \multirow{2}{*}{\textbf{OAS}} & \multirow{2}{*}{\textbf{KE}} & \multirow{2}{*}{\textbf{OANA}} & \multicolumn{2}{c}{\textbf{Action Type $\uparrow$}} & \multicolumn{2}{c}{\textbf{A. Type + Princ. Param. $\uparrow$}} & \multicolumn{2}{c}{\textbf{A. Type + All Param. $\uparrow$}} \\
\cline{6-11}
& & & & & \textbf{R} & \textbf{H} & \textbf{R} & \textbf{H} & \textbf{R} & \textbf{H} \\
\hline

\multirow{5}{*}{8B} 
& Decoupled Partial & \xmark & \xmark & \cmark & 31.8 & 29.5 & 17.2 & 15.8 & 15.7 & 14.1 \\
& Decoupled Partial & \xmark & \cmark & \cmark & 33.3 & 31.1 & 18.4 & 17.8 & 16.6 & 15.4 \\
& Decoupled Partial & \cmark & \xmark & \cmark & 36.1 & 33.8 & 17.8 & 17.9 & 16.4 & 15.8 \\
& Decoupled Partial & \cmark & \cmark & \cmark & \underline{37.9} & 34.2 & 18.9 & 17.2 & \underline{16.9} & 15.5 \\
& Full & \xmark & - & \cmark & 29.3 & 33.2 & 17.0 & 20.9 & 14.1 & 16.4 \\
& Full & \cmark & - & \cmark & 33.4 & \underline{35.5} & \underline{20.1} & \underline{22.1} & \underline{16.9} & \underline{18.4} \\

\hline

\multirow{5}{*}{70B} 
& Decoupled Partial & \xmark & \xmark & \cmark & 38.5 & 41.8 & 24.1 & 26.5 & 22.1 & 24.3 \\
& Decoupled Partial & \xmark & \cmark & \cmark & 43.6 & 49.6 & 29.5 & 32.0 & 26.5 & 28.7 \\
& Decoupled Partial & \cmark & \xmark & \cmark & 52.9 & 55.8 & 35.2 & 34.7 & 31.7 & 32.3 \\
& Decoupled Partial & \cmark & \cmark & \cmark & 55.5 & 55.8 & 37.3 & 35.8 & 34.1 & 33.2 \\
& Full & \xmark & - & \cmark & 57.0 & 64.2 & 40.8 & \textbf{\underline{44.6}} & 35.5 & \textbf{\underline{39.3}} \\
& Full & \cmark & - & \cmark & \textbf{\underline{65.6}} & \textbf{\underline{70.8}} & \textbf{\underline{44.5}} & 43.2 & \textbf{\underline{38.5}} & 38.5 \\

\hline
\end{tabularx}
\end{table*}

\begin{table*}[t]
\centering
\setlength{\tabcolsep}{1.4pt}
\renewcommand{\arraystretch}{1.8}
\scriptsize

\caption{\textbf{OANA prediction accuracy (\%)} comparison across Decoupled Partial and Full observability settings for robot (R) and human (H) agents. The target action is the \textbf{next action} of the other agent. Results highlight that full observability improves prediction for the Llama 70B and the 8B human agent, while the 8B robot's Action Type accuracy falls below its best partial-observability configuration; under full observability, adding OAS does not improve the 70B's next-action prediction. Subgroup bests (per model scale) are underlined, and overall bests are bolded.}
\label{tab:oana_prediction_next_action_dp_full}

\begin{tabularx}{\textwidth}{
>{\centering\arraybackslash}p{1.4cm} 
>{\centering\arraybackslash}p{2.2cm} 
>{\centering\arraybackslash}p{0.6cm} 
>{\centering\arraybackslash}p{0.5cm} 
>{\centering\arraybackslash}p{0.8cm} 
>{\centering\arraybackslash}X 
>{\centering\arraybackslash}X 
>{\centering\arraybackslash}X 
>{\centering\arraybackslash}X 
>{\centering\arraybackslash}X 
>{\centering\arraybackslash}X 
}
\hline
\multirow{2}{*}{\textbf{Model Size}} & \multirow{2}{*}{\textbf{Observability}} & \multirow{2}{*}{\textbf{OAS}} & \multirow{2}{*}{\textbf{KE}} & \multirow{2}{*}{\textbf{OANA}} & \multicolumn{2}{c}{\textbf{Action Type $\uparrow$}} & \multicolumn{2}{c}{\textbf{A. Type + Princ. Param. $\uparrow$}} & \multicolumn{2}{c}{\textbf{A. Type + All Param. $\uparrow$}} \\
\cline{6-11}
& & & & & \textbf{R} & \textbf{H} & \textbf{R} & \textbf{H} & \textbf{R} & \textbf{H} \\
\hline

\multirow{5}{*}{8B} 
& Decoupled Partial & \xmark & \xmark & \cmark & 23.1 & 20.7 & 11.5 & 10.4 & 10.1 & 8.8 \\
& Decoupled Partial & \xmark & \cmark & \cmark & 24.0 & 21.6 & 12.2 & 12.2 & 10.4 & 9.9 \\
& Decoupled Partial & \cmark & \xmark & \cmark & 25.8 & 23.6 & 11.3 & 12.3 & 10.0 & 10.2 \\
& Decoupled Partial & \cmark & \cmark & \cmark & \underline{26.2} & 23.4 & 11.9 & 11.4 & 9.9 & 9.6 \\
& Full & \xmark & - & \cmark & 23.3 & 26.3 & 14.6 & 17.6 & 11.7 & 13.3 \\
& Full & \cmark & - & \cmark & 23.4 & \underline{26.7} & \underline{15.5} & \underline{18.6} & \underline{12.2} & \underline{14.8} \\

\hline

\multirow{5}{*}{70B} 
& Decoupled Partial & \xmark & \xmark & \cmark & 27.5 & 26.9 & 17.3 & 18.5 & 15.4 & 16.5 \\
& Decoupled Partial & \xmark & \cmark & \cmark & 30.7 & 31.2 & 21.5 & 21.4 & 18.6 & 18.4 \\
& Decoupled Partial & \cmark & \xmark & \cmark & 34.7 & 33.3 & 25.0 & 24.5 & 21.7 & 22.2 \\
& Decoupled Partial & \cmark & \cmark & \cmark & 35.4 & 32.6 & 26.3 & 24.4 & 23.2 & 21.9 \\
& Full & \xmark & - & \cmark & \textbf{\underline{43.8}} & \textbf{\underline{45.9}} & \textbf{\underline{35.3}} & \textbf{\underline{36.5}} & \textbf{\underline{30.3}} & \textbf{\underline{31.3}} \\
& Full & \cmark & - & \cmark & 43.5 & 43.1 & \textbf{\underline{35.3}} & 33.9 & 29.5 & 29.3 \\

\hline
\end{tabularx}
\end{table*}

\textbf{The union-to-strict accuracy loss is accordingly concentrated in the configurations where OAS is enabled.} Under full observability, the relative drop from union to strict target is 20.5--28.5\% without OAS and 24.8--39.1\% with it, the two largest values in the table being the 70B's OAS-enabled cells ($-33.7\%$ and $-39.1\%$). Averaged over the two Llama planners, the drop is 27.6\% under full observability against 32.5\% under Decoupled Partial Observability, so complete world knowledge makes a larger share of measured prediction accuracy reflect anticipation rather than state recognition, while adding OAS on top pushes it back the other way. Any evaluation of partner modeling that credits the ongoing action should therefore report the strict variant alongside it, because the two scorings respond in opposite directions to the same intervention.

\textbf{Under full observability, OAS degrades the conditional term for the larger planner, which mirrors its behavior in task solving.} With the union target, enabling OAS costs the 70B $-3.7/{-}8.6$ points of conditional principal-parameter accuracy and $-3.6/{-}7.0$ of conditional all-parameter accuracy, and it lowers the human agent's unconditional Principal Par., A. Type+Princ. Par., All Params, and A. Type+All Params scores as well. The pattern matches what we report for task performance in Appendix \ref{sec:appendix_task_analysis_full}: once perceptual uncertainty is removed, OAS retains the functions that are intrinsic to partner state and loses the ones that were compensating for missing world knowledge, while its prompt cost remains. For the 8B, whose baseline competence is lower, OAS still helps on both targets, so the cost only dominates where the remaining headroom is small.

\textbf{Full observability largely dissolves the parameter bottleneck while leaving the type bottleneck intact.} The highest conditional accuracies in this paper occur under full observability with the strict target: 80.6/79.5\% and 81.2/78.5\% conditional principal-parameter accuracy for the 70B, against 74.4/75.0\% at its best in the partial regime and 67.6/64.5\% under the union target there. Roughly one correctly typed prediction in five now carries a wrong principal parameter, down from about one in three. Unconditional Action Type under the strict target is 43.8/45.9\% without OAS and 43.5/43.1\% with it. An agent with a complete, correct, and continuously updated model of the world, predicting a partner acting in that same fully known world, still names the wrong next action type more than half the time. The residual difficulty in partner modeling is therefore not informational in the sense our interventions address: it concerns which of several individually plausible actions a partner will select, and neither partner-state observation, nor object knowledge exchange, nor complete observability resolves it.

\textbf{The human's coarse-prediction advantage reappears under full observability and survives the strict target, consistent with the action-space-size account holding once object uncertainty is removed.} The human agent leads on Action Type in all four union-target cells ($+3.9$, $+2.1$, $+7.2$, $+5.2$ points H$-$R) and in three of four strict-target cells ($+3.0$, $+3.3$, $+2.1$, $-0.4$). This contrasts with Decoupled Partial Observability, where the robot matched or led in 14 of 16 strict-target cells. The resolution the two regimes together suggest is that the type-space advantage is real but only expressible once object uncertainty is removed: under partial observability, not knowing where things are dominates the prediction problem for both agents and masks the difference in how many verbs each must choose among; under full observability, that source of difficulty is gone, and the smaller hypothesis space shows through. This account rests on two planners of one family.

\textbf{The conditional asymmetry, by contrast, is stable in direction within each planner and opposite between them.} Under full observability, the 8B human leads on conditional principal-parameter accuracy in all four cells ($+4.6$, $+2.0$, $+4.7$, $+3.4$ points H$-$R across both targets), while the 70B robot leads in all four ($-2.0$, $-6.9$, $-1.1$, $-2.7$). The 8B ordering matches what it showed under the strict target in the partial regime, whereas the 70B robot led in only one of four contexts there (Appendix \ref{subsubsec:appendix_action_pred_eval_next}), so its lead is stable under full observability but not across regimes. The split is thus a planner property rather than a target-definition artifact, which is a further reason to treat the arity-based account offered in Section~\ref{sec:prediction} as unconfirmed: an explanation grounded in the structure of the two action spaces should predict the same direction for both planners, and it does not.

Overall, full observability helps a capable planner much more than a weak one: it largely solves the parameter part of the prediction problem, leaves the action-type part largely unsolved, and reduces OAS observation to information about the partner's present state, which improves scores that credit present-state recognition but not anticipation. The best strict-target accuracy we measure anywhere (43.8/45.9\% at the action-type level, 30.3/31.3\% with all parameters) is attained by agents that know everything about the world and are told nothing about their partner. Even in this best case, more than half of the predicted action types are wrong, so the OANA cost reported in Section \ref{sec:task} comes from acting on mostly incorrect predictions.

\subsection{Prompts for the agents}\label{app:prompts}
\label{sec:appendix_prompts}

See the prompts below, which were used for the agents in various experiments. We are not using the \textit{\{rag\_examples\}} tag, since we are using zero-shot planners. The \textit{\{input\}} tag would be substituted with the task description, e.g., \textit{Place the candle and candle holder on the white table in the living room. Move the plant to the same table.}  

The instruction prompt for the Decentralized ReAct agents \textbf{without} OANA prediction (the PARTNR baseline prompt; \citealp{Chang2024PARTNR:}): 
\begin{promptbox}{Decentralized | Zero-Shot | ReAct | Without OANA prediction}
\{system\_tag\}You are an agent that solves multi-agent planning problems. The task assigned to you will be situated in a house and will generally involve navigating to objects, picking and placing them on different receptacles to achieve rearrangement. You strictly follow any format specifications and pay attention to the previous actions taken in order to avoid repeating mistakes. Rooms do not need to be explored more than once.
\\ \\
        There will be another agent trying to solve the same task that you are at the same time. You may find that that agent has picked up relevant objects or is in the process of completing parts of the task. If that is the case you may want to move on to a different part of the task.
\\ \\
        Rooms do not need to be explored more than once.
        This means if you have explored the living room and have not found the object, then you should explore the kitchen, if a relevant object is still not found, you should explore the hallway etc...
\\ \\
        \{agent\_role\_description\}
\\ \\
        Many calls to the same action in a row are a sign that something has gone wrong and you should try a different action.\{eot\_tag\}\{rag\_examples\}\{user\_tag\}Task: \{input\}
\\ \\
        \{world\_description\}
\\ \\
        Possible Actions: \\
        \{tool\_descriptions\} \\
        - Done: Used to indicate that the agent has finished the task. Example (Done[])
\\ \\
        What is the next action to make progress towards completing the task? \\
        Return your response in the following format
\\ \\
        Thought: <reasoning for why you are taking the next action> \\
        <next action call> \\
        Assigned!
\\ \\
        Here is an example: \\
        Thought: Since there are no objects found I should explore a room I have not explored yet \\
        Explore[<room name>] \\
        Assigned! \\
        \{eot\_tag\}\{assistant\_tag\}
\end{promptbox}

Our instruction prompt for the Decentralized ReAct agents \textbf{with} OANA prediction is shown below. Since the human uses extra object state modifier actions as well, the \textit{\{your\_tool\_descriptions\}} and the \textit{\{other\_agent\_tool\_descriptions\}} would be different based on the given roles (see the tool description later).
\begin{promptbox}{ Decentralized | Zero-Shot | ReAct | With OANA prediction}
\{system\_tag\}You are an agent (Agent A) that solves multi-agent planning problems. The task assigned to you will be situated in a house and will generally involve navigating to objects, picking and placing them on different receptacles to achieve rearrangement. You strictly follow any format specifications and pay attention to the previous actions taken in order to avoid repeating mistakes. Rooms do not need to be explored more than once.
\\ \\
        There will be another agent (Agent B) trying to solve the same task that you are at the same time. You may find that that agent has picked up relevant objects or is in the process of completing parts of the task. If that is the case, you may want to move on to a different part of the task. \\
        The other agent (Agent B) is not controlled by you, but you are working together to solve the common task, so you should predict what the other agent (Agent B) will do next and take that into account when deciding your next action.
\\ \\
        Rooms do not need to be explored more than once.
        This means if you have explored the living room and have not found the object, then you should explore the kitchen, if a relevant object is still not found, you should explore the hallway etc...
\\ \\
        \{agent\_role\_description\}
\\ \\
        Many calls to the same action in a row are a sign that something has gone wrong and you should try a different action.\{eot\_tag\}\{rag\_examples\}\{user\_tag\}Task: \{input\}
\\ \\
        \{world\_description\}
\\ \\
        Your (Agent A) possible actions: \\
        \{your\_tool\_descriptions\} \\
        - Done: Used to indicate that the agent has finished the task. Example (Done[])
\\ \\
        The other agent's (Agent B) possible actions: \\
        \{other\_agent\_tool\_descriptions\} \\
        - Done: Used to indicate that the agent has finished the task. Example (Done[])
\\ \\
        What is your prediction of the other agent's (Agent B) next action to make progress towards completing the task? \\
        What is your (Agent A) next action to make progress towards completing the task? \\
        Return your response in the following format
\\ \\
        Thought: <reasoning for what is the other agent's (Agent B) next action and why you (Agent A) are taking your next action> \\
        The other agent's predicted next action: <other agent's (Agent B) predicted next action call> \\ 
        My next action: <your (Agent A) next action call> \\
        Assigned!
\\ \\
        Here is an example: \\
        Thought: The other agent is probably exploring a room for the objects, but since no objects were found, I should explore a room too, which I have not explored yet \\
        The other agent's predicted next action: Explore[<room name>] \\
        My next action: Explore[<room name>] \\
        Assigned! \\
        \{eot\_tag\}\{assistant\_tag\}
\end{promptbox}

The agent’s role description depends on whether it is assigned the role of the human or the robot, and would be specified accordingly \citep{Chang2024PARTNR:}: 
\begin{promptbox}{Agent Role Description}
\textbf{Human's Role Description:}
\\ \\
You are playing the role of the task giver. This means if the instruction says something like "You should move the object and I will wash it", then the other agent should be moving the object, and you should washing the it.
\\ \\
\textbf{Robot's Role Description:}
\\ \\
You are playing the role of the task receiver. This means if the instruction says something like "You should move the object and I will wash it", then you should move the object and the other agent should wash it.
\end{promptbox}

The action tool descriptions are shown below (from \citealp{Chang2024PARTNR:}).
\begin{promptbox}{Tool Description}
- Clean: Used for cleaning an object. You need to provide the name of the object to clean. \\
- Close: Used for closing an articulated entity. You must provide the name of the furniture you want to close. Example (Close[chest\_of\_drawers\_1]) \\
- Explore: Search a specific room by visiting various receptacles or furnitures in that room. The input to the skill is the EXACT name of the room to be visited. Use the room names provided in the house description. This tool exhaustivly explores the specified room. Example (Explore[kitchen\_1]) \\
- Fill: Used for filling an object. You need to provide the name of the object to fill. \\
- Navigate: Used for navigating to an entity. You must provide the name of the entity you want to navigate to. Example (Navigate[counter\_22]) \\
- Open: Used for opening an articulated entity. You must provide the name of the furniture you want to open. Example (Open[chest\_of\_drawers\_1]) \\
- Pick: Used for picking up an object. You must provide the name of the object to be picked. The agent cannot hold more than one object at a time. Example (Pick[cup\_1]) \\
- Place: Used for placing an object on a target location. You need to provide the name of the object to be placed, the name of the furniture where it should be placed, spatial relation ("on" or "within") describing the relation between the object and furniture. The object to be placed must already be held by the agent (i.e. picked previously). In addition to these, you can request to place the object near another object. For that you can optionally provide a spatial constraints ("next\_to") and the name of the reference object. To place next to an object, the reference object must already be on the target furniture. API template - Place[<object\_to\_be\_placed>, <spatial\_relation>, <furniture to be placed on>, <spatial\_constraint>, <reference\_object>]. spatial\_constraint and reference\_object should be set to "None" when necessary. \\
- Pour: Used for pouring from one container to another. This skill will pour into the specified container from whichever container is currently held by the agent. \\
- PowerOff: Used for turning off a powered object. You need to provide the name of the object to be turned off. \\
- PowerOn: Used for turning on a powered object. You need to provide the name of the object to be turned on. \\
- Rearrange: Used for moving an object from its current location to the target location. You need to provide the name of the object to be moved, the name of the furniture where is should be moved, spatial relation ("on" or "within") describing the relation between the object and furniture. This will automatically pick the specified object and move to the target furniture and attempt to place it. In addition to these, you can request to place the object near another object. For that you can optionally provide a spatial constraint ("next\_to") and the name of the reference object. To place next to an object, the reference object must already be on the target furniture. API template Rearrange[<object\_to\_be\_moved>, <spatial\_relation>, <furniture to be placed on>, <spatial\_constraint>, <reference\_object>]. spatial\_constraint and reference\_object should be set to "None" when necessary. \\
- Wait: Used to make agent stay idle for some time. Example (Wait[]) \\
- Done: Used to indicate that the agent has finished the task. Example (Done[])
\end{promptbox}

Only the human has access to these tools, which modify the object state on top of the base tools. The tool descriptions for the agents differ on this (from \citealp{Chang2024PARTNR:}). 
\begin{promptbox}{Human Only Tool Descriptions}
- Clean: Used for cleaning an object. You need to provide the name of the object to clean. \\
- Fill: Used for filling an object. You need to provide the name of the object to fill. \\
- Pour: Used for pouring from one container to another. This skill will pour into the specified container from whichever container is currently held by the agent. \\
- PowerOff: Used for turning off a powered object. You need to provide the name of the object to be turned off. \\
- PowerOn: Used for turning on a powered object. You need to provide the name of the object to be turned on. \\
\end{promptbox}

The world description provides a textual overview of all rooms, their furniture, and the objects currently observed. In (Decoupled) Partial observability, the objects are not known in advance, and exploration is required. An example of a scene is shown below:
\begin{promptbox}{World Description Example}
Furniture:
living\_room\_1: floor\_living\_room\_1, table\_10, couch\_18, shelves\_26, chair\_28, chair\_29, table\_32, chair\_33, chair\_34, chair\_35, chair\_36, table\_46, table\_47, stand\_55, chest\_of\_drawers\_56 \\
other\_room\_1: floor\_other\_room\_1 \\
bathroom\_1: floor\_bathroom\_1, chair\_23, table\_37, unknown\_38, bench\_48 \\
bathroom\_2: floor\_bathroom\_2, unknown\_14, table\_50 \\
bedroom\_1: floor\_bedroom\_1, chair\_16, table\_22, bed\_45, bench\_49, chest\_of\_drawers\_52, chest\_of\_drawers\_54 \\
laundryroom\_1: floor\_laundryroom\_1, washer\_dryer\_11, washer\_dryer\_12, shelves\_15 \\
entryway\_1: floor\_entryway\_1, bench\_51 \\
bedroom\_2: floor\_bedroom\_2, table\_13, chair\_17, bed\_21, table\_24, table\_25, chair\_27, stool\_31, bench\_44, table\_59, chest\_of\_drawers\_61 \\
hallway\_1: floor\_hallway\_1, table\_30 \\
kitchen\_1: floor\_kitchen\_1, counter\_19, chair\_20, cabinet\_39, cabinet\_40, chair\_41, chair\_42, chair\_43, counter\_53, cabinet\_57, fridge\_58, unknown\_60
\\ \\
The following furnitures have a faucet: cabinet\_57 \\
Objects: \\
No objects found yet
\end{promptbox}

For the experiments, we use a summarized representation of the scene; after each action call, we append the textual description of the objects observed by the agent. Note that the agents could move through different rooms in the house in order to get to another room and could discover and observe objects meanwhile.
\begin{promptbox}{Scene Summary Example Object Only Baseline}
<|start\_header\_id|>assistant<|end\_header\_id|>
\\ \\
Thought: Since there are no objects found yet, I should explore a room I have not explored yet. The task requires the candle, candle holder, and plant, so I'll start by exploring the living room to see if I can find any of these objects. \\
Explore[living\_room\_1] \\
Assigned!<|eot\_id|><|start\_header\_id|>user<|end\_header\_id|> \\
\\ 
Result: Successful execution! \\
Objects: candle\_holder\_1: table\_30 in hallway\_1 \\
candle\_0: table\_30 in hallway\_1 \\
plant\_container\_2: table\_30 in hallway\_1 \\
box\_5: table\_46 in living\_room\_1 \\
jug\_4: chair\_27 in bedroom\_2 \\
box\_3: washer\_dryer\_12 in laundryroom\_1<|eot\_id|> \\<|start\_header\_id|>assistant<|end\_header\_id|> \\
\\
Thought: I found the candle, candle holder, and plant container in the hallway, which is one of the rooms I needed to explore. Now that I have the objects, I can start working on the task. I'll navigate to the living room to pick up the candle holder, candle, and plant container. \\
Navigate[table\_30] \\
Assigned!<|eot\_id|><|start\_header\_id|>user<|end\_header\_id|> \\
\\
Result: Successful execution! \\
Objects: candle\_holder\_1: table\_30 in hallway\_1 \\
candle\_0: table\_30 in hallway\_1 \\
plant\_container\_2: table\_30 in hallway\_1 \\
box\_5: table\_46 in living\_room\_1 \\
jug\_4: chair\_27 in bedroom\_2 \\
box\_3: washer\_dryer\_12 in laundryroom\_1<|eot\_id|>
\end{promptbox}

In the case of Other Agent's State Observation, the observed OAS is added to the textual observation description. If the Other Agent's Next Action Prediction is also enabled, then the prediction is made between the reasoning (Thought) and the action tool call at each replanning. 
\begin{promptbox}{Scene Summary Example With OAS Observation and OANA Prediction}
<|start\_header\_id|>assistant<|end\_header\_id|>

Thought: Since the task requires placing the candle and candle holder on the white table in the living room, and moving the plant to the same table, I predict that Agent B will start by exploring the living room to find the objects. As Agent A, I will also explore a room to find the objects, but I will choose a different room to avoid duplication of effort. I will start by exploring the kitchen, as it is a common place to find objects like candles and plants. \\
The other agent's predicted next action: Explore[living\_room\_1] \\
My next action: Explore[kitchen\_1] \\
Assigned!<|eot\_id|><|start\_header\_id|>user<|end\_header\_id|>
\\ \\
Result of your action: Successful execution! \\
The other agent's actions since the previous observation: Acting in an other room, Walking in living\_room\_1, Acting in an other room, Walking in living\_room\_1, Acting in an other room, Walking in living\_room\_1, Acting in an other room \\
Your observations: \\
Objects: jug\_4: chair\_27 in bedroom\_2 \\
vase\_7: table\_13 in bedroom\_2 \\
clock\_6: chair\_17 in bedroom\_2 \\
candle\_holder\_1: table\_30 in hallway\_1 \\
candle\_0: table\_30 in hallway\_1 \\
plant\_container\_2: table\_30 in hallway\_1 \\
box\_5: table\_46 in living\_room\_1 \\
box\_3: washer\_dryer\_12 in laundryroom\_1<|eot\_id|> \\<|start\_header\_id|>assistant<|end\_header\_id|> \\
\\
Thought: Since I found some objects in the kitchen, but not the ones required for the task, I will continue exploring other rooms to find the remaining objects. I noticed that Agent B was walking in the living room, but didn't find any objects there. I also found some objects in the hallway, including the candle holder and candle, which are required for the task. I will navigate to the hallway to pick up these objects. \\
The other agent's predicted next action: Explore[bedroom\_2] \\
My next action: Navigate[table\_30] \\
Assigned!<|eot\_id|><|start\_header\_id|>user<|end\_header\_id|> \\
\\
Result of your action: Successful execution! \\
The other agent's actions since the previous observation: Walking in living\_room\_1, Acting in an other room \\
Your observations: \\
Objects: jug\_4: chair\_27 in bedroom\_2 \\
vase\_7: table\_13 in bedroom\_2 \\
clock\_6: chair\_17 in bedroom\_2 \\
candle\_holder\_1: table\_30 in hallway\_1 \\
candle\_0: table\_30 in hallway\_1 \\
plant\_container\_2: table\_30 in hallway\_1 \\
box\_5: table\_46 in living\_room\_1 \\
box\_3: washer\_dryer\_12 in laundryroom\_1<|eot\_id|>
\end{promptbox}

\end{document}